%% file: main.tex
\documentclass[11pt]{article}
\input{commands}

\title{Solving Pasur Using GPU-Accelerated Counterfactual Regret Minimization}
\author{Sina Baghal \\ \texttt{siinabaghal@gmail.com}}
\begin{document}
\maketitle
\begin{abstract}

Pasur is a fishing card game played over six rounds and is played similarly to games such as Cassino and Scopa, and Bastra \footnote{\href{https://en.wikipedia.org/wiki/Pasur_(card_game)}{Wikipedia: Pasur (card game)}}. This paper introduces a CUDA-accelerated computational framework for simulating Pasur, emphasizing efficient memory management.  We use our framework to compute near-Nash equilibria via Counterfactual Regret Minimization (CFR), a well-known algorithm for solving large imperfect-information games. 

\vspace{1mm}

Solving Pasur presents unique challenges due to its intricate rules and the large size of its game tree. We handle rule complexity using PyTorch CUDA tensors and to address the memory-intensive nature of the game, we decompose the game tree into two key components: (1) actual game states, and (2) inherited scores from previous rounds. We construct the \textit{Full Game Tree} by pairing card states with accumulated scores in the \textit{Unfolding Process}. This design reduces memory overhead by storing only essential strategy values and node connections. To further manage computational complexity, we apply a round-by-round backward training strategy, starting from the final round and recursively propagating average utilities to earlier stages. Our approach constructs the complete game tree, which on average consists of over $10^9$ nodes. We provide detailed implementation snippets to illustrate the structure and training logic of our framework and the CFR algorithm.

\vspace{1mm}

\vspace{1mm}

After computing a near-Nash equilibrium strategy, we train a tree-based model to predict these strategies for use during gameplay. We then estimate the fair value of each deck through large-scale self-play between equilibrium strategies by simulating, for instance, 10,000 games per matchup, executed in parallel using GPU acceleration. Our analysis reveals that the distribution of high-value cards heavily influences match outcomes, accounting for much of the variation in deck fairness. Finally, each trained model is lightweight, making it feasible to be used as a real-time AI agent for Pasur.

\vspace{1mm}
Similar frameworks can be extended to other reinforcement learning algorithms in settings where the action tree naturally decomposes into multiple rounds such as turn-based strategy games or sequential trading decisions in financial markets.
\end{abstract}



\section{Introduction and Background}

Imperfect-information games pose significant challenges due to hidden information and strategic uncertainty. Counterfactual Regret Minimization (CFR) has emerged as a powerful algorithmic framework for approximating Nash equilibria in such settings, notably achieving success in poker variants and other complex card games \cite{zinkevich2007regret, lanctot2009monte, brown2019deep, brown2019solving, xu2024minimizing}. By iteratively minimizing regret over possible strategies, CFR converges towards equilibrium policies without exhaustive search, making it an attractive method for large-scale sequential games.

In this work, we focus on \emph{Pasur}, a traditional fishing card game popular in Middle Eastern cultures, with a six-round structure and unique scoring rules. Each round, players receive new cards while aiming to capture valuable cards from the table through matching and summation mechanics. Unlike many extensively studied games, Pasur has not been rigorously solved, in part due to its combinatorial complexity and memory-intensive game tree.

In this paper, we present a CUDA-accelerated implementation of CFR in PyTorch tailored to Pasur, addressing its computational challenges by decomposing the game tree into game states and accumulated scores. Our approach computes the complete game tree, enabling near-Nash equilibrium calculations rather than relying on sampling methods. To further optimize GPU memory usage, we train the CFR models round-by-round in reverse order, propagating average utilities to ensure tractability. At any given time, only the tensors for the current round are kept on the GPU; all others are stored in CPU memory. Our carefully designed data structures made it possible to train and develop the entire CFR algorithm on a game tree with an average size of $10^9$, all within the constraints of a system with 32 GB of RAM and 10 GB of VRAM. After computing a near-Nash equilibrium strategy, we train a tree-based model to calculate the strategies for use during gameplay. These tree-model also facilites the estimation of deck fair values through self-play. 

This framework not only advances the strategic understanding of Pasur and fishing card games more broadly, but also enables the development of practical AI agents capable of playing at near-optimal levels. Moreover, this document details the construction and update procedures of every step in the framework through CodeSnippets, including the implementation of the CFR algorithm.

Several clarifications about the CodeSnippets are in order. The CodeSnippets aim to convey as much detail as possible; however, certain operations are omitted for clarity. These include routine tasks such as data type conversions and the use of \py{.clone()} to avoid in-place modifications, if necessary. Additionally, we omit the \py{torch.} prefix in all code expressions. For example, \py{torch.any(t\_act[:,1,:], dim=1)} is written as \py{any(t\_act[:,1,:], dim=1)}. It is also worth noting that not all lines of code are shown in the CodeSnippets. For complete implementations, we refer the reader to the GitHub repository\footnote{\href{https://sinabaghal.github.io/pasur/}{Paper Homepage}}. Finally, we use symbolic notation to simplify certain expressions. For instance, instead of \py{t\_1.repeat\_interleave(t\_2)}, we write \py{t\_1}$\otimes$\py{t\_2}. Similarly, \py{t\_1}$\otimes_1$\py{t\_2} denotes the dimension-specific form 
\[
    \py{t\_1.repeat\_interleave(t\_2, dim=1)}.
\]

\subsection{Extensive Games and CFR}
\input{notation}

\input{cfr}
\subsection{Pasur}

\input{pasur}

\section{PyTorch-Based Framework}
\input{pytorch}

\input{tree}
\input{fwork}
\bibliographystyle{plain}
\bibliography{bib} 
\input{appendix}
\end{document}

%% file: commands.tex
\usepackage{algorithm}
\usepackage{algpseudocode}
\usepackage{xcolor}
\usepackage{amsmath}
\usepackage{tikz}
\usepackage{pifont}
\usepackage{bm}
\usepackage{hyperref}
\usepackage{booktabs}
\usepackage{wrapfig}
\usepackage{algorithm}
\usepackage{algpseudocode}
\usepackage{caption}
\usepackage{amssymb}  
\usepackage{longtable}
\usepackage{booktabs}
\usepackage{graphicx} 
\input{cardscommands}
\usepackage{xcolor}
\usepackage{caption}
\usepackage{colortbl}
\usepackage{float} 
\definecolor{myrowcolor}{HTML}{CCFFCC}  
\newtheorem{theorem}{Theorem}
\usepackage{geometry}

\newcommand{\py}[1]{\texttt{#1}}
\usetikzlibrary{decorations.markings}
\usetikzlibrary{decorations.pathmorphing}
\tikzset{
  midarrow/.style={
    decoration={markings, mark=at position 0.5 with {\arrow{>}}},
    postaction={decorate}
  }
}

\newcommand{\drawboxwitharrows}[6]{%
    \begin{scope}[shift={(#5)}]
      \node[draw, fill=#6, minimum width=.5cm, minimum height=.5cm] (box) at (#1,0) {#4};
      
      \foreach \col/\x in {#2} {
        \draw[\col, thick, midarrow] 
          (#1+\x,0.7+\x/2) 
          to[out=270,in=90] 
          (box.north);
      }
    \end{scope}
}

\tikzset{
  midarrow/.style={
    decoration={markings, mark=at position 0.5 with {\arrow{>}}},
    postaction={decorate}
  }
}

\usetikzlibrary{patterns}

%% file: cardscommands.tex
\newcommand{\ThH}{\textcolor{red}{3\heartsuit}}  
\newcommand{\TwH}{\textcolor{red}{2\heartsuit}}  
\newcommand{\FoH}{\textcolor{red}{4\heartsuit}}   
\newcommand{\FiH}{\textcolor{red}{5\heartsuit}}   
\newcommand{\SiH}{\textcolor{red}{6\heartsuit}}    
\newcommand{\SeH}{\textcolor{red}{7\heartsuit}}  
\newcommand{\EH}{\textcolor{red}{8\heartsuit}}  
\newcommand{\NH}{\textcolor{red}{9\heartsuit}}   
\newcommand{\TeH}{\textcolor{red}{10\heartsuit}}   
\newcommand{\JH}{\textcolor{red}{J\heartsuit}}    
\newcommand{\QH}{\textcolor{red}{Q\heartsuit}}   
\newcommand{\KH}{\textcolor{red}{K\heartsuit}}    
\newcommand{\AH}{\textcolor{red}{A\heartsuit}}     
\floatname{algorithm}{CodeSnippet}
\usepackage{tikz}
\usetikzlibrary{patterns, decorations.pathmorphing, arrows.meta}

\usepackage{graphicx,scalerel}
\usepackage{adjustbox}
\usetikzlibrary{calc}
\newcommand{\ThD}{\textcolor{red}{3\diamondsuit}}  
\newcommand{\TwD}{\textcolor{red}{2\diamondsuit}}  
\newcommand{\FoD}{\textcolor{red}{4\diamondsuit}}   
\newcommand{\FiD}{\textcolor{red}{5\diamondsuit}}   
\newcommand{\SiD}{\textcolor{red}{6\diamondsuit}}    
\newcommand{\SeD}{\textcolor{red}{7\diamondsuit}}  
\newcommand{\ED}{\textcolor{red}{8\diamondsuit}}  
\newcommand{\ND}{\textcolor{red}{9\diamondsuit}}   
\newcommand{\TeD}{\textcolor{red}{10\diamondsuit}}   
\newcommand{\JD}{\textcolor{red}{J\diamondsuit}}    
\newcommand{\QD}{\textcolor{red}{Q\diamondsuit}}   
\newcommand{\KD}{\textcolor{red}{K\diamondsuit}}    
\newcommand{\AD}{\textcolor{red}{A\diamondsuit}}     

\newcommand{\ThC}{\textcolor{black}{3\clubsuit}}   
\newcommand{\TwC}{\textcolor{black}{2\clubsuit}}   
\newcommand{\FoC}{\textcolor{black}{4\clubsuit}}    
\newcommand{\FiC}{\textcolor{black}{5\clubsuit}}    
\newcommand{\SiC}{\textcolor{black}{6\clubsuit}}     
\newcommand{\SeC}{\textcolor{black}{7\clubsuit}}   
\newcommand{\EC}{\textcolor{black}{8\clubsuit}}   
\newcommand{\NC}{\textcolor{black}{9\clubsuit}}    
\newcommand{\TeC}{\textcolor{black}{10\clubsuit}}    
\newcommand{\JC}{\textcolor{black}{J\clubsuit}}    
\newcommand{\QC}{\textcolor{black}{Q\clubsuit}}   
\newcommand{\KC}{\textcolor{black}{K\clubsuit}}    
\newcommand{\AC}{\textcolor{black}{A\clubsuit}}     

\newcommand{\ThS}{\textcolor{black}{3\spadesuit}}   
\newcommand{\TwS}{\textcolor{black}{2\spadesuit}}   
\newcommand{\FoS}{\textcolor{black}{4\spadesuit}}    
\newcommand{\FiS}{\textcolor{black}{5\spadesuit}}    
\newcommand{\SiS}{\textcolor{black}{6\spadesuit}}     
\newcommand{\SeS}{\textcolor{black}{7\spadesuit}}   
\newcommand{\ES}{\textcolor{black}{8\spadesuit}}   
\newcommand{\NS}{\textcolor{black}{9\spadesuit}}    
\newcommand{\TeS}{\textcolor{black}{10\spadesuit}}    
\newcommand{\JS}{\textcolor{black}{J\spadesuit}}    
\newcommand{\QS}{\textcolor{black}{Q\spadesuit}}   
\newcommand{\KS}{\textcolor{black}{K\spadesuit}}    
\newcommand{\AS}{\textcolor{black}{A\spadesuit}}     

%% file: notation.tex
In \textit{Imperfect-Information Extensive-Form Games} there is a finite set of players ($\mathcal{P}$). A \textit{node} $h$ at time $T$ is defined by all the information revealed at a , particularly the actions taken by players $\mathcal{P}$. \textit{Terminal nodes}, denoted by $\mathcal{Z}$, are defined as those nodes where no further action is available. For each player $p \in \mathcal{P}$, there is a payoff function $u_p:\mathcal{Z}\to \mathbb{R}$. In this paper, we focus on the \textit{zero-sum} two-player setting \textit{i.e.,}  $\mathcal{P}=\{0,1\}$ and $u_0=-u_1$. 

Imperfect information is represented by \textit{information sets}
(infosets) for each player $p\in \mathcal{P}$. It is emphasized that at player $p$'s turn at infoset $I$, all nodes $h,h'\in I$ are identical from $p$'s perspective. In this situation, we say infoset $I$ belongs to $p$ and we denote the set of all such infosets by $\mathcal{I}_p$. Actions available at infoset $I$ are also denoted by $A(I)$. A \textit{strategy} $\sigma_p(I):A(I)\to \mathbb{R}^{\geq 0}$ is a distribution over actions in $A(I)$. The strategy of other players is denoted by $\sigma_{-p}$. We denote by $u_p(\sigma_p,\sigma_{-p})$ the expected payoff for $p$ if players' actions are governed by strategy profile $\sigma:=\{\sigma_p\}_{p\in \mathcal{P}}$. 

\textit{Reach probability} $\pi^{\sigma}(h)$ is defined as the probability of arriving at node $h$, if all players play according to $\sigma$. For $I\in \mathcal{I}_p$ and $h\in A(I)$, we denote by $\pi^{\sigma}_{-p}(h)$ the probability of arriving at $h$ in the event where $p$ chooses to reach $h$ and other players follow $\sigma$. Define $\pi^{\sigma}_p(I):=\sum_{h\in I} \pi^{\sigma}_p(h)$ and $\pi^{\sigma}_{-p}(I):=\sum_{h\in I} \pi^{\sigma}_{-p}(h)$. \textit{Counterfactual utility} at infoset $I\in \mathcal{I}_p$ is defined as 
\begin{equation}\label{eq:cfutl}
    u^{\sigma}(I) := \sum_{h\in I, h'\in \mathcal{Z}}  \pi^{\sigma}_{-p}(h)\pi^{\sigma}(h,h')u_p(h')
\end{equation}
Similarly, counterfactual utility for $a\in A(I)$, $u^{\sigma}(I,a)$ is defined as in \eqref{eq:cfutl}, except that $p$ chooses $a$ with probability 1 once it reaches $I$. Formally, if $h.a$ denotes the node wherein action $a$ is chosen at node $h$, then 
\begin{equation}
    u^{\sigma}_p(I,a) := \sum_{h\in I, h'\in \mathcal{Z}}  \pi^{\sigma}_{-p}(h)\pi^{\sigma}(h.a,h')u_p(h')
\end{equation}
Finally, in a two-player extensive game a \textit{Nash equilibrium} \cite{nash1950equilibrium} is a strategy profile $\boldsymbol{\sigma}^*$ for which the following holds
\[
u_p(\sigma^*_p,\sigma^*_{-p}) = \max_{\sigma'_p}u_p(\sigma'_p, \sigma^*_{-p}).
\]
In other words, $\sigma_p$ is the \textit{best response} to $\sigma_{-p}$ for each $p\in \mathcal{P}$. An $\epsilon$-Nash equilibrium (in a two-player game, for example) is also defined as 
\[
u_p(\sigma^*_p,\sigma^*_{-p}) +\epsilon \geq \max_{\sigma'_p}  u_p(\sigma'_p, \sigma^*_{-p}), \quad \forall p\in \{0,1\}.
\]
We are now ready to provide an overview of CFR next; for a complete discussion, see Zinkevich et al.\ (2007).

%% file: cfr.tex
CFR is an iterative algorithm that converges to a Nash equilibrium in any finite two-player zero-sum game with a theoretical convergence bound of \( O\left( \frac{1}{\sqrt{T}} \right) \). 
At the heart of CFR lies the concept of \textit{regret}. For a strategy profile $\sigma$, \textit{instantaneous regret} of playing $a$ vs. $\sigma$ at $I\in \mathcal{I}_p$ is denoted by 
\begin{equation}\label{eq:reg}
    r_p(I,a) := u_p^{\sigma}(I,a)-u_p^{\sigma}(I)
\end{equation}
In CFR, a \textit{regret matching} (RM) is performed at each iteration. According to RM, at iteration $T$, $\sigma^{T+1}_p(I)$  is determined using regrets \eqref{eq:reg} accumulated up to time $T$. Formally, 
\begin{equation}\label{eq:RM}
    \sigma^{T+1}_p(I) \propto R^T_+(I,a):= \left(\sum_{t=1}^T r^t(I,a)\right)_+
\end{equation}
$R^T_+(I,a)$ is called \textit{counterfactual regret} for infoset $I$ action $a$. Under update rule \eqref{eq:RM}, average strategy is then defined as follows
\begin{equation}\label{eq:avsigma}
\bar{\sigma}_p^T(I)(a) \propto \sum_{t=1}^{T} \pi_p^{\sigma^t}(I) \, \sigma_p^t(I)(a).
\end{equation}
The following two well-established results show that under RM \eqref{eq:RM}, the average strategy \eqref{eq:avsigma} converges to a Nash equilibrium in zero-sum two-player games. 
\begin{theorem}\label{thm:avst}
    In a zero-sum game, average strategy \eqref{eq:avsigma} is a $2\epsilon$-Nash equilibrium provided that the total regret which is defined below is less than $\epsilon$ for $p\in \{0,1\}$.
    \[R^T_p:=\max_{\sigma'_p}\frac{1}{T}\sum_{t=1}^T\left(u_p(\sigma'_p,\sigma_{-p}^t)-u_p(\sigma_p^t,\sigma^t_{-p})\right)\]
\end{theorem}
The following theorem states that $R_p^T \to 0$ as $T\to +\infty$ under RM \eqref{eq:RM}.
\begin{theorem}[Theorem 3 \& 4, \cite{zinkevich2007regret}]\label{thm:rm}
    Under RM \eqref{eq:RM}, the following bound holds 
    \[
    R_p^T = \mathcal{O}\left(\frac{1}{\sqrt{T}}\right).
    \]
    Moreover, total regret is lower bounded by counterfactual regrets defined in \eqref{eq:RM}.
\end{theorem}

There are different variants of CFR where the difference is in the way it updates counterfactual regrets. For example, in CFR$^+$~\cite{tammelin2014solving}, regrets, initialized at zero, are updated according to the following rule:
\[
R^{+,t}(I, a) = \left( R^{+,t-1}(I, a) + u^{\sigma^t}(I, a) - u^{\sigma^t}(I) \right)_+
\]
Another example that is also used in this paper is the Discounted CFR (DCFR) \cite{brown2019solving}, where prior
iterations when determining both regrets and the average strategy are discounted. The update rule for DCFR is as follows:

\begin{equation}\label{eq:DFCR_REG}
    R_p^t(I,a) = R_p^{t-1}(I,a) \cdot d_p^{t-1}(I,a) + r_p^t(I,a)
\end{equation}
where
\begin{equation}
d_p^t(I,a) = 
\begin{cases}
\displaystyle \frac{t^\alpha}{t^\alpha + 1} & \text{if } R_p^t(I,a) > 0 \\
\displaystyle \frac{t^\beta}{t^\beta + 1} & \text{otherwise}
\end{cases}    
\end{equation}
Moreover, the average strategies are updated according to the following rule:
\begin{equation}\label{eq:DFCR_AVG}    
    \bar{\sigma}_p^t(I)(a) \propto \left(1-\tfrac{1}{t}\right)^{\gamma}\cdot\bar{\sigma}_p^{t-1}(I)(a)+\pi_p^{\sigma^t}(I)\sigma^t_p(I)(a)
\end{equation}
    

%% file: pasur.tex
Pasur is a traditional card game played with a standard 52-card deck (excluding jokers), and it supports 2 to 4 players. In this paper, we focus on the two-player variant, where the players are referred to as \textbf{Alex} and \textbf{Bob}, along with a \textbf{Dealer} who manages the game.

The game begins with four cards placed face-up on the table to form the initial pool. This pool must not contain any Jacks. If a Jack appears among the initial four cards, it is returned to the deck and replaced. If multiple Jacks are dealt, or if a replacement card is also a Jack, the dealer reshuffles and redeals.

Once the pool is valid and face-up, the dealer deals four cards to each player, starting with the player on their left (assumed to be Alex). Players then take turns beginning with Alex. On each turn, a player must play one card from their hand. The played card will either: Be added to the pool of face-up cards, or Capture one or more cards from the pool, following rules described in Table \ref{tab:cap_rules}. If a capture is possible, the player \emph{must} capture; they cannot simply add a card to the pool. Captured cards are retained and used to calculate each player’s score at the end. 

\begin{table}[h!]
\centering
\scriptsize
\begin{tabular}{|c|p{10cm}|}
\hline
\multicolumn{1}{|c|}{\textbf{Card Type}} & \multicolumn{1}{c|}{\textbf{Capture Rule}} \\
\hline
Numeric & One or more numeric cards from the pool if their total sum equals 11 \\
\hline
Jack & All cards in the pool, except Kings and Queens (can also capture other Jacks) \\
\hline
Queen & A single Queen \\
\hline
King & A single King \\
\hline
\end{tabular}
\caption{Capture rules for each card type}
\label{tab:cap_rules}
\end{table}

A \textbf{Sur} occurs when a player captures all the cards from the pool in a single move. There are two important exceptions: (1) a Sur cannot be made using a Jack, and (2) Surs are not permitted during the final round of play. Table~\ref{tab:score} outlines Pasur's scoring system, and Figure~\ref{fig:game_0} provides a gameplay example.
\vspace{10 mm}
\begin{table}[h!]
\centering
\begin{tabular}{|c|c|}
\hline
\textbf{Rule} & \textbf{Points} \\
\hline
Most Clubs & 7 \\
Each Jack & 1 \\
Each Ace & 1 \\
Each \;Sur & 5 \\
10{\color{red}\ding{117}} & 3 \\
2{\ding{168}} & 2 \\
\hline
\end{tabular}
\caption{Pasur Scoring System}
\label{tab:score}
\end{table}

\input{figs/game}

%% file: figs/game.tex
\begin{figure}[!p]  
\caption*{Pasur: A Step-by-Step Gameplay Snapshot}
\begin{center}
\resizebox{1\textwidth}{!}{%
\begin{tabular}{|>{\centering\arraybackslash}p{0.1\textwidth}|
                >{\centering\arraybackslash}p{0.2\textwidth}|
                >{\centering\arraybackslash}p{0.2\textwidth}|
                >{\centering\arraybackslash}p{0.7\textwidth}|
                >{\centering\arraybackslash}p{0.05\textwidth}|
                 >{\centering\arraybackslash}p{0.7\textwidth}|
                >{\centering\arraybackslash}p{0.05\textwidth}|
                >{\centering\arraybackslash}p{0.05\textwidth}|
                 >{\centering\arraybackslash}p{0.05\textwidth}|
                  >{\centering\arraybackslash}p{0.05\textwidth}|
                   >{\centering\arraybackslash}p{0.05\textwidth}|
                   >{\centering\arraybackslash}p{0.05\textwidth}|
                   >{\centering\arraybackslash}p{0.05\textwidth}|
                   >{\centering\arraybackslash}p{0.05\textwidth}|>{\centering\arraybackslash}p{0.05\textwidth}|}
\hline
\textbf{Stage}& \textbf{Alex} & \textbf{Bob}& \textbf{Pool} & \textbf{Lay} & \textbf{Pick} & \textbf{Acl} &  \textbf{Bcl} & \textbf{Apt} &  \textbf{Bpt} & \textbf{Asr} & \textbf{Bsr} & \textbf{$\Delta$} & \textbf{L} & \textbf{CL}\\
\hline
\rowcolor{myrowcolor}

 0\_0\_0 &$\FoC\;\FoD\;\SeD\;\QC$&$\ThD\;\ThH\;\FiC\;\KS$&$\AC\;\AS\;\ND\;\KD$&$\FoD$&&0 &0 &0 &0 &0 &0 &0 &0&-\\ \hline

 0\_0\_1 &$\FoC\;\SeD\;\QC$&$\ThD\;\ThH\;\FiC\;\KS$&$\AC\;\AS\;\FoD\;\ND\;\KD$&$\KS$&$\KD$&0 &0 &0 &0 &0 &0 &0 &B&-\\ \hline
 \rowcolor{myrowcolor}

 0\_1\_0 &$\FoC\;\SeD\;\QC$&$\ThD\;\ThH\;\FiC$&$\AC\;\AS\;\FoD\;\ND$&$\QC$&&0 &0 &0 &0 &0 &0 &0 &B&-\\ \hline

 0\_1\_1 &$\FoC\;\SeD$&$\ThD\;\ThH\;\FiC$&$\AC\;\AS\;\FoD\;\ND\;\QC$&$\FiC$&$\AC\;\AS\;\FoD$&0 &2 &0 &2 &0 &0 &0 &B&-\\ \hline
 \rowcolor{myrowcolor}

 0\_2\_0 &$\FoC\;\SeD$&$\ThD\;\ThH$&$\ND\;\QC$&$\FoC$&&0 &2 &0 &2 &0 &0 &0 &B&-\\ \hline

 0\_2\_1 &$\SeD$&$\ThD\;\ThH$&$\FoC\;\ND\;\QC$&$\ThD$&&0 &2 &0 &2 &0 &0 &0 &B&-\\ \hline
 \rowcolor{myrowcolor}

 0\_3\_0 &$\SeD$&$\ThH$&$\ThD\;\FoC\;\ND\;\QC$&$\SeD$&$\FoC$&1 &2 &0 &2 &0 &0 &0 &A&-\\ \hline

 0\_3\_1 &&$\ThH$&$\ThD\;\ND\;\QC$&$\ThH$&&1 &2 &0 &2 &0 &0 &0 &A&-\\ \hline
 \rowcolor{myrowcolor}
\hline

 1\_0\_0 &$\SiC\;\SiD\;\NH\;\JC$&$\SiH\;\SeC\;\JD\;\KH$&$\ThD\;\ThH\;\ND\;\QC$&$\JC$&$\ThD\;\ThH\;\ND$&2 &2 &1 &0 &0 &0 &-2 &A&-\\ \hline

 1\_0\_1 &$\SiC\;\SiD\;\NH$&$\SiH\;\SeC\;\JD\;\KH$&$\QC$&$\JD$&&2 &2 &1 &0 &0 &0 &-2 &A&-\\ \hline
 \rowcolor{myrowcolor}

 1\_1\_0 &$\SiC\;\SiD\;\NH$&$\SiH\;\SeC\;\KH$&$\JD\;\QC$&$\SiC$&&2 &2 &1 &0 &0 &0 &-2 &A&-\\ \hline

 1\_1\_1 &$\SiD\;\NH$&$\SiH\;\SeC\;\KH$&$\SiC\;\JD\;\QC$&$\KH$&&2 &2 &1 &0 &0 &0 &-2 &A&-\\ \hline
 \rowcolor{myrowcolor}

 1\_2\_0 &$\SiD\;\NH$&$\SiH\;\SeC$&$\SiC\;\JD\;\QC\;\KH$&$\SiD$&&2 &2 &1 &0 &0 &0 &-2 &A&-\\ \hline

 1\_2\_1 &$\NH$&$\SiH\;\SeC$&$\SiC\;\SiD\;\JD\;\QC\;\KH$&$\SeC$&&2 &2 &1 &0 &0 &0 &-2 &A&-\\ \hline
 \rowcolor{myrowcolor}

 1\_3\_0 &$\NH$&$\SiH$&$\SiC\;\SiD\;\SeC\;\JD\;\QC\;\KH$&$\NH$&&2 &2 &1 &0 &0 &0 &-2 &A&-\\ \hline

 1\_3\_1 &&$\SiH$&$\SiC\;\SiD\;\SeC\;\NH\;\JD\;\QC\;\KH$&$\SiH$&&2 &2 &1 &0 &0 &0 &-2 &A&-\\ \hline
 \rowcolor{myrowcolor}
\hline

 2\_0\_0 &$\FiD\;\NC\;\TeH\;\KC$&$\AD\;\SiS\;\EC\;\ES$&$\SiC\;\SiD\;\SiH\;\SeC\;\NH\;\JD\;\QC\;\KH$&$\TeH$&&2 &2 &0 &0 &0 &0 &-1 &0&-\\ \hline

 2\_0\_1 &$\FiD\;\NC\;\KC$&$\AD\;\SiS\;\EC\;\ES$&$\SiC\;\SiD\;\SiH\;\SeC\;\NH\;\TeH\;\JD\;\QC\;\KH$&$\EC$&&2 &2 &0 &0 &0 &0 &-1 &0&-\\ \hline
 \rowcolor{myrowcolor}

 2\_1\_0 &$\FiD\;\NC\;\KC$&$\AD\;\SiS\;\ES$&$\SiC\;\SiD\;\SiH\;\SeC\;\EC\;\NH\;\TeH\;\JD\;\QC\;\KH$&$\KC$&$\KH$&3 &2 &0 &0 &0 &0 &-1 &A&-\\ \hline

 2\_1\_1 &$\FiD\;\NC$&$\AD\;\SiS\;\ES$&$\SiC\;\SiD\;\SiH\;\SeC\;\EC\;\NH\;\TeH\;\JD\;\QC$&$\AD$&$\TeH$&3 &2 &0 &1 &0 &0 &-1 &B&-\\ \hline
 \rowcolor{myrowcolor}

 2\_2\_0 &$\FiD\;\NC$&$\SiS\;\ES$&$\SiC\;\SiD\;\SiH\;\SeC\;\EC\;\NH\;\JD\;\QC$&$\FiD$&$\SiH$&3 &2 &0 &1 &0 &0 &-1 &A&-\\ \hline

 2\_2\_1 &$\NC$&$\SiS\;\ES$&$\SiC\;\SiD\;\SeC\;\EC\;\NH\;\JD\;\QC$&$\ES$&&3 &2 &0 &1 &0 &0 &-1 &A&-\\ \hline
 \rowcolor{myrowcolor}

 2\_3\_0 &$\NC$&$\SiS$&$\SiC\;\SiD\;\SeC\;\EC\;\ES\;\NH\;\JD\;\QC$&$\NC$&&3 &2 &0 &1 &0 &0 &-1 &A&-\\ \hline

 2\_3\_1 &&$\SiS$&$\SiC\;\SiD\;\SeC\;\EC\;\ES\;\NC\;\NH\;\JD\;\QC$&$\SiS$&&3 &2 &0 &1 &0 &0 &-1 &A&-\\ \hline
 \rowcolor{myrowcolor}
\hline

 3\_0\_0 &$\TwC\;\ThC\;\JS\;\QS$&$\SeS\;\ED\;\EH\;\QH$&$\SiC\;\SiD\;\SiS\;\SeC\;\EC\;\ES\;\NC\;\NH\;\JD\;\QC$&$\TwC$&$\NC$&5 &2 &2 &0 &0 &0 &-2 &A&-\\ \hline

 3\_0\_1 &$\ThC\;\JS\;\QS$&$\SeS\;\ED\;\EH\;\QH$&$\SiC\;\SiD\;\SiS\;\SeC\;\EC\;\ES\;\NH\;\JD\;\QC$&$\ED$&&5 &2 &2 &0 &0 &0 &-2 &A&-\\ \hline
 \rowcolor{myrowcolor}

 3\_1\_0 &$\ThC\;\JS\;\QS$&$\SeS\;\EH\;\QH$&$\SiC\;\SiD\;\SiS\;\SeC\;\EC\;\ED\;\ES\;\NH\;\JD\;\QC$&$\JS$&$\SiC\;\SiD\;\SiS\;\SeC\;\EC\;\ED\;\ES\;\NH\;\JD$&8 &2 &4 &0 &0 &0 &-2 &A&-\\ \hline

 3\_1\_1 &$\ThC\;\QS$&$\SeS\;\EH\;\QH$&$\QC$&$\SeS$&&8 &2 &4 &0 &0 &0 &-2 &A&-\\ \hline
 \rowcolor{myrowcolor}

 3\_2\_0 &$\ThC\;\QS$&$\EH\;\QH$&$\SeS\;\QC$&$\ThC$&&8 &2 &4 &0 &0 &0 &-2 &A&-\\ \hline

 3\_2\_1 &$\QS$&$\EH\;\QH$&$\ThC\;\SeS\;\QC$&$\QH$&$\QC$&8 &3 &4 &0 &0 &0 &-2 &B&-\\ \hline
 \rowcolor{myrowcolor}

 3\_3\_0 &$\QS$&$\EH$&$\ThC\;\SeS$&$\QS$&&8 &3 &4 &0 &0 &0 &-2 &B&-\\ \hline

 3\_3\_1 &&$\EH$&$\ThC\;\SeS\;\QS$&$\EH$&$\ThC$&8 &4 &4 &0 &0 &0 &-2 &B&-\\ \hline
 \rowcolor{myrowcolor}
\hline

 4\_0\_0 &$\FoH\;\FoS\;\SeH\;\JH$&$\TwD\;\FiH\;\NS\;\TeS$&$\SeS\;\QS$&$\FoH$&$\SeS$&0 &0 &0 &0 &0 &0 &2 &A&A\\ \hline

 4\_0\_1 &$\FoS\;\SeH\;\JH$&$\TwD\;\FiH\;\NS\;\TeS$&$\QS$&$\FiH$&&0 &0 &0 &0 &0 &0 &2 &A&A\\ \hline
 \rowcolor{myrowcolor}

 4\_1\_0 &$\FoS\;\SeH\;\JH$&$\TwD\;\NS\;\TeS$&$\FiH\;\QS$&$\SeH$&&0 &0 &0 &0 &0 &0 &2 &A&A\\ \hline

 4\_1\_1 &$\FoS\;\JH$&$\TwD\;\NS\;\TeS$&$\FiH\;\SeH\;\QS$&$\TeS$&&0 &0 &0 &0 &0 &0 &2 &A&A\\ \hline
 \rowcolor{myrowcolor}

 4\_2\_0 &$\FoS\;\JH$&$\TwD\;\NS$&$\FiH\;\SeH\;\TeS\;\QS$&$\JH$&$\FiH\;\SeH\;\TeS$&0 &0 &1 &0 &0 &0 &2 &A&A\\ \hline

 4\_2\_1 &$\FoS$&$\TwD\;\NS$&$\QS$&$\TwD$&&0 &0 &1 &0 &0 &0 &2 &A&A\\ \hline
 \rowcolor{myrowcolor}

 4\_3\_0 &$\FoS$&$\NS$&$\TwD\;\QS$&$\FoS$&&0 &0 &1 &0 &0 &0 &2 &A&A\\ \hline

 4\_3\_1 &&$\NS$&$\TwD\;\FoS\;\QS$&$\NS$&$\TwD$&0 &0 &1 &0 &0 &0 &2 &B&A\\ \hline
 \rowcolor{myrowcolor}
\hline

 5\_0\_0 &$\AH\;\ThS\;\TeC\;\QD$&$\TwH\;\TwS\;\FiS\;\TeD$&$\FoS\;\QS$&$\AH$&&0 &0 &0 &0 &0 &0 &3 &0&A\\ \hline

 5\_0\_1 &$\ThS\;\TeC\;\QD$&$\TwH\;\TwS\;\FiS\;\TeD$&$\AH\;\FoS\;\QS$&$\FiS$&&0 &0 &0 &0 &0 &0 &3 &0&A\\ \hline
 \rowcolor{myrowcolor}

 5\_1\_0 &$\ThS\;\TeC\;\QD$&$\TwH\;\TwS\;\TeD$&$\AH\;\FoS\;\FiS\;\QS$&$\TeC$&$\AH$&1 &0 &1 &0 &0 &0 &3 &A&A\\ \hline

 5\_1\_1 &$\ThS\;\QD$&$\TwH\;\TwS\;\TeD$&$\FoS\;\FiS\;\QS$&$\TeD$&&1 &0 &1 &0 &0 &0 &3 &A&A\\ \hline
 \rowcolor{myrowcolor}

 5\_2\_0 &$\ThS\;\QD$&$\TwH\;\TwS$&$\FoS\;\FiS\;\TeD\;\QS$&$\QD$&$\QS$&1 &0 &1 &0 &0 &0 &3 &A&A\\ \hline

 5\_2\_1 &$\ThS$&$\TwH\;\TwS$&$\FoS\;\FiS\;\TeD$&$\TwS$&$\FoS\;\FiS$&1 &0 &1 &0 &0 &0 &3 &B&A\\ \hline
 \rowcolor{myrowcolor}

 5\_3\_0 &$\ThS$&$\TwH$&$\TeD$&$\ThS$&&1 &0 &1 &0 &0 &0 &3 &B&A\\ \hline

 5\_3\_1 &&$\TwH$&$\ThS\;\TeD$&$\TwH$&&1 &0 &1 &0 &0 &0 &3 &B&A\\ \hline
\hline 

CleanUp &&&&&&1 &0 &1 &3 &0 &0 &2 &B&A\\ \hline
\end{tabular}
}
\end{center}
\caption{Columns \texttt{Acl}, \texttt{Apt}, and \texttt{Asr} represent the number of clubs, points, and surs collected or earned by Alex so far. \texttt{Bcl}, \texttt{Bpt}, and \texttt{Bsr} are defined similarly for Bob.  Once, at the end of any round, \texttt{Acl} exceeds 7, both \texttt{Acl} and \texttt{Bcl} reset to zero, and the column \texttt{CL} indicates which player collected at least 7 clubs. According to Pasur rules, this player earns 7 points. Collecting more than 7 clubs yields no additional points unless the card is also a point card (i.e., \textit{e.g.}, $\AC$ or 2\ding{168} ). The column $\Delta$ shows the cumulative point difference up to the end of the previous round. $\Delta$ updates at the end of each round to reflect the points earned in that round. Finally, in the \textit{CleanUp} phase, the player who made the last pick (as shown in column \texttt{L}) collects all remaining cards from the pool. If any point cards are present, the corresponding point columns will be updated. If there are club cards in the CleanUp phase and neither player has yet reached 7 clubs, then the remaining clubs in the pool determine who earns the 7-club bonus. In such situations, the identity of the last player to pick becomes critical. 
\\
\\
In the displayed game above, Alex earns the 7-club bonus. Bob gains 3 points from collecting $\TeD$, but he is trailing by 3 points from the rounds preceding the last. Additionally, Alex earns 1 point in the final round from capturing the $\AH$ card. This results in a final score with Alex leading by 8 points. See Table~\ref{tab:score} for the Pasur scoring system.
}
\label{fig:game_0}
\end{figure}

\newpage

%% file: pytorch.tex
In this section, we present our computational framework for implementing for simulating Pasur using PyTorch.  The game consists of 6 rounds, and in each round, both players take 4 turns, giving rise to a game tree of depth 48. We begin by describing how a set of cards is represented as a PyTorch tensor. As illustrated in Figure~\ref{fig:deck}, each card is mapped to a unique tensor index according to a fixed, natural order. The interpretation of the tensor values at these positions varies depending on the specific tensor in which they appear: it may indicate card ownership (e.g., who holds the card) or an action involving that card (e.g., laid or picked), depending on whether the tensor encodes game state, actions, or other game-related structures. This indexing convention serves as the foundation for constructing all tensors listed in Table~\ref{table:tensors}.

\begin{figure}[h]
    \centering
\begin{tikzpicture}
    \def\n{1} 
    \def\m{20} 
    \def\cellsize{0.7} 

    \foreach \i in {0,...,\n} {
        \draw[lightgray] (0,\i*\cellsize) -- (\m*\cellsize,\i*\cellsize); 
    }

    Place dots in the middle row (row 3, columns 2, 3, and 4)
    \foreach \j in {0,1,2} {
        \fill[black] ({(1-\j)/2+\m*\cellsize/2}, 0.5*\cellsize) circle (0.03); 
    }

    \foreach \j in {0,...,\m} {
        \ifnum\j<7 
            \draw[lightgray] (\j*\cellsize,0) -- (\j*\cellsize,\n*\cellsize); 
        \else
            \ifnum\j>13 
                \draw[lightgray] (\j*\cellsize,0) -- (\j*\cellsize,\n*\cellsize); 
            \fi
        \fi
        }

    \draw[black] (0,0) rectangle (\m*\cellsize, \n*\cellsize);



    \node at (0.5*\cellsize, \n*\cellsize-0.4*\cellsize) {\scriptsize $\AC$};
    \node at (1.5*\cellsize, \n*\cellsize-0.4*\cellsize) {\scriptsize $\AD$};
    \node at (2.5*\cellsize, \n*\cellsize-0.4*\cellsize) {\scriptsize $\AH$};
    \node at (3.5*\cellsize, \n*\cellsize-0.4*\cellsize) {\scriptsize $\AS$};
    \node at (4.5*\cellsize, \n*\cellsize-0.4*\cellsize) {\scriptsize $\TwC$};
    \node at (5.5*\cellsize, \n*\cellsize-0.4*\cellsize) {\scriptsize $\TwD$};

    \node at (14.5*\cellsize, \n*\cellsize-0.4*\cellsize) {\scriptsize $\QH$};
    \node at (15.5*\cellsize, \n*\cellsize-0.4*\cellsize) {\scriptsize $\QS$};
    \node at (16.5*\cellsize, \n*\cellsize-0.4*\cellsize) {\scriptsize $\KC$};
    \node at (17.5*\cellsize, \n*\cellsize-0.4*\cellsize) {\scriptsize $\KD$};
    \node at (18.5*\cellsize, \n*\cellsize-0.4*\cellsize) {\scriptsize $\KH$};
    \node at (19.5*\cellsize, \n*\cellsize-0.4*\cellsize) {\scriptsize $\KS$};

\end{tikzpicture}
\caption{Mapping Cards to Indices}
\label{fig:deck}
\end{figure}

To manage memory efficiently during game tree generation, we maintain two boolean tensors that track card availability throughout the game. Let \py{m} denote the number of \emph{in-play cards}—those currently in play, either held by a player or present in the pool. For instance, in the first round, \py{m = 12}, since each player is dealt four cards and four cards are placed in the pool. At the end of each round, we update the set of in-play cards: we remove any cards that have been picked across all nodes and add new cards that are about to be dealt. 

This information is captured using the \py{t\_inp} tensor. This binary tensor marks which cards are currently involved in the game—either held by players or present in the pool—and serves as the basis for constructing and updating all relevant tensors during game tree generation. It is emphasized that as a result of this, the shapes of certain tensors—particularly those that represent actions or action history, such as \py{t\_act} and \py{t\_gme}—may vary across rounds to reflect the current number of in-play cards. To illustrate this point, consider the initial configuration of the game shown in Table~\ref{table:deck9}, with the corresponding \py{t\_inp} tensor depicted in Figure~\ref{fig:tinp_ex}.

\begin{table}[h!]
\centering
\caption{Initial setup for a single game instance}
\label{table:deck9}
\begin{tabular}{lcccc}
\toprule
\textbf{Alex} & $\FoC$ & $\FoD$ & $\SeD$ & $\QC$ \\
\textbf{Bob} & $\ThD$ & $\ThH$ & $\FiC$ & $\KS$ \\
\textbf{Pool} & $\AC$ & $\AS$ & $\ND$ & $\KD$ \\
\bottomrule
\end{tabular}
\end{table}
\begin{figure}[h]
    \centering
\begin{tikzpicture}
    \def\n{1} 
    \def\m{25} 
    \def\cellsize{0.6} 

    \foreach \i in {0,...,\n} {
        \draw[lightgray] (0,\i*\cellsize) -- (\m*\cellsize,\i*\cellsize); 
    }

    \foreach \j in {0,1,2} {
        \fill[black] ({(1-\j)/2+\m*\cellsize/2+0.5}, 0.5*\cellsize) circle (0.03); 
    }

    \foreach \j in {0,...,\m} {
        \ifnum\j<12 
            \draw[lightgray] (\j*\cellsize,0) -- (\j*\cellsize,\n*\cellsize); 
        \else
            \ifnum\j>15 
                \draw[lightgray] (\j*\cellsize,0) -- (\j*\cellsize,\n*\cellsize); 
            \fi
        \fi
        }

    \draw[black] (0,0) rectangle (\m*\cellsize, \n*\cellsize);



    \node at (0.5*\cellsize, \n*\cellsize-0.4*\cellsize) {\tiny \py{True}};
    \node at (1.5*\cellsize, \n*\cellsize-0.4*\cellsize) {\tiny \py{False}};
    \node at (2.5*\cellsize, \n*\cellsize-0.4*\cellsize) {\tiny \py{False}};
    \node at (3.5*\cellsize, \n*\cellsize-0.4*\cellsize) {\tiny \py{True}};
    \node at (4.5*\cellsize, \n*\cellsize-0.4*\cellsize) {\tiny \py{False}};
    \node at (5.5*\cellsize, \n*\cellsize-0.4*\cellsize) {\tiny \py{False}};
    \node at (6.5*\cellsize, \n*\cellsize-0.4*\cellsize) {\tiny \py{False}};
    \node at (7.5*\cellsize, \n*\cellsize-0.4*\cellsize) {\tiny \py{False}};
    \node at (8.5*\cellsize, \n*\cellsize-0.4*\cellsize) {\tiny \py{False}};
    \node at (9.5*\cellsize, \n*\cellsize-0.4*\cellsize) {\tiny \py{True}};
    \node at (10.5*\cellsize, \n*\cellsize-0.4*\cellsize) {\tiny\py{True}};
    
    \node at (0.5*\cellsize, \n*\cellsize-1.4*\cellsize) {\tiny $\AC$};
    \node at (1.5*\cellsize, \n*\cellsize-1.4*\cellsize) {\tiny $\AD$};
    \node at (2.5*\cellsize, \n*\cellsize-1.4*\cellsize) {\tiny $\AH$};
    \node at (3.5*\cellsize, \n*\cellsize-1.4*\cellsize) {\tiny $\AS$};
    \node at (4.5*\cellsize, \n*\cellsize-1.4*\cellsize) {\tiny $\TwC$};
    \node at (5.5*\cellsize, \n*\cellsize-1.4*\cellsize) {\tiny $\TwD$};
    \node at (6.5*\cellsize, \n*\cellsize-1.4*\cellsize) {\tiny $\TwH$};
    \node at (7.5*\cellsize, \n*\cellsize-1.4*\cellsize) {\tiny $\TwS$};
    \node at (8.5*\cellsize, \n*\cellsize-1.4*\cellsize) {\tiny $\ThC$};
    \node at (9.5*\cellsize, \n*\cellsize-1.4*\cellsize) {\tiny $\ThD$};
    \node at (10.5*\cellsize,\n*\cellsize-1.4*\cellsize) {\tiny $\ThH$};

    \node at (16.5*\cellsize, \n*\cellsize-0.4*\cellsize) {\tiny \py{False}};
    \node at (17.5*\cellsize, \n*\cellsize-0.4*\cellsize) {\tiny \py{True}};
    \node at (18.5*\cellsize, \n*\cellsize-0.4*\cellsize) {\tiny \py{False}};
    \node at (19.5*\cellsize, \n*\cellsize-0.4*\cellsize) {\tiny \py{False}};
    \node at (20.5*\cellsize, \n*\cellsize-0.4*\cellsize) {\tiny \py{False}};
    \node at (21.5*\cellsize, \n*\cellsize-0.4*\cellsize) {\tiny \py{False}};
    \node at (22.5*\cellsize, \n*\cellsize-0.4*\cellsize) {\tiny \py{True}};
    \node at (23.5*\cellsize, \n*\cellsize-0.4*\cellsize) {\tiny \py{False}};
    \node at (24.5*\cellsize, \n*\cellsize-0.4*\cellsize) {\tiny \py{True}};
    
    \node at (16.4*\cellsize, \n*\cellsize-1.4*\cellsize) {\tiny $\JS$};
    \node at (17.4*\cellsize, \n*\cellsize-1.4*\cellsize) {\tiny $\QC$};
    \node at (18.4*\cellsize, \n*\cellsize-1.4*\cellsize) {\tiny $\QD$};
    \node at (19.4*\cellsize, \n*\cellsize-1.4*\cellsize) {\tiny $\QH$};
    \node at (20.4*\cellsize, \n*\cellsize-1.4*\cellsize) {\tiny $\QS$};
    \node at (21.4*\cellsize, \n*\cellsize-1.4*\cellsize) {\tiny $\KC$};
    \node at (22.4*\cellsize, \n*\cellsize-1.4*\cellsize) {\tiny $\KD$};
    \node at (23.4*\cellsize, \n*\cellsize-1.4*\cellsize) {\tiny $\KH$};
    \node at (24.4*\cellsize, \n*\cellsize-1.4*\cellsize) {\tiny $\KS$};

\end{tikzpicture}
\caption{\py{t\_inp} tensor at the beginning of the game for initial setup in Table~\ref{table:deck9}}
\label{fig:tinp_ex}
\end{figure}

We next provide an overview how the game tree is generated and recorded using tensor operations. During the tree construction process, we distinguish between the \emph{card state}—including the cards held by Alex, the cards held by Bob, and the cards in the pool—and the \emph{score information}, which is updated independently. A given card state may correspond to multiple incoming edges, each representing a different score inherited from earlier rounds. Figure~\ref{fig:gametreegeneration} illustrates this structural design. Alongside the \emph{Game Tree} (GT) illustrated in Figure~\ref{fig:gametreegeneration}, we construct a \emph{Full Game Tree} (FGT)  via an \emph{unfolding process} (Figure \ref{fig:Unfolding Process}), which systematically expands the tree by combining each card state with all compatible incoming scores.

\begin{center}
\textit{To ensure memory efficiency, the only parameters stored for FGT are the strategy values at each node and the edges linking nodes between successive layers.}
\end{center}

Thus, each node in FGT is represented as a pair: a GT node and its corresponding incoming score. We explicitly maintain the edge structure between FGT nodes, which is crucial for updating strategies during the CFR iterations. Figure~\ref{fig:Unfolding Process} provides a visual representation of the unfolding process. Further details on this mechanism and the edge-tracking procedure are discussed in the subsections that follow.

\begin{figure}
\centering
\scalebox{0.5}{
    \begin{tikzpicture}

        \node[rotate=0] at (-2, -4) {\Large Game Tree (GT)}; 
        \node[rotate=0] at (11, -4) {\Large Full Game Tree (FGT)}; 
 \draw[->, very thick, decorate, decoration={snake, amplitude=2mm, segment length=5mm}] (2.5cm, -9cm) -- (4.5cm, -9cm);
        \node[rotate=0] at (3.5, -8) {\Large Unfold}; 
        
      \drawboxwitharrows{0}{blue!50/0.4, red/0.05, green/-0.2}{dummy}{\py{0}}  {-2.5cm,  -8cm}{none}
      \drawboxwitharrows{0}{blue!50/0.4, orange/0.05}{dummy}{\py{1}}  {-1.5cm,   -8cm}{none}

      \drawboxwitharrows{0}{blue!50/0.4, red/0.05, green/-0.2}{dummy}                                       {\py{0}}    {-4cm,  -10cm}{none}
      \drawboxwitharrows{0}{blue!50/0.4, red/0.05, green/-0.2}{dummy}                                       {\py{1}}    {-3cm,  -10cm}{none}
      \drawboxwitharrows{0}{blue!50/0.4, orange/0.05}{dummy}                                       {\py{2}}    {-2cm,  -10cm}{none}
      \drawboxwitharrows{0}{blue!50/0.4, orange/0.05}{dummy}                                       {\py{3}}    {-1cm,  -10cm}{none}
      \drawboxwitharrows{0}{blue!50/0.4, orange/0.05}{dummy}                                       {\py{4}}    {0cm,   -10cm}{none}
       \draw (-4.5cm,    -10.5cm) -- (-2.5cm, -10.5cm);
        \draw (-2.2cm, -10.5cm) -- (0.4cm, -10.5cm);
         \drawboxwitharrows{0}{}{dummy}{\py{0}}  {9cm,   -7cm}{green}
         \drawboxwitharrows{0}{}{dummy}{\py{0}}  {10cm,  -7cm}{red}
         \drawboxwitharrows{0}{}{dummy}{\py{0}}  {11cm,  -7cm}{blue!50}
         \drawboxwitharrows{0}{}{dummy}{\py{1}}  {12cm,  -7cm}{orange}
         \drawboxwitharrows{0}{}{dummy}{\py{1}}  {13cm,  -7cm}{blue!50}

         \drawboxwitharrows{0}{}{dummy}{\py{0}}  {6cm,   -11cm}{green}
         \drawboxwitharrows{0}{}{dummy}{\py{0}}  {7cm,   -11cm}{red}
         \drawboxwitharrows{0}{}{dummy}{\py{0}}  {8cm,   -11cm}{blue!50}
          \drawboxwitharrows{0}{}{dummy}{\py{1}} {9cm,   -11cm}{green}
         \drawboxwitharrows{0}{}{dummy}{\py{1}}  {10cm,  -11cm}{red}
         \drawboxwitharrows{0}{}{dummy}{\py{1}}  {11cm,  -11cm}{blue!50}
         \drawboxwitharrows{0}{}{dummy}{\py{2}}  {12cm,  -11cm}{orange}
         \drawboxwitharrows{0}{}{dummy}{\py{2}}  {13cm,  -11cm}{blue!50}
         \drawboxwitharrows{0}{}{dummy}{\py{3}}  {14cm,  -11cm}{orange}
         \drawboxwitharrows{0}{}{dummy}{\py{3}}  {15cm,  -11cm}{blue!50}
         \drawboxwitharrows{0}{}{dummy}{\py{4}}  {16cm,  -11cm}{orange}
         \drawboxwitharrows{0}{}{dummy}{\py{4}}  {17cm,  -11cm}{blue!50}


         \draw[thick, decoration={markings, mark=at position 0.5 with {\arrow{>}}}, postaction={decorate}] (9, -7.25) to[out=-80, in=60] (6, -10.75); 
          \draw[thick, decoration={markings, mark=at position 0.5 with {\arrow{>}}}, postaction={decorate}] (9, -7.25) to[out=-80, in=60] (9, -10.75); 

        \draw[thick, decoration={markings, mark=at position 0.5 with {\arrow{>}}}, postaction={decorate}] (10, -7.25) to[out=-80, in=60] (7, -10.75); 
          \draw[thick, decoration={markings, mark=at position 0.5 with {\arrow{>}}}, postaction={decorate}] (10, -7.25) to[out=-80, in=60] (10, -10.75);

          \draw[thick, decoration={markings, mark=at position 0.5 with {\arrow{>}}}, postaction={decorate}] (11, -7.25) to[out=-80, in=60] (8, -10.75); 
          \draw[thick, decoration={markings, mark=at position 0.5 with {\arrow{>}}}, postaction={decorate}] (11, -7.25) to[out=-80, in=60] (11, -10.75); 

        \draw[thick, decoration={markings, mark=at position 0.5 with {\arrow{>}}}, postaction={decorate}] (12, -7.25) to[out=-80, in=60] (12, -10.75); 
          \draw[thick, decoration={markings, mark=at position 0.5 with {\arrow{>}}}, postaction={decorate}] (12, -7.25) to[out=-80, in=60] (14, -10.75); 
          \draw[thick, decoration={markings, mark=at position 0.5 with {\arrow{>}}}, postaction={decorate}] (12, -7.25) to[out=-80, in=60] (16, -10.75); 

          \draw[thick, decoration={markings, mark=at position 0.5 with {\arrow{>}}}, postaction={decorate}] (13, -7.25) to[out=-80, in=60] (13, -10.75); 
          \draw[thick, decoration={markings, mark=at position 0.5 with {\arrow{>}}}, postaction={decorate}] (13, -7.25) to[out=-80, in=60] (15, -10.75); 
          \draw[thick, decoration={markings, mark=at position 0.5 with {\arrow{>}}}, postaction={decorate}] (13, -7.25) to[out=-80, in=60] (17, -10.75);

        \node[rotate=0] at (4.5, -11.5) {\py{{t\_edg = [}}}; 
         \node[rotate=0] at (6, -11.5) {\py{{0}}}; 
         \node[rotate=0] at (7, -11.5) {\py{{1}}}; 
         \node[rotate=0] at (8, -11.5) {\py{{2}}}; 
         \node[rotate=0] at (9, -11.5) {\py{{0}}}; 
         \node[rotate=0] at (10, -11.5) {\py{{1}}}; 
         \node[rotate=0] at (11, -11.5) {\py{{2}}}; 
         \node[rotate=0] at (12, -11.5) {\py{{3}}}; 
         \node[rotate=0] at (13, -11.5) {\py{{4}}}; 
         \node[rotate=0] at (14, -11.5) {\py{{3}}}; 
         \node[rotate=0] at (15, -11.5) {\py{{4}}}; 
         \node[rotate=0] at (16, -11.5) {\py{{3}}}; 
         \node[rotate=0] at (17, -11.5) {\py{{4}}}; 
         \node[rotate=0] at (18, -11.5) {\py{]}}; 
    \end{tikzpicture}
    }
\caption{Unfolding Process}
    \label{fig:Unfolding Process}
\end{figure}
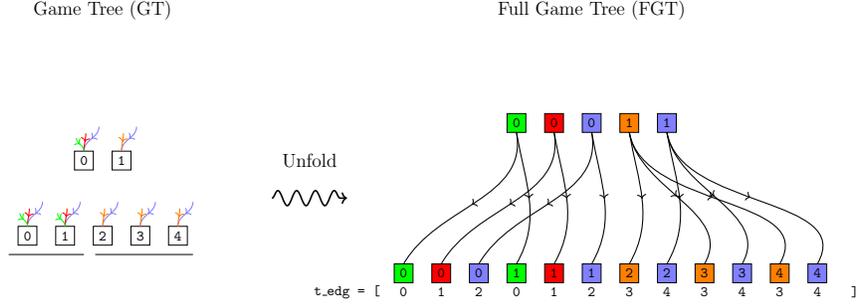

\begin{figure}[!p]
\centering
\caption*{Game Tree (GT)}
\scalebox{0.5}{
    \begin{tikzpicture}

      \drawboxwitharrows{0}{red/0.2}{dummy}{0}{2cm,0cm}{none}
      \drawboxwitharrows{0}{red/0.2}{dummy}{0}{0cm, -2cm}{none}
      \drawboxwitharrows{0}{red/0.2}{dummy}{1}{1cm, -2cm}{none}
      \drawboxwitharrows{0}{red/0.2}{dummy}{2}{2cm, -2cm}{none}
      \drawboxwitharrows{0}{red/0.2}{dummy}{3}{3cm, -2cm}{none}
      \drawboxwitharrows{0}{red/0.2}{dummy}{4}{4cm, -2cm}{none}
      \draw (-0.5cm, -2.5cm) -- (4.5cm, -2.5cm);

      \node[rotate=60] at  (0.5cm, -3.2cm) {\dots\dots};    
      \node[rotate=90] at (2cm, -3.2cm) {\dots\dots};        
      \node[rotate=-60] at (3.5cm, -3.2cm) {\dots\dots};

      \drawboxwitharrows{0}{red/0.2}{dummy}{0}{-2cm, -5cm}{none}
      \drawboxwitharrows{0}{red/0.2}{dummy}{1}{-1cm, -5cm}{none}
      \drawboxwitharrows{0}{red/0.2}{dummy}{2}{0cm, -5cm} {none}
      \node[rotate=0] at  (2cm, -5cm) {\dots\dots};
      \drawboxwitharrows{0}{red/0.2}{dummy}{-3}{4cm,  -5cm}{none}
      \drawboxwitharrows{0}{red/0.2}{dummy}{-2}{5cm,  -5cm}{none}
      \drawboxwitharrows{0}{red/0.2}{dummy}{-1}{6cm,  -5cm}{none}
        \draw (-2.5cm, -5.5cm) -- (-0.5cm, -5.5cm);
        \draw (-0.3cm, -5.5cm) -- (0.4cm, -5.5cm);
        \draw (3.5cm, -5.5cm) -- (5.5cm, -5.5cm);
        \draw (5.7cm, -5.5cm) -- (6.4cm, -5.5cm);

         \draw[blue!50, thick, decoration={markings, mark=at position 0.5 with {\arrow{>}}}, postaction={decorate}] (-2, -5.3) to[out=-80, in=60] (0, -7.75); 
         \draw[orange, thick, decoration={markings, mark=at position 0.5 with {\arrow{>}}}, postaction={decorate}] (4, -5.3) to[out=-80, in=60] (-1, -7.75); 
         \draw[red, thick, decoration={markings, mark=at position 0.5 with {\arrow{>}}}, postaction={decorate}] (-1, -5.3) to[out=-80, in=120] (5, -7.75); 
         \draw[blue!50, thick, decoration={markings, mark=at position 0.5 with {\arrow{>}}}, postaction={decorate}] (0, -5.3) to[out=-80, in=60] (-1, -7.75); 
         \draw[blue!50, thick, decoration={markings, mark=at position 0.5 with {\arrow{>}}}, postaction={decorate}] (5, -5.3) to[out=-80, in=60] (0, -7.75); 
         \draw[green, thick, decoration={markings, mark=at position 0.5 with {\arrow{>}}}, postaction={decorate}] (6, -5.3) to[out=-80, in=60] (5, -7.75); 
        
      \drawboxwitharrows{0}{}{dummy}{0}  {-1cm,  -8cm}{none}
      \drawboxwitharrows{0}{}{dummy}{1}  {0cm,   -8cm}{none}
      \drawboxwitharrows{0}{blue!50/0.4, red/0.05}{dummy}{2}  {1cm,   -8cm}   {none}
      \drawboxwitharrows{0}{blue!50/0.4, black/0.05}{dummy}{3}  {2cm,   -8cm} {none}
      \drawboxwitharrows{0}{blue!50/0.4, orange/0.05}{dummy}{4}  {3cm,   -8cm}{none}
      \drawboxwitharrows{0}{blue!50/0.4, green/0.05}{dummy}{5}  {4cm,   -8cm} {none}
      \drawboxwitharrows{0}{}{dummy}{6}  {5cm,   -8cm}{none}

      \drawboxwitharrows{0}{blue!50/0.4, orange/0.05}{dummy}                                       {0}    {-4cm,  -10cm}{none}
      \drawboxwitharrows{0}{blue!50/0.4, orange/0.05}{dummy}                                       {1}    {-3cm,  -10cm}{none}
      \drawboxwitharrows{0}{blue!50/0.4}{dummy}                                       {2}    {-2cm,  -10cm}{none}
      \drawboxwitharrows{0}{blue!50/0.4}{dummy}                                       {3}    {-1cm,  -10cm}{none}
      \drawboxwitharrows{0}{blue!50/0.4}{dummy}                                       {4}    {0cm,   -10cm}{none}
      \node[rotate=0] at  (3cm, -10cm) {\dots\dots};
      \drawboxwitharrows{0}{green/0.4, red/0.05}{dummy}                                       {-3}   {6cm,   -10cm}{none}
      \drawboxwitharrows{0}{green/0.4, red/0.05}{dummy}                                       {-2}   {7cm,   -10cm}{none}
      \drawboxwitharrows{0}{green/0.4, red/0.05}{dummy}                                       {-1}   {8cm,   -10cm}{none}
       \draw (-4.5cm,    -10.5cm) -- (-2.5cm, -10.5cm);
        \draw (-2.2cm, -10.5cm) -- (0.4cm, -10.5cm);
        \draw (5.7cm, -10.5cm) -- (8.5cm, -10.5cm);

      \node[rotate=-60] at  (-1cm, -12.2cm) {\dots\dots\dots};    
      \node[rotate=90] at (3cm, -12.2cm) {\dots\dots\dots};        
      \node[rotate=60] at (7cm, -12.2cm) {\dots\dots\dots};      

      \drawboxwitharrows{0}{blue!50/0.4, orange/0.05}{dummy}                                       {0}    {-8cm,  -14cm}{none}
      \drawboxwitharrows{0}{blue!50/0.4, orange/0.05}{dummy}                                       {1}    {-7cm,  -14cm}{none}
      \drawboxwitharrows{0}{blue!50/0.4, orange/0.05}{dummy}                                       {2}    {-6cm,  -14cm}{none}
      \drawboxwitharrows{0}{blue!50/0.4, orange/0.05}{dummy}                                       {3}    {-5cm,  -14cm}{none}
      \drawboxwitharrows{0}{blue!50/0.4, orange/0.05}{dummy}                                       {4}    {-4cm,   -14cm}{none}
      \node[rotate=0] at  (3cm, -14cm) {\dots\dots\dots\dots\dots\dots};
      \drawboxwitharrows{0}{green/0.4, red/0.05}{dummy}                                       {-3}   {10cm,   -14cm}{none}
      \drawboxwitharrows{0}{green/0.4, red/0.05}{dummy}                                       {-2}   {11cm,   -14cm}{none}
      \drawboxwitharrows{0}{green/0.4, red/0.05}{dummy}                                       {-1}   {12cm,   -14cm}{none}
       \draw (-4.5cm,    -10.5cm) -- (-2.5cm, -10.5cm);
        \draw (-2.2cm, -10.5cm) -- (0.4cm, -10.5cm);
        \draw (5.7cm, -10.5cm) -- (8.5cm, -10.5cm);

       \draw (-8.4cm,    -14.5cm) -- (-7.6cm, -14.5cm);
       \draw (-7.4cm,    -14.5cm) -- (-6.6cm, -14.5cm);
       \draw (-6.4cm,    -14.5cm) -- (-5.6cm, -14.5cm);
       \draw (-5.4cm,    -14.5cm) -- (-3.6cm, -14.5cm);
        \draw (9.7cm, -14.5cm) -- (12.5cm, -14.5cm);

        \draw[blue!50, thick, decoration={markings, mark=at position 0.5 with {\arrow{>}}}, postaction={decorate}] (-8, -14.3) to[out=-140, in=-2] (-1, -16.3); 
         \draw[green, thick, decoration={markings, mark=at position 0.5 with {\arrow{>}}}, postaction={decorate}] (-8, -14.3) to[out=-100, in=-2] (-1, -16.3); 
         \draw[red, thick, decoration={markings, mark=at position 0.5 with {\arrow{>}}}, postaction={decorate}] (-7, -14.3) to[out=-120, in=1] (0, -16.3); 
          \draw[orange, thick, decoration={markings, mark=at position 0.5 with {\arrow{>}}}, postaction={decorate}] (-7, -14.3) to[out=-90, in=1] (0, -16.3); 
         \draw[orange, thick, decoration={markings, mark=at position 0.5 with {\arrow{>}}}, postaction={decorate}] (-6, -14.3) to[out=-120, in=-2] (-2, -16.3); 
         \draw[red, thick, decoration={markings, mark=at position 0.5 with {\arrow{>}}}, postaction={decorate}] (-6, -14.3) to[out=-90, in=-2] (-2, -16.3); 
         \draw[orange, thick, decoration={markings, mark=at position 0.5 with {\arrow{>}}}, postaction={decorate}] (-5, -14.3) to[out=-120, in=-2] (-2, -16.3); 
          \draw[blue!50, thick, decoration={markings, mark=at position 0.5 with {\arrow{>}}}, postaction={decorate}] (-5, -14.3) to[out=-90, in=-2] (-2, -16.3); 

          \draw[red, thick, decoration={markings, mark=at position 0.5 with {\arrow{>}}}, postaction={decorate}] (10, -14.3) to[out=-60, in=10] (4, -16.3); 
            \draw[blue!50, thick, decoration={markings, mark=at position 0.5 with {\arrow{>}}}, postaction={decorate}] (10, -14.3) to[out=-40, in=10] (4, -16.3); 
           \draw[green, thick, decoration={markings, mark=at position 0.5 with {\arrow{>}}}, postaction={decorate}] (11, -14.3) to[out=-60, in=10] (7, -16.3); 
           \draw[orange, thick, decoration={markings, mark=at position 0.5 with {\arrow{>}}}, postaction={decorate}] (11, -14.3) to[out=-40, in=10] (7, -16.3); 
         \draw[blue!50, thick, decoration={markings, mark=at position 0.5 with {\arrow{>}}}, postaction={decorate}] (12, -14.3) to[out=-60, in=10] (5, -16.3); 
          \draw[red, thick, decoration={markings, mark=at position 0.5 with {\arrow{>}}}, postaction={decorate}] (12, -14.3) to[out=-40, in=10] (5, -16.3);

      \node[rotate=90] at (1cm, -19cm) {\dots\dots\dots};        
      \node[rotate=90] at (3cm, -19cm) {\dots\dots\dots};        
      \node[rotate=90] at (5cm, -19cm) {\dots\dots\dots};

       \drawboxwitharrows{0}{red/-0.2, blue!50/0.4, green/0.05}{dummy}{0}  {-1cm,  -22cm}{none}
      \drawboxwitharrows{0}{red/-0.2, blue!50/0.4, orange/0.05}{dummy}{1}  {0cm,   -22cm}{none}
       \drawboxwitharrows{0}{green/-0.2, blue!50/0.4, red/0.05}{dummy}{2}  {1cm,   -22cm}{none}
        \node[rotate=0] at (3cm, -22cm) {\dots\dots\dots};   
      \drawboxwitharrows{0}{black/-0.2, blue!50/0.4, orange/0.05}{dummy}{-3}  {5cm,  -22cm}{none}
      \drawboxwitharrows{0}{red/-0.2, blue!50/0.4, green/0.05}{dummy}{-2}  {6cm,   -22cm}  {none}
      \drawboxwitharrows{0}{black/-0.2, blue!50/0.4, red/0.05}{dummy}{-1}  {7cm,   -22cm}  {none}
      \node[rotate=60] at (1cm, -25cm) {\dots\dots\dots};        
      \node[rotate=90]  at (3cm, -25cm) {\dots\dots\dots};        
      \node[rotate=-60]  at (5cm, -25cm) {\dots\dots\dots};    

             \drawboxwitharrows{0}{red/-0.2, blue!50/0.4, green/0.05}{dummy}{0}  {-5cm,  -28cm} {none}
      \drawboxwitharrows{0}{red/-0.2, blue!50/0.4, orange/0.05}{dummy}{1}        {-4cm,   -28cm}{none}
      \drawboxwitharrows{0}{green/-0.2, blue!50/0.4, red/0.05}{dummy}{2}         {-3cm,   -28cm}{none}
        \node[rotate=0] at (3cm, -28cm) {\dots\dots\dots\dots\dots\dots\dots\dots\dots\dots\dots\dots};   
      \drawboxwitharrows{0}{black/-0.2, blue!50/0.4, orange/0.05}{dummy}{-3}     {9cm,   -28cm} {none}
      \drawboxwitharrows{0}{red/-0.2, blue!50/0.4, green/0.05}{dummy}{-2}        {10cm,   -28cm}{none}
      \drawboxwitharrows{0}{black/-0.2, blue!50/0.4, red/0.05}{dummy}{-1}        {11cm,   -28cm}{none}

            \draw (-5.3cm,    -28.5cm) -- (-4.7cm, -28.5cm);
            \draw (-4.3cm,    -28.5cm) -- (-3.7cm, -28.5cm);
            \draw (-3.3cm,    -28.5cm) -- (-2.7cm, -28.5cm);
            \draw (8.7cm,    -28.5cm) -- (9.4cm, -28.5cm);
            \draw (9.7cm,    -28.5cm) -- (10.4cm, -28.5cm);
            \draw (10.7cm,    -28.5cm) -- (11.4cm, -28.5cm);

         \draw[blue!50, thick, decoration={markings, mark=at position 0.5 with {\arrow{>}}}, postaction={decorate}] (-5, -28.3) to[out=-70, in=60] (5, -30.7); 
         \draw[red, thick, decoration={markings, mark=at position 0.5 with {\arrow{>}}}, postaction={decorate}] (-5, -28.3) to[out=-90, in=60] (5, -30.7); 
         \draw[green, thick, decoration={markings, mark=at position 0.5 with {\arrow{>}}}, postaction={decorate}] (-4, -28.3) to[out=-80, in=60] (1, -30.7); 
         \draw[orange, thick, decoration={markings, mark=at position 0.5 with {\arrow{>}}}, postaction={decorate}] (-4, -28.3) to[out=-60, in=60] (1, -30.7); 
         \draw[red, thick, decoration={markings, mark=at position 0.5 with {\arrow{>}}}, postaction={decorate}] (-3, -28.3) to[out=-80, in=60] (2, -30.7); 
         \draw[black, thick, decoration={markings, mark=at position 0.5 with {\arrow{>}}}, postaction={decorate}] (-3, -28.3) to[out=-60, in=60] (2, -30.7); 
         \draw[green, thick, decoration={markings, mark=at position 0.5 with {\arrow{>}}}, postaction={decorate}] (-4, -28.3) to[out=-80, in=60] (1, -30.7); 
         \draw[orange, thick, decoration={markings, mark=at position 0.5 with {\arrow{>}}}, postaction={decorate}] (-4, -28.3) to[out=-60, in=60] (1, -30.7); 
         \draw[black, thick, decoration={markings, mark=at position 0.5 with {\arrow{>}}}, postaction={decorate}] (10, -28.3) to[out=-80, in=60] (4, -30.7); 
         \draw[orange, thick, decoration={markings, mark=at position 0.5 with {\arrow{>}}}, postaction={decorate}] (10, -28.3) to[out=-60, in=60] (4, -30.7); 
         \draw[green, thick, decoration={markings, mark=at position 0.5 with {\arrow{>}}}, postaction={decorate}] (9, -28.3) to[out=-80, in=60] (3, -30.7); 
         \draw[red, thick, decoration={markings, mark=at position 0.5 with {\arrow{>}}}, postaction={decorate}] (9, -28.3) to[out=-60, in=60] (3, -30.7); 
         \draw[green, thick, decoration={markings, mark=at position 0.5 with {\arrow{>}}}, postaction={decorate}] (11, -28.3) to[out=-80, in=60] (3, -30.7); 
         \draw[red, thick, decoration={markings, mark=at position 0.5 with {\arrow{>}}}, postaction={decorate}] (11, -28.3) to[out=-60, in=60] (3, -30.7);

         \draw[green, thick, decoration={markings, mark=at position 0.5 with {\arrow{>}}}, postaction={decorate}] (5, -28.3) to[out=-80, in=60] (6, -30.7); 
         \draw[red, thick, decoration={markings, mark=at position 0.5 with {\arrow{>}}}, postaction={decorate}] (5, -28.3) to[out=-60, in=60] (6, -30.7);

         \draw[blue!50, thick, decoration={markings, mark=at position 0.5 with {\arrow{>}}}, postaction={decorate}] (2, -28.3) to[out=-80, in=60] (7, -30.7); 
         \draw[red, thick, decoration={markings, mark=at position 0.5 with {\arrow{>}}}, postaction={decorate}] (2, -28.3) to[out=-60, in=60] (7, -30.7); 
         
         \draw[blue!50, thick, decoration={markings, mark=at position 0.5 with {\arrow{>}}}, postaction={decorate}] (-2, -28.3) to[out=-80, in=60] (7, -30.7); 
         \draw[green, thick, decoration={markings, mark=at position 0.5 with {\arrow{>}}}, postaction={decorate}] (-2, -28.3) to[out=-60, in=60] (7, -30.7);

      \drawboxwitharrows{0}{}{dummy}{0}                     {1cm,  -31cm} {none}
      \drawboxwitharrows{0}{}{dummy}{1}                     {2cm,  -31cm} {none}
      \drawboxwitharrows{0}{}{dummy}{2}                     {3cm,  -31cm} {none}
      \drawboxwitharrows{0}{}{dummy}{3}                     {4cm,   -31cm}{none}
      \drawboxwitharrows{0}{}{dummy}{4}                     {5cm,   -31cm}{none}
      \drawboxwitharrows{0}{}{dummy}{5}                     {6cm,   -31cm}{none}
      \drawboxwitharrows{0}{}{dummy}{6}                     {7cm,   -31cm}{none}

    \end{tikzpicture}
    }
\caption{We distinguish between the \emph{card state} and the \emph{score information}, which are updated independently. A given card state may correspond to multiple incoming edges, each representing a different score inherited from earlier rounds. Underlines indicate the number of available actions for each parent node.}

    \label{fig:gametreegeneration}
\end{figure}

This section is organized as follows. We begin by describing the mask and padding tensors used during game tree generation in Subsection \ref{sec:mask}. These tensors help track the indices of cards within the current tensor as the size of these tensors may vary from round to round due to the fact that the set of in-play cards changes dynamically. Next, in Subsection~\ref{sec:gametensor}, we describe the construction of the \emph{Game Tensors}. These tensors encode the state and action history. Subsection~\ref{sec:actions} explains how the \emph{Action Tensors} are built and how they are used to update the Game Tensors. Once the construction of the Action Tensors is understood, we proceed to explain the construction of the Compressed Tensor in Subsection~\ref{sec:comptensor}, where all game information up to the current point is encoded in a tensor of shape \py{[58]}. Next, Subsection~\ref{sec:scoretensor} explains the two types of score tensors used: the \emph{Running-Score Tensor}, which accumulates scores across a single round, and the score tensors that are passed along the edges of GT, containing scores inherited from previous rounds. Then, in Subsection~\ref{sec:inhandupdates}, we detail how \emph{FGT Tensors} are updated within each round; it also introduces the construction of \emph{Edge Tensors}. Following that, Subsection~\ref{sec:bethand} describes the Between-Hand updates, including how the Score Tensor and FGT tensors are updated at the end of each round, along with the \emph{Linkage Tensor}. Finally, Subsection~\ref{sec:infosettensor} explains how \emph{Infoset Tensors} are constructed. 

\newpage
Before proceeding to the next section, we summarize the key components of our framework in Table~\ref{table:tensors}. Padding-related tensors are described separately in Table~\ref{table:mask_pad_vars}.

\begin{table}[h!]
\centering
\scriptsize
\caption{Summary of Tensors Used in Our Framework}
\label{tab:gametree_tensors_detailed}
\begin{tabular}{
>{\centering\arraybackslash}m{1.3cm}  
>{\centering\arraybackslash}m{1.3cm}  
>{\centering\arraybackslash}m{1cm}    
>{\centering\arraybackslash}m{1.3cm}  
>{\centering\arraybackslash}m{2cm}    
>{\raggedright\arraybackslash}p{5.5cm}
}
\toprule
\textbf{Component} & \textbf{Tensor} & \textbf{Shape} & \textbf{Type} & \textbf{Update} & \textbf{Description} \\
\midrule
Game & \py{t\_gme} & \py{[M,3,m]} & \py{int8} & Per turn & Encodes state and action history per node. Each slice represents a singleton node in the tree layer. \\
\midrule
In-Play Cards & \py{t\_inp} & \py{[52]} & \py{int8} & Per round & Boolean indicator for whether each of the 52 cards is still in play (held or in pool). \\
\midrule
Dealt Cards & \py{t\_dlt} & \py{[52]} & \py{int8} & Per round & Boolean indicator marking cards that have been dealt in previous rounds. \\
\midrule
Full Game & \py{t\_fgm} & \py{[Q, 2]} & \py{int32} & Per turn/round & Each row \py{[g, s]} in \py{t\_fgm} indicates that GT's node \py{t\_gme[g,:]} inherits the score with ID \py{s} from \py{t\_scr}. Here, \py{Q} denotes FGT's current layer's number of nodes.
 \\
\midrule
Scores & \py{t\_scr} & \py{[Q, 4]} & \py{int8} & Per turn/round & Unique score in form (Alex Clubs, Bob Clubs, Point Difference, 7-Clubs Bonus). Scaled each turn using $\otimes$\py{t\_brf}. See Section~\ref{sec:inhandupdates} for in-between hands updates. \\
\midrule
Runnsing Scores & \py{t\_rus} & \py{[M,7]} & \py{int8} & Per turn & Unique score in form (Alex Club, Bob Club, Last Picker, Alex Points, Alex Sur, Bob Points, Bob Sur). \\
\midrule
Action & \py{t\_act} & \py{[M',2,m]} & \py{int8} & Per turn & Action representation per node. \py{[0,:]} encodes the lay card; \py{[1,:]} encodes picked cards. Here, \py{M'} denotes the number of nodes in GT's next layer. \\
\midrule
Branch Factor & \py{t\_brf} & \py{[M]} & \py{int8} & Per turn & Number of valid actions available from each node; used to replicate game states before applying actions. Indicated using underlines in Figure~\ref{fig:gametreegeneration}. \\
\midrule
Linkage & \py{t\_lnk} & \py{[Q]} & \py{int32} & Per round & Connects FGT nodes between consecutive hands to identify how scores and states map across hands. \\
\midrule
Edge & \py{t\_edg} & \py{[Q']} & \py{int32} & Per turn & Records edges between FGT nodes to track the structure of the overall graph. Here, \py{Q'} denotes the number of nodes in the FGT's next layer. \\
\midrule
Strategy & \py{t\_sgm} & \py{[Q]} & \py{float32} & Per turn & Stores strategy values (e.g., probabilities) associated with each FGT's node. \\
\midrule
Compressed Game & \py{t\_cmp} & \py{[M, m]} & \py{int8} & Per turn & Stores the entire game information up to the current turn in a compressed format. \\
\midrule
Infoset & \py{t\_inf} & \py{[Q,58]} & \py{int8} & Per turn & Encodes information available to player whose turn it is. \py{t\_inf} hides information that is not observable by the acting player. Includes metadata such as round index, turn counter, and cumulative score up to the previous round. \\
\bottomrule
\end{tabular}
\label{table:tensors}
\end{table}

\newpage 

\subsection{Mask and Padding Tensors}\label{sec:mask}

We now describe how masks and padding tensors are constructed from the in-play card tensor \py{t\_inp}, which has shape \py{[M, 52]} and indicates which cards are active during a round.

\begin{table}[H]
\centering
\scriptsize
\caption{Mask and Padding Variables}
\label{table:mask_pad_vars}
\begin{tabular}{>{\ttfamily}l p{10cm}}
\toprule
\textbf{Variable} & \textbf{Description} \\
\midrule
t\_inn,\ t\_inq,\ t\_ink,\ t\_inj,\ t\_inc,\ t\_ins & Masks for Numerical, Queen, King, Jack, Club, and Point cards resp. \\
\midrule
i\_pdn,\ i\_pdj,\ i\_pdq,\ i\_pdk & Padding sizes for Numerical, Jack, Queen, and King actions resp. \\
\midrule
t\_pdq,\ t\_pdk & Permutations to restore Queen and King actions to original order, resp. \\
\bottomrule
\end{tabular}
\end{table}

\begin{table}[H]
\centering
\scriptsize
\caption{Index Sets for Card Categories and Scoring}
\label{table:card_index_sets}
\begin{tabular}{>{\ttfamily}l p{10cm}}
\toprule
\textbf{Index Set} & \textbf{Description} \\
\midrule
\py{l\_n = range(40)} & Numerical cards (2 to 10 of all suits including Aces). \\
\py{l\_j = [40, 41, 42, 43]} & Jacks. \\
\py{l\_q = [44, 45, 46, 47]} & Queens. \\
\py{l\_k = [48, 49, 50, 51]} & Kings. \\
\midrule
\py{l\_c = range(0, 52, 4)} & Clubs (every 4th card). \\
\py{l\_p = [0,1,2,3,4,37,40,41,42,43]} & Scoring cards: all Aces, $\TwC$, $\TeD$, all Jacks. \\
\py{l\_s = [1,1,1,1,2,3,1,1,1,1]} & Scores corresponding to entries in \py{l\_p}. \\
\bottomrule
\end{tabular}
\end{table}
To compute masks in Table~\ref{table:mask_pad_vars}, we use the index sets from Table~\ref{table:card_index_sets}, which are computed as follows:
\begin{flushleft}
\py{t\_inn, t\_inq, t\_ink, t\_inj, t\_inc, t\_ins =}
\end{flushleft}
\begin{flushright}
  \py{(f(index\_set) for index\_set in [l\_n,l\_q,l\_k,l\_j,l\_c,l\_s])}
\end{flushright}
Here the helper function \py{f} is defined below:
\[
\py{f = lambda\ lm:\ tensor([id in lm for id,card in enumerate(t\_inp)\ if card])}
\]
We compute the counts of each category:
\begin{center}
\py{i\_n,\ i\_j,\ i\_q,\ i\_k\ =\ (t.sum()\ for\ t\ in\ [t\_inn,\ t\_inj,\ t\_inq,\ t\_ink])}
\end{center}
We also define intermediate sums:
\begin{align*}
\py{i\_kq}    &= \py{i\_k + i\_q} \\
\py{i\_jqk}   &= \py{i\_j + i\_q + i\_k} \\
\py{i\_njk}   &= \py{i\_n + i\_j + i\_k} \\
\py{i\_njq}   &= \py{i\_n + i\_j + i\_q} \\
\py{i\_njqk}  &= \py{i\_n + i\_j + i\_q + i\_k}
\end{align*}
Using these, we define:
\begin{center}
\py{i\_pdn, i\_pdj, i\_pdk, i\_pdq = i\_jqk, i\_kq, i\_njq, i\_njk}
\end{center}
And the corresponding padding permutations:
\begin{align*}
\py{t\_pdk} &= \py{hstack((arange(i\_k,\ i\_njqk),\ arange(i\_k)))} \\
\py{t\_pdq} &= \py{hstack((arange(i\_q,\ i\_njq),\ arange(i\_q),\ arange(i\_njq,\ i\_njqk)))}
\end{align*}

\subsection{Game Tensor}\label{sec:gametensor}
Each row in the game tree (Figure~\ref{fig:gametreegeneration}) is represented using an \py{[M,3,m]} GT tensor \py{t\_gme}, where \py{M} corresponds to the number of nodes in the current GT's layer. The associated \py{[3,m]} tensor is constructed for each node: Each slice \py{t\_gme[g,:,:]} encodes a single game state. The first row of the tensor, \py{t\_gme[g,0,:]}, represents the current card holdings and pool status. To encode the game state numerically, we assign integer identifiers to the entities involved, as shown in Table~\ref{tab:state_encoding}.

\begin{table}[h!]
\centering
\caption{State Encoding in \py{t\_gme[g,0,:]}}
\label{tab:state_encoding}
\begin{tabular}{ll}
\toprule
\textbf{Element} & \textbf{Encoding} \\
\midrule
Alex                 & 1 \\
Bob                  & 2 \\
Pool                 & 3 \\
\bottomrule
\end{tabular}
\end{table}

The remaining two rows of \py{t\_gme}, namely \py{t\_gme[g,1,:]} and \py{t\_gme[g,2,:]}, record the action histories of Alex and Bob, resp. We describe the construction of \py{t\_gme[g,1,:]} in detail below; the construction of \py{t\_gme[g,2,:]} is analogous.

Each round consists of four turns, meaning that Alex plays a card four times per round. During each turn \py{i\_trn = i}, for \py{i = 0, 1, 2, 3}, he may or may not collect cards from the pool.

Suppose Alex plays the card $\AH$ on his first turn and collects $\TeH$. We update \py{t\_gme[g, 1,:]} by recording the value \py{1} at the position corresponding to $\AH$, and adding the value \py{10} at the position corresponding to $\TeH$. If Alex collects multiple cards from the pool on his first turn, the value \py{10} is added to all corresponding positions in \py{t\_gme[g, 1,:]} for each collected card. For subsequent turns, similar updates are performed, except we use the value pairs \py{(2,20)}, \py{(3,30)}, and \py{(4,40)} instead of \py{(1,10)} for the second, third, and fourth turns, respectively.

Importantly, \py{t\_gme[g,0,:]} is updated after each move to reflect the current state of the game. For cards that are played but not collected, the corresponding position in \py{t\_gme[g,0,:]} is set to \py{3}. If a played card is used to collect one or more cards from the pool, then the corresponding positions are updated as follows: hand cards that were played or picked change from \py{1} to \py{0}, and pool cards that were collected change from \py{3} to \py{0}. See CodeSnippet \ref{alg:applyactions}.

\subsection{Actions}\label{sec:actions}

At each turn, we first compute two key tensors: the \emph{Branch Factor Tensor} \py{t\_brf} and the \emph{Action Tensor} \py{t\_act}: \py{t\_brf} is a tensor of length \py{M}, where \py{M} is the number of nodes in the current round. Each entry \py{t\_brf[g]} records the number of valid actions available from node \py{g}. Correspondingly, \py{t\_act} is a binary tensor of shape \py{[M',2,m]}, where \py{m} is the number of in-play cards. Here, \py{M'=t\_brf.sum()} is the total number of resulting nodes in the next round, created by enumerating all valid actions from the current nodes. Each slice \py{t\_act[j,\ :,\ :]} encodes a single action, consisting of two rows as described in Table~\ref{table:action_tensor_structure}.
\begin{table}[H]
\centering
\scriptsize
\caption{Structure of the Action Tensor \py{t\_act}}
\label{table:action_tensor_structure}
\begin{tabular}{>{\ttfamily}l p{10cm}}
\toprule
\textbf{Tensor Slice} & \textbf{Description} \\
\midrule
\py{t\_act[j, 0, :]} & Encodes the \emph{lay card}. Contains exactly one \py{1} at the index of the played card; all other entries are \py{0}. \\
\py{t\_act[j, 1, :]} & Encodes the \emph{picked cards} from the pool. May contain zero or more \py{1}s depending on the number of picked cards. \\
\bottomrule
\end{tabular}
\end{table}


To update the game state tensor \py{t\_gme} for the next round, we first replicate each current node \py{g} according to its corresponding \py{t\_brf[g]} count, effectively expanding \py{t\_gme} to match the total number of action slices \py{M'}. This is done using the \py{repeat\_interleave} operator, denoted by the Kronecker product symbol: $\otimes$.
\begin{center}
    \py{t\_gme} $\gets$ \py{t\_gme} $\otimes$ \py{t\_brf}
\end{center}
This expansion ensures that each valid action is paired with its own copy of the corresponding game state. Then, each slice of \py{t\_act} is applied to the corresponding replicated game state to produce the updated states. Finally, we apply the encoded actions by updating \py{t\_gme} using \py{t\_act}. See CodeSnippet~\ref{alg:applyactions}. 

\begin{algorithm}
\caption{ApplyActions}
\label{alg:applyactions}
\begin{algorithmic}[1]
\State Input \py{t\_gme, t\_act, i\_ply, i\_trn}
\Statex \textcolor{green!50!black}{\scriptsize \# \py{t\_act.shape[0] = t\_gme.shape[0]} after expansion \py{t\_gme} $\gets$ \py{t\_gme} $\otimes$ \py{t\_brf}}
\State \py{t\_mpk} $\gets$ \py{any(t\_act[:,1,:], dim=1)}
\Statex \textcolor{green!50!black}{\scriptsize \# Whether a pick action occurs}

\State \py{t\_gme[t\_mpk,0,:]+=(2-i\_ply)*t\_act[t\_mpk,0,:]}
\Statex \textcolor{green!50!black}{\scriptsize \# Add lay card to pool (only if no pick)}

\State \py{t\_gme[t\_mpk,0,:]-=(1+i\_ply)*t\_act[t\_mpk,0,:]}
\Statex \textcolor{green!50!black}{\scriptsize \# Remove lay card from player's hand}

\State \py{t\_gme[t\_mpk,0,:]-=3*t\_act[t\_mpk,1,:]}
\Statex \textcolor{green!50!black}{\scriptsize \# Remove picked cards from the pool}

\State \py{t\_gme[:,i\_ply+1,:]+=(i\_trn+1)*t\_act[:,0,:]+10*(i\_trn+1)*t\_act[:,1,:]}
\Statex \textcolor{green!50!black}{\scriptsize \# Update player record: lay (weighted by i\_trn+1) + pick (weighted by 10 × i\_trn+1)}
\end{algorithmic}
\end{algorithm}

We now present how the Action Tensor \py{t\_act} is constructed using the GT tensor \py{t\_gme}. There are four types of actions: Numerical, Jack, King, and Queen. The construction process is explained below. To derive actions from a given \py{t\_gme} tensor, we first construct an \py{int8} tensor of shape \py{[2, 52]} denoted as \py{t\_2x52}, along with various views such as \py{t\_2x40}, \py{t\_2x44}, and two \py{t\_2x4} tensors corresponding to different action types. The construction of \py{t\_2x52} is explained in CodeSnippet~\ref{cs:t_2x52}. 
\begin{algorithm}
  \caption{\py{t\_2x52}}
  \label{cs:t_2x52}
  \begin{algorithmic}[1]
    \State Input \py{t\_gme}
    \State \py{t\_2x52 $\gets$ zeros((t\_gme.shape[0],2,t\_gme.shape[2]))}
    \State  \py{t\_2x52[:,0,:][t\_gme[:,0,:]==i\_ply+1]$\gets$1}
    \State \py{t\_2x52[:,1,:][t\_gme[:,0,:]==3]$\gets$1}
  \end{algorithmic}
  \end{algorithm}
Once \py{t\_2x52} is obtained, we define the inputs to each action operator as:
\begin{center}
\py{t\_2x52[:,:,t\_msk] for t\_msk in [t\_inn, t\_inn+t\_inj, t\_ink, t\_inq]}
\end{center}
These serve as inputs to the Numerical, Jack, King, and Queen action routines, respectively. For memory efficiency, we first apply \py{unique} to these tensors and pass the result to \py{n\_act}, \py{j\_act}, \py{k\_act}, and \py{q\_act} to compute the action and branch factor tensors. We next pass each resulting action and branch tensor pair \py{t\_act, t\_brft} to CodeSnippet~\ref{cs:actionpropagation} to map them back to the full tensor. 

\begin{algorithm}[H]
\caption{Mapping Back Actions from Unique Tensor to Full Tensor}
\label{cs:actionpropagation}
\begin{algorithmic}[1]
  \State \textbf{Input} \py{t}, \py{f\_act} \textcolor{green!50!black}{\scriptsize \# \py{t}: input tensor,\ \py{f\_act}: function to find actions (e.g., numerical, jack, etc.)}
  \State \py{tu, t\_inx $\gets$ unique(t, sorted=Fasle, return\_inverse=True)}
  \State \py{tu\_act, tu\_brf $\gets$ f\_act(tu)}
  \State \py{t\_brf $\gets$ tu\_brf[t\_inx]}
  \State \py{t\_act $\gets$ tu\_act[inverseunique(tu\_brf,t\_inx)]}
  \State return \py{t\_brf, t\_act}

  \Statex
  \Function{inverseunique}{\py{t\_cnt,t\_inx}} \textcolor{green!50!black}{\scriptsize \# Ex: \py{t\_cnt = [2,3], t\_inx = [1,0,0]}}
    \State \py{t\_cms $\gets$ zeros(t\_cnt.shape[0]+1, dtype=int32)} \textcolor{green!50!black}{\scriptsize \# \py{= [0,2,5]}}
    \State \py{t\_cms[1:] $\gets$ t\_cnt.cumsum(0)}
    \State \py{t\_bgn $\gets$ t\_cms[t\_inx]} \textcolor{green!50!black}{\scriptsize \# \py{= [2,0,0]}}
    \State \py{t\_len $\gets$ t\_cnt[t\_inx]} \textcolor{green!50!black}{\scriptsize \# \py{= [3,2,2]}}
    \State \py{t\_lns $\gets$ t\_len.cumsum(0)} \textcolor{green!50!black}{\scriptsize \# \py{= [3,5,7]}}
    \State \py{t\_ofs $\gets$ arange(t\_lns[-1])-(t\_lns-t\_len)}$\otimes$\py{t\_len}
    \Statex \textcolor{green!50!black}{\scriptsize \# \py{arange(t\_lns[-1])=[0,1,2,3,4,5,6], t\_lns-t\_len=[0,3,5]}}
    \Statex \textcolor{green!50!black}{\scriptsize \# \py{\py{(t\_lns - t\_len)}$\otimes$\py{t\_len = [0,0,0,3,3,5,5]}, t\_ofs = [0,1,2,0,1,0,1]}}
    \State \textbf{return} \py{t\_bgn}$\otimes$\py{t\_len + t\_ofs} \textcolor{green!50!black}{\scriptsize \# = [2,2,2,0,0,0,0]+[0,1,2,0,1,0,1] = [2,3,4,0,1,0,1]}
  \EndFunction
\end{algorithmic}
\end{algorithm}
We denote the resulting action tensors as \py{t\_pck, t\_lay, t\_kng, t\_jck, t\_qun} and the corresponding branch factor tensors as \py{c\_pck, c\_lay, c\_kng, c\_jck, c\_qun}.  It is emphasized that \py{n\_act} produces two pairs of action and branch factor tensors: Pick and Lay pairs. Pick actions correspond to cases where at least one card (along with the card in hand) is picked from the pool, while Lay actions correspond to cases where no card is picked and only one card is laid down and added to the pool. 

We pad these tensors to match the original shape of tensor \py{t\_2x52} as follows:
\begin{center}
\begin{tabular}{l}
\py{t\_pck = pad(t\_pck,\ (0,\ i\_pdn))} \\
\py{t\_lay = pad(t\_lay,\ (0,\ i\_pdn))} \\
\py{t\_kng = pad(t\_kng,\   (0,\ i\_pdk))[:,:,t\_pdk]} \\
\py{t\_qun = pad(t\_qun,\   (0,\ i\_pdq))[:,:,t\_pdq]} \\
\py{t\_jck = pad(t\_jck,\   (0,\ i\_pdj))} \\
\end{tabular}
\end{center}
The branch factor is also easily calculated as below 
\begin{center}
  \py{t\_brf = sum(stack([c\_pck, c\_lay, c\_kng, c\_qun, c\_jck]), dim=0)}
\end{center}
We perform one final step to construct the action tensor \py{t\_act}. This step ensures that all actions corresponding to each row of the game tensor are grouped together. To achieve this, we repeat each row index of the game tensor according to its corresponding branch factor, and then use the sorted indices to reorder the concatenated action tensors. 
\[
  \py{\_, t\_inx = sort(cat([arange(M)} \otimes \py{t for t in [c\_pck,c\_lay,c\_kng,c\_qun,c\_jck]]))}
\]
  Once \py{t\_inx} is obtained, the final action tensor is constructed by concatenating all action components and reordering them using \py{t\_inx}:
\begin{center}
  \py{t\_act = cat([t\_pck,\ t\_lay,\ t\_kng,\ t\_qun,\ t\_jck])[t\_inx]}
\end{center}

\subsubsection{Numerical Actions}

In this section, we explain how Numerical Actions are constructed.  Recall that the input tensor to \py{n\_act} is given by \py{t\_2x40 = t\_2x52[:,:,t\_inn]}. 

We begin by identifying all possible combinations of numerical cards such that the sum of each subset equals \py{11}. This is a classical instance of the \emph{Subset Sum Problem} (SSP), which we solve using dynamic programming. The solutions are stored in a tensor \py{t\_40x2764} of shape \py{[40, 2764]}, where each row corresponds to one of the \py{40} numerical cards, and each column represents a valid subset whose sum equals \py{11}. We also define an auxiliary tensor \py{t\_tpl} as in Table~\ref{table:tpl_structure}.
\begin{table}[H]
\centering
\scriptsize
\caption{Tuple Tensor \py{t\_tpl} of shape \py{[2, 2764]}}
\label{table:tpl_structure}
\begin{tabular}{>{\ttfamily}l p{10cm}}
\toprule
\textbf{Row} & \textbf{Description} \\
\midrule
\py{t\_tpl[0,:]} & All entries are \py{1}, representing a fixed lay card for each action. \\
\py{t\_tpl[1,:]} & Stores the number of cards from the pool that are involved in each action. \\
\bottomrule
\end{tabular}
\end{table}
Since we mask only for in-play cards, we need to exclude those that are not in-play. To this end, we apply the mask \py{t\_inp[:40]} to \py{t\_40x2764} by computing \py{t\_40x2764[:, t\_inp[:40]]}, and pass the resulting tensor to \py{n\_act}. We still denote \py{t\_40x2764} as the masked tensor for convenience. Moreover, denote by \py{t\_2764x40} the transpose of \py{t\_40x2764}. We now describe how tensors \py{t\_pck} and \py{t\_lay} are constructed. The process begins with \py{t\_pck}, from which \py{t\_lay} is subsequently derived.  Consider the matrix product
\[
\py{matmul(t\_2x40, t\_40x2764)}
\]
which results in a tensor of shape \py{[M,2,2764]}, where \py{M} is the number of rows in \py{t\_2x40}. This tensor evaluates how each input pair of player hand and pool (i.e., each row of \py{t\_2x40}) overlaps with the precomputed subset-sum solutions in \py{t\_40x2764}. Notice masking out \py{t\_40x2764} does not affect the outcome of the matrix multiplication, as masked entries are \py{0} and thus do not contribute to the result.

A pair of player hand and pool (i.e., row \py{i} of \py{t\_2x40}) is said to have the action \py{j} available if and only if the count of matching cards exactly matches the template \py{t\_tpl[:, j]}. The following code yields all indices \py{[i, j]} such that action \py{j} is valid for row \py{i} of \py{t\_2x40}.

\begin{center}
\py{t\_inx = nonzero(((t\_tpl - matmul(t\_2x40, t\_40x2764)) == 0).all(dim=1))}
\end{center}
We use the index tensor \py{t\_inx} to construct the pick action tensor \py{t\_pck}. Computing the corresponding \py{t\_brf} is straightforward from \py{t\_inx}; it is computed as follows:
\begin{center}
  \py{c\_pck = t\_inx[:,0].bincount(minlength=M)}
\end{center}
Now that we have the number of actions availble for each row and also the indices of the corresponding actions in \py{t\_inx}, we could obtain \py{t\_pck} as follows:  We repeat each row of \py{t\_2x40} based on the number of available actions for that row. Simultaneously, we repeat each available action twice along \py{dim=1}. Then, we apply the \py{logical\_and} operator to identify matching cards. The following performs this computation:
\[
\py{t\_pck = logical\_and(t\_2x40}\otimes\py{c\_pck, t\_2764x40[t\_inx[:,1]].unsqueeze(1)}\otimes_1\py{2})
\]
Here each row of \py{t\_2764x40[t\_inx[:,1]]} corresponds to a valid available action. 

We are now ready to construct the Lay Action tensor \py{t\_lay}. The main idea is to identify which cards in the player's hand have not been used in any Pick Action. To achieve this, we first build a tensor \py{t\_hnd} that records how many times each card in the hand has been selected for a Pick Action. This is accomplished using the \py{scatter\_add\_} operation. The index tensor used for this purpose has the same shape as \py{t\_pck[:,0,:]}, and rows corresponding to the same row in \py{t\_2x40} are grouped together through this index tensor. Once \py{t\_hnd} is constructed, we subtract it from the player's original hand tensor \py{t\_2x40[:,0,:]} and apply \py{relu} to isolate the remaining cards that were not picked. The result is a binary tensor indicating candidate cards for Lay Actions. The remainder of the construction is straightforward and presented in CodeSnippet~\ref{cs:construct_lay_tensor}.

\begin{algorithm}[H]
\caption{Constructing Lay Action Tensor from Pick Tensor}
\label{cs:construct_lay_tensor}
\begin{algorithmic}[1]
\State \py{m\_pck $\gets$ c\_pck > 0}
\State \py{t\_cnt $\gets$ c\_pck[m\_pck]}
\State \py{i\_cnt  $\gets$ t\_cnt.shape[0]}
\vspace{2mm}
\State \textcolor{green!50!black}{\scriptsize \# Construct hand tensor by summing picked cards}
\State \py{t\_hnd $\gets$ zeros((i\_cnt, m))}
\State \py{t\_inx $\gets$ (arange(i\_cnt)}$\otimes$\py{t\_cnt).view(-1,1).expand(-1, m)}
\State \py{t\_src $\gets$ t\_pck[:,0,:]}
\State \py{t\_hnd.scatter\_add\_(0, t\_inx, t\_src)}
\vspace{2mm}
\State \textcolor{green!50!black}{\scriptsize \# Remove picked cards from hand to construct lay}
\State \py{t\_cln $\gets$ t\_2x40[:,0,:].clone()}
\State \py{t\_cln[m\_pck,:] $\gets$ relu(t\_2x40[m\_pck,0,:]-t\_hnd)}
\vspace{2mm}
\State \textcolor{green!50!black}{\scriptsize \# Find positions to lay the remaining card}
\State \py{t\_inx $\gets$ nonzero(t\_cln == 1)}
\State \py{c\_lay $\gets$ bincount(t\_inx[:,0], minlength=M)}
\State \py{t\_lay $\gets$ zeros((t\_inx.shape[0], 2, m))}
\State \py{t\_lay[arange(t\_lay.shape[0]), 0, t\_inx[:,1]] $\gets$ 1}
\end{algorithmic}
\end{algorithm}

\subsubsection{Jack Actions}
As discussed earlier, \py{t\_2x44 = t\_2x52[:,:,t\_inn+t\_inj]} serves as the input tensor to the \py{j\_act} function. The first step is to identify all rows in \py{t\_2x44} that contain Jack cards. This is accomplished by:
\begin{center}
  \py{t\_inx $\gets$ nonzero(t\_2x44[:, 0, i\_pdn:])}
\end{center}
Recall from Table \ref{table:mask_pad_vars} that \py{i\_pdn} denotes the number of numerical cards among the current in-play cards. The corresponding branch factor tensor is obtained via:
\begin{center}
  \py{c\_jck $\gets$ t\_inx[:, 0].bincount(minlength=M)}
\end{center}
To construct the Jack Action tensor \py{t\_jck}, we first let \py{i\_cnt = t\_inx.shape[0]}. Since each Jack card corresponds to exactly one action, the tensor \py{t\_jck} is initialized with shape \py{[i\_cnt, 2, m]}. This tensor is then populated as follows:
\begin{center}
  \begin{tabular}{l}
    \py{t\_rng $\gets$ arange(i\_cnt)} \\
    \py{t\_jck[t\_rng, 0, i\_pdn + t\_inx[:,1]] $\gets$ 1} \\
    \py{t\_jck[t\_rng, 1, :] $\gets$ t\_2x44[:,1,:]$\otimes$c\_jck}
  \end{tabular}
\end{center}
\subsubsection{King Actions}
In this section, we describe how King Action is constructed; the construction of Queen Action follows similarly. Recall that the input to \py{k\_act} is given by \py{t\_2x4 = t\_2x52[:, :, t\_ink]}. Since we apply \py{unique} before passing the tensor to \py{k\_act} (as explained earlier), the input tensor \py{t\_2x4} has a limited number of unique possibilities. Specifically, the number of rows in \py{t\_2x4} cannot exceed \py{3**4 = 81}. Each row of \py{t\_2x4} is first converted into a string of length \py{4}. To generate this string representation, we first apply the transformation
\[
\py{t\_2x4[:, 1, :][t\_2x4[:, 1, :] == 1] = 2},
\]
and then compute \py{t\_2x4.sum(dim=1)}, which is subsequently converted into a string key. For example, \py{'0201'} indicates that the player holds a $\KS$, while the pool contains a $\KD$. This key is then used to index into the dictionary to retrieve the corresponding action tensor. Finally, we use this key to look up the corresponding action tensor in our dictionary. CodeSnippet~\ref{cs:king_action_tensor} summarizes this discussion. Details regarding the lookup table are omitted here; please refer to the GitHub code repository for full implementation details.

\begin{algorithm}[H]
\caption{Constructing King Action Tensor}
\label{cs:king_action_tensor}
\begin{algorithmic}[1]
\State \py{t\_2x4[:, 1, :][t\_2x4[:, 1, :] == 1] $\gets$ 2}
\State \py{M0 $\gets$ t\_2x4.shape[0]}
\State Build \py{lookup\_tbl} using \py{t\_2x4.sum(dim=1)}.
\State \py{c\_k $\gets$ tensor([lookup\_tbl[i].shape[0] for i in range(M0)])}
\State \py{t\_k $\gets$ cat([lookup\_tbl[i] for i in range(M0)], dim=0)}
\end{algorithmic}
\end{algorithm}

\subsection{Compressed Game Tensor}\label{sec:comptensor}
In this subsection, we represent a game tensor \py{t\_gme} of shape \py{[M, 3, m]} using a compressed form \py{t\_cmp} of shape \py{[M, m]}. This compressed tensor is later used to build the Infoset representation in Section~\ref{sec:infosettensor}. During GT construction, all resulting \py{t\_cmp} tensors are stored, and it becomes straightforward to mask information not observable by the acting player through simple masking operations. Next, we describe how \py{t\_cmp} is constructed and explain why this construction uniquely encodes a GT layer, given the current In-Play tensor \py{t\_inp}. See CodeSnippet~\ref{cs:compressedgametensor}.

\begin{algorithm}
\caption{Compressed Game Tensor}
\label{cs:compressedgametensor}
\begin{algorithmic}[1]
\State \textbf{Input:} \py{t\_gme}
\State \py{t\_cmp} $\gets$ \py{t\_gme[:,1,:] - t\_gme[:,2,:]}
\State \py{t\_lpm} $\gets$ \py{logical\_and(t\_gme[:,0,:] == 0, (t\_cmp != 0) \& (t\_cmp.abs() < 5))}
\State \py{t\_cmp[t\_lpm]} $\gets$ \py{110 + t\_cmp[t\_lpm]}

\vspace{1mm}

\State \py{t\_cmp[logical\_and(t\_gme[:,0,:] == 3, t\_cmp == 0)]} $\gets$ \py{110}
\State \py{t\_cmp[logical\_and(t\_gme[:,0,:] == 1, t\_cmp == 0)]} $\gets$ \py{100}
\State \py{t\_cmp[logical\_and(t\_gme[:,0,:] == 2, t\_cmp == 0)]} $\gets$ \py{105}
\end{algorithmic}
\end{algorithm}

In Table~\ref{table:compressedgametensor}, we summarize all possible values that can appear in the compressed tensor \py{t\_cmp}. Each card can have up to two \emph{legs} in a round. The first leg corresponds to when the card is \emph{laid} into the pool, and the second leg corresponds to when it is \emph{picked} from the pool. A card may have only one leg in the current layer of the game tensor—this occurs when a player lays the card into the pool, but it has not yet been picked by any player. Alternatively, a card may have no legs in the current round; this happens when the card was already in the pool at the start of the round or is still held by one of the players. By design of the game tensor \py{t\_gme}, we can characterize these situations precisely:

\begin{table}[H]
\centering
\scriptsize
\caption{Leg Status Conditions for Cards}
\label{table:card_leg_conditions}
\begin{tabular}{
>{\centering\arraybackslash}p{2cm}
>{\centering\arraybackslash}p{4cm}
>{\centering\arraybackslash}p{8cm}
}
\toprule
\textbf{No Legs} & \textbf{One Leg} & \textbf{Both Legs in Same Turn} \\
\midrule
\py{t\_cmp==0} & \py{t\_gme[:,0,:]==3 \& t\_cmp!=0} & \py{t\_gme[:,0,:]==0 \& (t\_cmp != 0) \& (t\_cmp.abs() < 5)} \\
\bottomrule
\end{tabular}
\end{table}

\begin{table}[H]
\centering
\scriptsize
\caption{Compressed Game Tensor Values in \py{t\_cmp}}
\label{table:compressedgametensor}
\begin{tabular}{c c c l}
\toprule
\textbf{Value} & \textbf{First Leg} & \textbf{Second Leg} & \textbf{Description} \\
\midrule
\py{100} & — & — & Held by Alex \\
\py{105} & — & — & Held by Bob \\
\py{110} & — & — & Was already in the pool \\
\midrule
\py{111,112,113,114} & Alex laid at $\py{i\_trn=0:3}$ & Alex picked at same turn & Both legs in same turn by Alex \\
\py{109,108,107,106} & Bob laid at $\py{i\_trn=0:3}$ & Bob picked at same turn & Both legs in same turn by Bob \\
\midrule
\py{1,2,3,4} & Alex laid at $\py{i\_trn=0:3}$ & — & Not yet picked \\
\py{-1,-2,-3,-4} & Bob laid at $\py{i\_trn=0:3}$ & — & Not yet picked \\
\midrule
\py{-9,-19,-29,-39} & Alex laid at \quad$\py{i\_trn=0}$ & Bob picked at $\py{i\_trn=0:3}$ & \\
\py{-18,-28,-38} & Alex laid at \quad$\py{i\_trn=1}$ & Bob picked at $\py{i\_trn=1:3}$ & \\
\py{-27,-37} & Alex laid at \quad$\py{i\_trn=2}$ & Bob picked at $\py{i\_trn=2:3}$ & \\
\py{21,31,41} & Alex laid at \quad$\py{i\_trn=0}$ & Alex picked at $\py{i\_trn=1:3}$ & \\
\py{32,42} & Alex laid at \quad$\py{i\_trn=1}$ & Alex picked at $\py{i\_trn=2:3}$ & \\
\py{43} & Alex laid at \quad$\py{i\_trn=2}$ & Alex picked at \quad$\py{i\_trn=3}$ & \\
\midrule
\py{19,29,39} & Bob laid at \quad$\py{i\_trn=0}$ & Alex picked at $\py{i\_trn=1:3}$ & \\
\py{28,38} & Bob laid at \quad$\py{i\_trn=1}$ & Alex picked at $\py{i\_trn=2:3}$ & \\
\py{37} & Bob laid at \quad$\py{i\_trn=2}$ & Alex picked at \quad$\py{i\_trn=3}$ & \\
\py{-21,-31,-41} & Bob laid at \quad$\py{i\_trn=0}$ & Bob picked at $\py{i\_trn=1:3}$ & \\
\py{-32,-42} & Bob laid at \quad$\py{i\_trn=1}$ & Bob picked at $\py{i\_trn=2:3}$ & \\
\py{-43} & Bob laid at \quad$\py{i\_trn=2}$ & Bob picked at \quad$\py{i\_trn=3}$ & \\
\bottomrule
\end{tabular}
\end{table}

\subsection{Score Tensors}\label{sec:scoretensor}

There are two types of score tensors: those that store the score accumulated within a single round, called the \emph{running-score tensor} \py{t\_rus}, and those that are passed along edges in GT, referred to as the \emph{score tensor} \py{t\_scr}.  The tensor \py{t\_rus} is initialized to zero at the beginning of each round and is updated after every turn. At the end of the round, the final value of \py{t\_rus} is used to update \py{t\_scr}. Tables~\ref{table:scoretensorcols} and \ref{table:runningscorecols} outline the meaning of each index in the score tensors \py{t\_scr} and \py{t\_rus}, respectively. Note that the running-score tensor \py{t\_rus} includes additional components relevant to a single round.

\begin{table}[H]
    \centering
    \scriptsize
    \caption{Column definitions for the score tensor}
    \begin{tabular}{
        >{\centering\arraybackslash}p{2.8cm}
        >{\centering\arraybackslash}p{2.8cm}
        >{\centering\arraybackslash}p{2.8cm}
        >{\centering\arraybackslash}p{2.8cm}
        }
        \toprule
        \textbf{Alex Club} & \textbf{Bob Club} & \textbf{Point Difference} & \textbf{7-Clubs Bonus} \\
        \midrule
        \py{0} & \py{1} & \py{2} & \py{3} \\
        \bottomrule
    \end{tabular}
    \label{table:scoretensorcols}
\end{table}

\begin{table}[H]
    \centering
    \scriptsize
    \caption{Column definitions for the running-score tensor}
    \begin{tabular}{
        >{\centering\arraybackslash}p{1.8cm}
        >{\centering\arraybackslash}p{1.8cm}
        >{\centering\arraybackslash}p{1.8cm}
        >{\centering\arraybackslash}p{1.8cm}
        >{\centering\arraybackslash}p{1.8cm}
        >{\centering\arraybackslash}p{1.8cm}
        >{\centering\arraybackslash}p{1.8cm}
        }
        \toprule
        \textbf{Alex Club} & \textbf{Bob Club} & \textbf{Last Picker} & \textbf{Alex Points} & \textbf{Alex Sur} & \textbf{Bob Points} & \textbf{Bob Sur} \\
\midrule
\py{0} & \py{1} & \py{2} & \py{3} & \py{4} & \py{5} & \py{6} \\
\bottomrule
\end{tabular}
\label{table:runningscorecols}
\end{table}
We need to clarify an important point about the score tensor \py{t\_scr}. After each round, once the number of Clubs collected by either Alex or Bob reaches 7, we update the last column---the 7-Clubs Bonus---to reflect that one of the players has achieved the bonus. Specifically, the value becomes 1 if Alex reaches 7 Clubs, or 2 if Bob does. At the same time, the counts in the first two columns, which track Clubs for Alex and Bob respectively, are reset to 0. 

For example, the score vector \py{[3, 8, 2, 0]} is not a valid state; it will be converted to \py{[0, 0, 2, 2]}, indicating that Bob has earned the 7-Clubs bonus. At this point, accumulating additional Clubs is irrelevant to the 7-Clubs bonus and is therefore not tracked. The primary advantage of this design choice is that it reduces the number of possible edges (and hence the size of the FGT), resulting in significant savings in both computation and memory. 

We omit the procedure for updating the running-score tensor and refer the reader to the Github repository.



\subsection{In-Hand Updates}\label{sec:inhandupdates}
In this section, we explain how in-hand updates are performed. A key operation required throughout is a generalized version of \py{torch.repeat\_interleave}, which we refer to as \textbf{RepeatBlocks}. Unlike \py{torch.repeat\_interleave}, which repeats individual elements, RepeatBlocks operates on entire contiguous blocks of elements and repeats them as a whole. This operation is detailed in Appendix~\ref{sec:appendix}.

We now explain how FGT is updated within each round. Two tensors are updated during this process: the FGT tensor \py{t\_fgm} and the Edge tensor \py{t\_edg}, which captures the connections between nodes across successive layers of FGT.

The FGT tensor \py{t\_fgm} is used to track which scores accumulated from previous rounds (stored in \py{t\_scr}) are linked to each node of the GT tensor \py{t\_gme}. A row entry \py{[g, s]} in \py{t\_fgm} indicates that node \py{t\_gme[g]} inherits the score \py{t\_scr[s]} from a previous round.  See Figure~\ref{fig:t_fgm_example} for an example consistent with the setting illustrated on the left-hand side of Figure~\ref{fig:Unfolding Process}.

\begin{figure}[h]
  \centering
\[
\py{t\_scr} = 
\begin{bmatrix}
\tikz\fill[green] (0,0) rectangle (0.2,0.2); \\
\tikz\fill[red] (0,0) rectangle (0.2,0.2); \\
\tikz\fill[fill=blue!50] (0,0) rectangle (0.2,0.2); \\
\tikz\fill[orange] (0,0) rectangle (0.2,0.2); \\
\end{bmatrix},
\qquad
\py{t\_fgm} = \begin{bmatrix}
\py{0} & \tikz\fill[green] (0,0) rectangle (0.2,0.2); \\
\py{0} & \tikz\fill[red] (0,0) rectangle (0.2,0.2); \\
\py{0} & \tikz\fill[fill=blue!50] (0,0) rectangle (0.2,0.2); \\
\py{1} & \tikz\fill[orange] (0,0) rectangle (0.2,0.2); \\
\py{1} & \tikz\fill[fill=blue!50] (0,0) rectangle (0.2,0.2); 
\end{bmatrix}
\]
\caption{An example illustrating how the FGT tensor \py{t\_fgm} associates nodes in the GT with entries in the Score Tensor \py{t\_scr}. Colors are used to indicate rows of \py{t\_scr}, consistent with Figure~\ref{fig:gametreegeneration}. Each row of \py{t\_scr} is represented by a tensor of size~4, as described in Table~\ref{table:scoretensorcols}.}
  \label{fig:t_fgm_example}
\end{figure}

\begin{figure}[h]
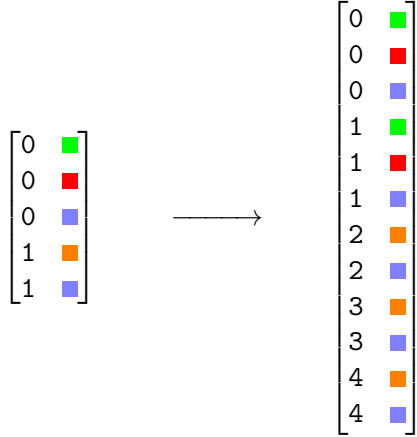

  \centering
\[
\begin{bmatrix}
\py{0} & \tikz\fill[green] (0,0) rectangle (0.2,0.2); \\
\py{0} & \tikz\fill[red] (0,0) rectangle (0.2,0.2); \\
\py{0} & \tikz\fill[fill=blue!50] (0,0) rectangle (0.2,0.2); \\
\py{1} & \tikz\fill[orange] (0,0) rectangle (0.2,0.2); \\
\py{1} & \tikz\fill[fill=blue!50] (0,0) rectangle (0.2,0.2); 
\end{bmatrix}
\;
\qquad
\xrightarrow{\hspace{1cm}}
\qquad
\begin{bmatrix}
\py{0} & \tikz\fill[green] (0,0) rectangle (0.2,0.2); \\
\py{0} & \tikz\fill[red] (0,0) rectangle (0.2,0.2); \\
\py{0} & \tikz\fill[fill=blue!50] (0,0) rectangle (0.2,0.2); \\
\py{1} & \tikz\fill[green] (0,0) rectangle (0.2,0.2); \\
\py{1} & \tikz\fill[red] (0,0) rectangle (0.2,0.2); \\
\py{1} & \tikz\fill[fill=blue!50] (0,0) rectangle (0.2,0.2); \\
\py{2} & \tikz\fill[orange] (0,0) rectangle (0.2,0.2); \\
\py{2} & \tikz\fill[fill=blue!50] (0,0) rectangle (0.2,0.2); \\
\py{3} & \tikz\fill[orange] (0,0) rectangle (0.2,0.2); \\
\py{3} & \tikz\fill[fill=blue!50] (0,0) rectangle (0.2,0.2); \\
\py{4} & \tikz\fill[orange] (0,0) rectangle (0.2,0.2); \\
\py{4} & \tikz\fill[fill=blue!50] (0,0) rectangle (0.2,0.2);
\end{bmatrix}
\]
  \caption{An illustration of how \py{t\_fgm} is updated.}
  \label{fig:t_fgm}
\end{figure}

Figure~\ref{fig:Unfolding Process} illustrates how \py{t\_edg} is constructed. On the left-hand side, we show two parent nodes—one with 2 child nodes and 3 inherited scores, and the other with 3 child nodes and 2 inherited scores. The right-hand side of the figure illustrates the corresponding structure within FGT. Here, colors denote inherited scores from the previous round, and numbers represent the row indices from the current GT tensor \py{t\_gme}. An edge is drawn when the score colors match and the nodes are connected in the folded GT. In this example, the tensor
\[
\py{t\_edg = [0,1,2,0,1,2,3,4,3,4,3,4]}
\]
records the parent indices in FGT. The counts of inherited scores and available actions per parent are given by \py{c\_scr = [3,2]} and \py{t\_brf = [2,3]}, respectively. This structure can be generated compactly by the RepeatBlocks operator as follows:
\[
\py{t\_edg} \leftarrow \py{arange(t\_fgm.shape[0])} \otimes_{\py{c\_scr}} \py{t\_brf}.
\]
We will next explain how the FGT tensor \py{t\_fgm} is updated. Figure~\ref{fig:t_fgm} shows the \py{t\_fgm} tensors before and after the update, corresponding to the setting of Figure~\ref{fig:Unfolding Process}. The update to \py{t\_fgm} is performed in two phases: first, we update \py{t\_fgm[:,1]}; then, we update \py{t\_gme} and \py{c\_scr} before proceeding to update \py{t\_fgm[:,0]}. We have 
\[
    \py{t\_fgm[:,1]} \leftarrow 
    \begin{bmatrix}
        \tikz[baseline] \fill[green] (0,0) rectangle (0.2,0.2); &
        \tikz[baseline] \fill[red] (0,0) rectangle (0.2,0.2); &
        \tikz[baseline] \fill[fill=blue!50] (0,0) rectangle (0.2,0.2); &
        \tikz[baseline] \fill[orange] (0,0) rectangle (0.2,0.2); &
        \tikz[baseline] \fill[fill=blue!50] (0,0) rectangle (0.2,0.2); 
    \end{bmatrix}
    \otimes_{\py{[3,2]}}\py{[2,3]}
    \]
    where the first tensor on the right-hand side is \py{t\_fgm[:,1]}. The first three colors are repeated twice on the right-hand side of Figure~\ref{fig:t_fgm} because \py{t\_brf[0] = 2}, and the last two are repeated three times because \py{t\_brf[1] = 3}. Therefore
  \[
  \py{t\_fgm[:,1]} \leftarrow \py{t\_fgm[:,1]}\otimes_{\py{c\_scr}} \py{t\_brf}
  \]
  Furthermore, 
  \[
  \py{t\_fgm[:,0]} \leftarrow \py{[0,1,2,3,4]$\otimes$[3,3,2,2,2]} 
  \]
  In words, each node in GT's second layer in the left-hand side of Figure~\ref{fig:Unfolding Process} is repeated as many times as the number of scores it inherited from the previous round. Namely,
  \[
  \py{t\_fgm[:,0]}\leftarrow \py{arange(t\_gme.shape[0])$\otimes$c\_scr}
  \]
  Both \py{t\_gme} and \py{c\_scr} are updated before the equation above. 
                
\subsection{Between-Hand Updates}\label{sec:bethand}
\begin{figure}
\centering
\scalebox{0.5}{
    \begin{tikzpicture}

             \drawboxwitharrows{0}{blue!50/0.4, green/0.05}{dummy}{0}  {-5cm,  -28cm} {none}
      \drawboxwitharrows{0}{blue!50/0.4, orange/0.05}{dummy}{1}        {-4cm,   -28cm}{none}
      \drawboxwitharrows{0}{green/-0.2, blue!50/0.4}{dummy}{2}         {-3cm,   -28cm}{none}
        \node[rotate=0] at (3cm, -28cm) {\dots\dots\dots\dots\dots\dots\dots\dots\dots\dots\dots\dots};   
      \drawboxwitharrows{0}{black/-0.2, blue!50/0.4}{dummy}{-3}     {9cm,   -28cm} {none}
      \drawboxwitharrows{0}{red/-0.2, blue!50/0.4}{dummy}{-2}        {10cm,   -28cm}{none}
      \drawboxwitharrows{0}{blue!50/0.4, red/0.05}{dummy}{-1}        {11cm,   -28cm}{none}

         \draw[blue!50, thick, decoration={markings, mark=at position 0.5 with {\arrow{>}}}, postaction={decorate}] (-5, -28.3) to[out=-70, in=60] (5, -30.7); 
         \draw[red, thick, decoration={markings, mark=at position 0.5 with {\arrow{>}}}, postaction={decorate}] (-5, -28.3) to[out=-90, in=60] (5, -30.7); 
         \draw[green, thick, decoration={markings, mark=at position 0.5 with {\arrow{>}}}, postaction={decorate}] (-4, -28.3) to[out=-80, in=60] (1, -30.7); 
         \draw[orange, thick, decoration={markings, mark=at position 0.5 with {\arrow{>}}}, postaction={decorate}] (-4, -28.3) to[out=-60, in=60] (1, -30.7); 
         \draw[red, thick, decoration={markings, mark=at position 0.5 with {\arrow{>}}}, postaction={decorate}] (-3, -28.3) to[out=-80, in=60] (2, -30.7); 
         \draw[black, thick, decoration={markings, mark=at position 0.5 with {\arrow{>}}}, postaction={decorate}] (-3, -28.3) to[out=-60, in=60] (2, -30.7); 
         \draw[green, thick, decoration={markings, mark=at position 0.5 with {\arrow{>}}}, postaction={decorate}] (-4, -28.3) to[out=-80, in=60] (1, -30.7); 
         \draw[orange, thick, decoration={markings, mark=at position 0.5 with {\arrow{>}}}, postaction={decorate}] (-4, -28.3) to[out=-60, in=60] (1, -30.7); 
         \draw[black, thick, decoration={markings, mark=at position 0.5 with {\arrow{>}}}, postaction={decorate}] (10, -28.3) to[out=-80, in=60] (4, -30.7); 
         \draw[orange, thick, decoration={markings, mark=at position 0.5 with {\arrow{>}}}, postaction={decorate}] (10, -28.3) to[out=-60, in=60] (4, -30.7); 
         \draw[green, thick, decoration={markings, mark=at position 0.5 with {\arrow{>}}}, postaction={decorate}] (9, -28.3) to[out=-80, in=60] (3, -30.7); 
         \draw[red, thick, decoration={markings, mark=at position 0.5 with {\arrow{>}}}, postaction={decorate}] (9, -28.3) to[out=-60, in=60] (3, -30.7); 
         \draw[green, thick, decoration={markings, mark=at position 0.5 with {\arrow{>}}}, postaction={decorate}] (11, -28.3) to[out=-80, in=60] (3, -30.7); 
         \draw[red, thick, decoration={markings, mark=at position 0.5 with {\arrow{>}}}, postaction={decorate}] (11, -28.3) to[out=-60, in=60] (3, -30.7);

         \draw[green, thick, decoration={markings, mark=at position 0.5 with {\arrow{>}}}, postaction={decorate}] (5, -28.3) to[out=-80, in=60] (6, -30.7); 
         \draw[red, thick, decoration={markings, mark=at position 0.5 with {\arrow{>}}}, postaction={decorate}] (5, -28.3) to[out=-60, in=60] (6, -30.7);

         \draw[blue!50, thick, decoration={markings, mark=at position 0.5 with {\arrow{>}}}, postaction={decorate}] (2, -28.3) to[out=-80, in=60] (7, -30.7); 
         \draw[red, thick, decoration={markings, mark=at position 0.5 with {\arrow{>}}}, postaction={decorate}] (2, -28.3) to[out=-60, in=60] (7, -30.7); 
         
         \draw[blue!50, thick, decoration={markings, mark=at position 0.5 with {\arrow{>}}}, postaction={decorate}] (-2, -28.3) to[out=-80, in=60] (7, -30.7); 
         \draw[green, thick, decoration={markings, mark=at position 0.5 with {\arrow{>}}}, postaction={decorate}] (-2, -28.3) to[out=-60, in=60] (7, -30.7);

      \drawboxwitharrows{0}{}{dummy}{0}                     {1cm,  -31cm} {none}
      \drawboxwitharrows{0}{}{dummy}{1}                     {2cm,  -31cm} {none}
      \drawboxwitharrows{0}{}{dummy}{2}                     {3cm,  -31cm} {none}
      \drawboxwitharrows{0}{}{dummy}{3}                     {4cm,   -31cm}{none}
      \drawboxwitharrows{0}{}{dummy}{4}                     {5cm,   -31cm}{none}
      \drawboxwitharrows{0}{}{dummy}{5}                     {6cm,   -31cm}{none}
      \drawboxwitharrows{0}{}{dummy}{6}                     {7cm,   -31cm}{none}

    \end{tikzpicture}
    }
\caption{An illustration of between-hand processing. Note that the number of incoming edges to each terminal node in the last round is equal to the number of edges going from that node to the root-level node in the next round. However, the color may change, as the corresponding running-score tensor is added to each score inherited from previous rounds.}
      \label{fig00:bethand}
\end{figure}
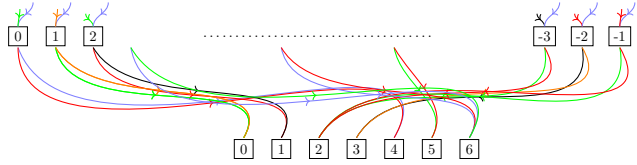

In this section, we introduce the processing framework used at the beginning of each hand. Figure~\ref{fig00:bethand} illustrates this process. By the end of each hand, we obtain a set of GT nodes, each associated with one or more inherited scores from previous rounds. Our goal is to construct the root-level nodes of GT's next round, along with their inherited scores. To begin, we identify the unique game states and running-score tensors at the terminal level of the current round. This is achieved via the following operation:
\begin{align*}
\py{t\_gme}, \py{t\_gnk} &= \py{unique(t\_gme, dim=0, sorted=False, return\_inverse=True)} \\
\py{t\_rus}, \py{t\_rnk} &= \py{unique(t\_rus, dim=0, sorted=False, return\_inverse=True)}
\end{align*}
We first update the second column of FGT tensor \py{t\_fgm}, namely \py{t\_fgm[:,1]}, which encodes the inherited scores. To propagate scores correctly, we take the score components from the previous round, stored in \py{t\_fgm[:,1]}, and add the scores earned during the current hand, recorded in \py{t\_rus}. To this end, we represent each player's total score—comprising both the inherited and running components—as follows:
\begin{center}
\py{t\_prs} $\gets$ \py{cat([t\_fgm[:,1], t\_rnk[t\_fgm[:,0]]], dim=1)}
\end{center}
Concatenate these two components to identify the unique score combinations:
\begin{align*}
\py{t\_prs, t\_pid} &\gets \py{unique(t\_prs, dim=0, sorted = False, return\_inverse=True)}
\end{align*}
Each unique pair in \py{t\_prs} represents a total score passed to the next round, and is computed by summing the individual components:
\begin{center}
\py{t\_scr} $\gets$ \py{t\_scr[t\_prs[:,0]]+t\_rus[t\_prs[:,1]]}
\end{center}
We next eliminate duplicate score values using another call to \py{unique}:
\begin{align*}
\py{t\_scr}, \py{t\_fid} &\gets \py{unique(t\_scr, dim=0, soretd=False, return\_inverse=True)}
\end{align*}
This gives us enough ingredients to update \py{t\_fgm[:,1]} as follows:
\begin{align*}
\py{t\_fgm[:,1]} &\gets \py{t\_fid[t\_pid]}
\end{align*}
Now we update the first column of the FGT \py{t\_fgm[:,0]} as follows:
\[
\py{t\_fgm[:,0]} \leftarrow \py{t\_gnk[t\_fgm[:,0]]}
\]
Note that this is an intermediate step: the final version of \py{t\_fgm} will be updated later in the process. Next, we determine how scores are passed from one hand to the next. That is, we construct the new \py{t\_fgm[:,1]} tensor for the upcoming round. Finally, we find unique rows in \py{t\_fgm} as follows:
\begin{align*}
  \py{t\_fgm, t\_lnk }& \gets \py{ unique(t\_fgm,dim=0, sorted=False, return\_inverse=True)}
\end{align*}
CodeSnippet \ref{cs:bethand} summarizes the steps discussed above.

\begin{figure}[h]
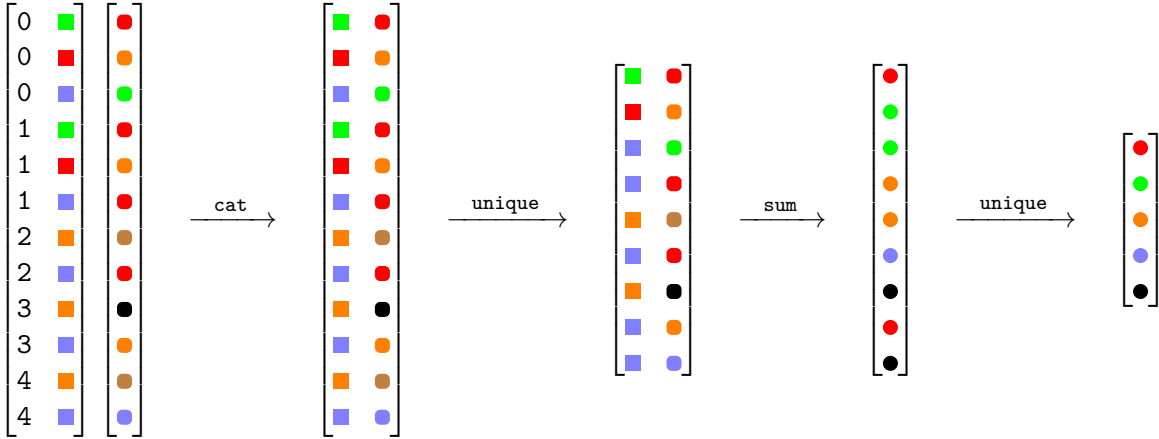

  \centering
\[
\begin{bmatrix}
\py{0} & \tikz\fill[green] (0,0) rectangle (0.2,0.2); \\
\py{0} & \tikz\fill[red] (0,0) rectangle (0.2,0.2); \\
\py{0} & \tikz\fill[fill=blue!50] (0,0) rectangle (0.2,0.2); \\
\py{1} & \tikz\fill[green] (0,0) rectangle (0.2,0.2); \\
\py{1} & \tikz\fill[red] (0,0) rectangle (0.2,0.2); \\
\py{1} & \tikz\fill[fill=blue!50] (0,0) rectangle (0.2,0.2); \\
\py{2} & \tikz\fill[orange] (0,0) rectangle (0.2,0.2); \\
\py{2} & \tikz\fill[fill=blue!50] (0,0) rectangle (0.2,0.2); \\
\py{3} & \tikz\fill[orange] (0,0) rectangle (0.2,0.2); \\
\py{3} & \tikz\fill[fill=blue!50] (0,0) rectangle (0.2,0.2); \\
\py{4} & \tikz\fill[orange] (0,0) rectangle (0.2,0.2); \\
\py{4} & \tikz\fill[fill=blue!50] (0,0) rectangle (0.2,0.2);
\end{bmatrix}
\begin{bmatrix}
\tikz\fill[red, rounded corners=2pt]               (0,0) rectangle (0.2,0.2); \\ 
\tikz\fill[orange, rounded corners=2pt]               (0,0) rectangle (0.2,0.2); \\ 
\tikz\fill[green, rounded corners=2pt]               (0,0) rectangle (0.2,0.2); \\ 
\tikz\fill[red, rounded corners=2pt]               (0,0) rectangle (0.2,0.2); \\ 
\tikz\fill[orange, rounded corners=2pt]               (0,0) rectangle (0.2,0.2); \\ 
\tikz\fill[red, rounded corners=2pt]               (0,0) rectangle (0.2,0.2); \\ 
\tikz\fill[brown, rounded corners=2pt]               (0,0) rectangle (0.2,0.2); \\ 
\tikz\fill[red, rounded corners=2pt]               (0,0) rectangle (0.2,0.2); \\ 
\tikz\fill[black, rounded corners=2pt]               (0,0) rectangle (0.2,0.2); \\ 
\tikz\fill[orange, rounded corners=2pt]               (0,0) rectangle (0.2,0.2); \\ 
\tikz\fill[fill=brown, rounded corners=2pt]               (0,0) rectangle (0.2,0.2); \\ 
\tikz\fill[fill=blue!50, rounded corners=2pt]               (0,0) rectangle (0.2,0.2); \\ 
\end{bmatrix}
\quad
\xrightarrow{\hspace{0.25cm}\py{cat}\hspace{0.25cm}}
\quad
\begin{bmatrix}
\tikz\fill[green] (0,0) rectangle (0.2,0.2);& \tikz\fill[red, rounded corners=2pt]               (0,0) rectangle (0.2,0.2); \\ 
\tikz\fill[red] (0,0) rectangle (0.2,0.2);& \tikz\fill[orange, rounded corners=2pt]               (0,0) rectangle (0.2,0.2); \\ 
\tikz\fill[fill=blue!50] (0,0) rectangle (0.2,0.2);& \tikz\fill[green, rounded corners=2pt]               (0,0) rectangle (0.2,0.2); \\ 
\tikz\fill[green] (0,0) rectangle (0.2,0.2);& \tikz\fill[red, rounded corners=2pt]               (0,0) rectangle (0.2,0.2); \\ 
\tikz\fill[red] (0,0) rectangle (0.2,0.2);& \tikz\fill[orange, rounded corners=2pt]               (0,0) rectangle (0.2,0.2); \\ 
\tikz\fill[fill=blue!50] (0,0) rectangle (0.2,0.2);& \tikz\fill[red, rounded corners=2pt]               (0,0) rectangle (0.2,0.2); \\ 
\tikz\fill[orange] (0,0) rectangle (0.2,0.2);& \tikz\fill[brown, rounded corners=2pt]               (0,0) rectangle (0.2,0.2); \\ 
\tikz\fill[fill=blue!50] (0,0) rectangle (0.2,0.2);& \tikz\fill[red, rounded corners=2pt]               (0,0) rectangle (0.2,0.2); \\ 
\tikz\fill[orange] (0,0) rectangle (0.2,0.2);& \tikz\fill[black, rounded corners=2pt]               (0,0) rectangle (0.2,0.2); \\ 
\tikz\fill[fill=blue!50] (0,0) rectangle (0.2,0.2);& \tikz\fill[orange, rounded corners=2pt]               (0,0) rectangle (0.2,0.2); \\ 
\tikz\fill[orange] (0,0) rectangle (0.2,0.2);& \tikz\fill[fill=brown, rounded corners=2pt]               (0,0) rectangle (0.2,0.2); \\ 
\tikz\fill[fill=blue!50] (0,0) rectangle (0.2,0.2); & \tikz\fill[fill=blue!50, rounded corners=2pt]               (0,0) rectangle (0.2,0.2); \\
\end{bmatrix}
\quad
\xrightarrow{\hspace{0.25cm}\py{unique}\hspace{0.25cm}}
\quad
\begin{bmatrix}
\tikz\fill[green] (0,0) rectangle (0.2,0.2);& \tikz\fill[red, rounded corners=2pt]               (0,0) rectangle (0.2,0.2); \\ 
\tikz\fill[red] (0,0) rectangle (0.2,0.2);& \tikz\fill[orange, rounded corners=2pt]               (0,0) rectangle (0.2,0.2); \\ 
\tikz\fill[fill=blue!50] (0,0) rectangle (0.2,0.2);& \tikz\fill[green, rounded corners=2pt]               (0,0) rectangle (0.2,0.2); \\ 
\tikz\fill[fill=blue!50] (0,0) rectangle (0.2,0.2);& \tikz\fill[red, rounded corners=2pt]               (0,0) rectangle (0.2,0.2); \\ 
\tikz\fill[orange] (0,0) rectangle (0.2,0.2);& \tikz\fill[brown, rounded corners=2pt]               (0,0) rectangle (0.2,0.2); \\ 
\tikz\fill[fill=blue!50] (0,0) rectangle (0.2,0.2);& \tikz\fill[red, rounded corners=2pt]               (0,0) rectangle (0.2,0.2); \\ 
\tikz\fill[orange] (0,0) rectangle (0.2,0.2);& \tikz\fill[black, rounded corners=2pt]               (0,0) rectangle (0.2,0.2); \\ 
\tikz\fill[fill=blue!50] (0,0) rectangle (0.2,0.2);& \tikz\fill[orange, rounded corners=2pt]               (0,0) rectangle (0.2,0.2); \\ 
\tikz\fill[fill=blue!50] (0,0) rectangle (0.2,0.2); & \tikz\fill[fill=blue!50, rounded corners=2pt]               (0,0) rectangle (0.2,0.2); \\
\end{bmatrix}
\quad
\xrightarrow{\hspace{0.25cm}\py{sum}\hspace{0.25cm}}
\quad
\begin{bmatrix}
\tikz\fill[red] (0,0) circle (0.1); \\ 
\tikz\fill[green] (0,0) circle (0.1);  \\ 
\tikz\fill[green] (0,0) circle (0.1);  \\ 
\tikz\fill[orange] (0,0) circle (0.1); \\ 
\tikz\fill[orange] (0,0) circle (0.1); \\ 
\tikz\fill[fill=blue!50] (0,0) circle (0.1);  \\ 
\tikz\fill[black] (0,0) circle (0.1); \\ 
\tikz\fill[red] (0,0) circle (0.1); \\ 
\tikz\fill[black] (0,0) circle (0.1);  \\
\end{bmatrix}
\quad
\xrightarrow{\hspace{0.25cm}\py{unique}\hspace{0.25cm}}
\quad
\begin{bmatrix}
\tikz\fill[red] (0,0) circle (0.1); \\ 
\tikz\fill[green] (0,0) circle (0.1);  \\ 
\tikz\fill[orange] (0,0) circle (0.1); \\ 
\tikz\fill[fill=blue!50] (0,0) circle (0.1);  \\ 
\tikz\fill[black] (0,0) circle (0.1); \\ 
\end{bmatrix}
\]
  \caption{An illustration of between-hand updates for \py{t\_fgm}.}
  \label{fig01:bethand}
\end{figure}

\begin{algorithm}
\caption{BetweenHands}
\label{cs:bethand}
\begin{algorithmic}[1]
\State Input \py{t\_gme,t\_fgm,t\_scr,t\_dck}
\State \py{t\_gme, t\_gnk }$\gets$\py{ unique(t\_gme,dim=0)}
\State \py{t\_rus, t\_rnk }$\gets$\py{ unique(t\_rus,dim=0)}\label{line:trus}
\vspace{2mm}
\State \py{t\_prs} $\gets$ \py{cat([t\_fgm[:,1],t\_rnk[t\_fgm[:,0]]], dim=1)}
\State \py{t\_prs,t\_pid }$\gets$\py{ unique(t\_prs,dim=0)}
\vspace{2mm}
\State \py{t\_scr} $\gets$ \py{t\_scr[t\_prs[:,0]]+t\_rus[t\_prs[:,1]]}
\State \py{t\_scr, t\_fid }$\gets$\py{ unique(t\_scr, dim=0)}
\vspace{2mm}
\State \py{t\_fgm[:,1]} $\gets$ \py{t\_fid[t\_pid]}
\State \py{t\_fgm[:,0]} $\gets$ \py{t\_gnk[t\_fgm[:,0]]}
\vspace{2mm}
\State \py{t\_fgm, t\_lnk }$\gets$\py{ unique(t\_fgm,dim=0)}\label{line:bethandlast}
\Statex \textcolor{green!50!black}{\scriptsize\py{\# Note: All the unique operations here consider sorted=False,return\_inverse=True}}
\end{algorithmic}
\end{algorithm}


\subsection{Infoset Tensors}\label{sec:infosettensor}
In this section, we represent FGT's nodes using the \py{int8} Infoset Tensor \py{t\_inf}, where each row has shape \py{[58]}. This tensor encodes all information available at that point in the game tree and can be easily adapted to mask any information not observable by the acting player. This design serves two primary purposes: first, to ensure memory efficiency; and second, to provide a compact representation of both the game state and associated action information. These representations are later used to train a tree-based model to approximate the strategy profile, which is subsequently used in self-play simulations, as described in Section~\ref{sec:experimental}. It is important to emphasize that these infoset tensors are used only after the Nash equilibrium has been learned using the CFR algorithm.  Tensor \py{t\_inf} consists of three distinct parts, summarized in Table~\ref{table:tinf_structure} and CodeSnippet~\ref{cs:infosettensor} explains its construction process.

\begin{table}[H]
    \centering
    \scriptsize
    \caption{Index structure of the \py{t\_inf} tensor}
    \begin{tabular}{
        >{\centering\arraybackslash}p{2.5cm}
        >{\centering\arraybackslash}p{3.5cm}
        >{\centering\arraybackslash}p{8.5cm}
        }
        \toprule
        \textbf{Index Range} & \textbf{Section} & \textbf{Description} \\
        \midrule
        \py{0:52} & Cards & Status/history of each card \\
        \py{52:55} & Score context & Inherited scores such as club points and point differential \\
        \py{55:58} & Metadata & Round index, turn counter, and current player indicator \\
        \bottomrule
    \end{tabular}
    \label{table:tinf_structure}
\end{table}

\begin{algorithm}
\caption{Infoset Tensor}
\label{cs:infosettensor}
\begin{algorithmic}[1]
\State \py{t\_inf} $\gets$ \py{zeros((Q, 58))}  \textcolor{green!50!black}{\scriptsize  \# \py{Q=t\_fgm.shape[0]}}
\vspace{3 mm}
\State \py{t\_inf[:, t\_inp]}$\gets$\py{t\_cmp[t\_fgm[:,0]]}
\State \py{t\_inf[:,52:55]}$\gets$\py{t\_scr[t\_fgm[:,1]]}
\vspace{3mm}
\State \py{t\_inf[logical\_and(t\_dlt==1,t\_inf==0)]}$\gets$\py{-127}\textcolor{green!50!black}{\scriptsize\py{\# for any dealt card represented in t\_dlt \& not present in t\_gme, -127 is assigned i.e., card is collected in past rounds.}}
\State \py{t\_inf[:,55:58]}$\gets$\py{tensor([i\_hnd,i\_trn,i\_ply])}

\end{algorithmic}
\end{algorithm}




\input{exsampling}
\newpage 
\subsection{CFR Algorithm}
In this section, we provide details on the PyTorch implementation of CFR for Pasur, following the framework outlined in the previous sections. 

Our approach is as follows: for each round, we compute the average utility at the root-level nodes of the FGT. This utility is then propagated backward to the preceding round, where CFR is applied to compute the utilities for that round’s root-level nodes. While this process is underway within a round, we simultaneously compute the average strategy profile (Equation \eqref{eq:avsigma}) for that round.

It is important to emphasize that when working within a round, the reach probabilities for the root-level nodes are all set to 1. Since Pasur consists of six rounds, we apply the CFR algorithm six times—once per round. The utility for the final round is computed directly by calculating the final score. Notably, the running-score tensor \py{t\_rus} is added to the utility of the final-layer nodes before applying CFR.

We treat utility values with equal weights, based on the logic that each unit of utility represents a point earned under the Nash Equilibrium at that terminal node within the round. Because \py{t\_rus} reflects the points accumulated during that round, it is appropriate to add it linearly to the computed utility.

This round-by-round implementation is both more accurate and memory-efficient. First, by computing utilities one round at a time rather than propagating utilities from terminal nodes all the way to the top of the full game tree, we improve numerical stability. This localized computation reduces the accumulation of floating-point errors that can arise when backpropagating through long sequences. Second, by isolating the computation to a single round at a time, we only need to move the relevant tensors for that round to the GPU, while keeping the rest on the CPU. Importantly, all tensors involved in our framework are stored on the CPU for use in the CFR algorithm. This design choice significantly reduces GPU memory pressure during training. As a result, we require significantly less GPU VRAM to compute the Nash Equilibrium. On the other hand, the main downside is the increased complexity in implementation and debugging, as computations must now be carefully coordinated across multiple rounds. 

We begin by obtaining the final scores from FGT's terminal nodes. This can be achieved using the following:



\begin{equation}\label{eq:final_util}
\begin{split}
  \py{t\_fsc}  &= \py{7*(2*(t\_scr[:,-1] \% 2) - 1)} \\
  \py{t\_utl} &= \py{t\_fsc[t\_fgm[:,1]]}
\end{split}
\end{equation}

Within CodeSnippet~\ref{cs:bethand}, we need to reverse Line~\eqref{line:bethandlast} to recover \py{t\_utl} at the terminal nodes. In other words, the \py{t\_utl} tensor computed above can be interpreted as the root-level utility of the next round, which must be passed back to the terminal level of the current round. Once \py{t\_utl} is obtained at the root level for each round, it should be propagated upward to the terminal nodes of the previous round and combined with the corresponding running-score tensor for each FGT node at that level. The following code accomplishes this:

\begin{equation}\label{eq:passuputil}
\begin{split}
  \py{t\_rus} &= \py{t\_rus} \otimes \py{c\_scr} \\
  \py{t\_utl} &= \py{t\_utl[t\_lnk]} + \py{t\_rus}
\end{split}
\end{equation}
Here \py{t\_lnk} is defined in the final step of CodeSnippet~\ref{cs:bethand} and \py{t\_rus} appears in the same CodeSnippet, prior to being passed through \py{unique} in line~\ref{line:trus}. 

The next three CodeSnippets present the CFR algorithm. There are two main components: the \texttt{FindUtility} and \texttt{FindReachProbability} functions. The overall CFR algorithm is shown in CodeSnippet~\ref{alg:cfr}.

\begin{algorithm}
\caption{CFR}
\label{alg:cfr}
\begin{algorithmic}[1]
\Function{runcfr}{}
  \State Compute \py{g\_utl} \textcolor{green!50!black}{\scriptsize \# \py{via Equation \eqref{eq:final_util}}}
  
  \For{\py{round} \textbf{in} \py{reversed(rounds)}} 

    \vspace{2 mm}
    \State \py{g\_utl} $\gets$ Pass back up  \py{g\_utl} via Equation \eqref{eq:passuputil}
    \State Initialize \py{g\_reg, g\_sgm, g\_UTL}
    \Statex \textcolor{green!50!black}{\scriptsize \# \py{g\_reg (init. to 0): regrets for all the availble actions, }$r^t$ \py{in Eq. \eqref{eq:DFCR_REG},}}
    \Statex \textcolor{green!50!black}{\scriptsize \# \py{g\_sgm (init. to uniform): strategy profile, }$\sigma^t$  \py{in Eq. \eqref{eq:RM}}}
    \Statex \textcolor{green!50!black}{\scriptsize \# \py{g\_UTL (init. to 0): root-node utilities for current round, }$R^t$ \py{in Eq. \eqref{eq:DFCR_REG}}}
    \For{\py{t} \textbf{in} \py{range(NUM\_ITER)}}
    \State \py{g\_rpr} $\gets$ \Call{findrpr}{\py{g\_sgm}}
      \State \py{t\_utl, g\_sgm, a\_sgm, g\_reg} $\gets$ \Call{FindUtility}{\py{g\_rpr, g\_utl}}
      \State \py{g\_UTL} $\gets$ \py{g\_UTL + t\_utl}
      \EndFor
      \State \py{g\_utl} $\gets$ \py{ g\_UTL/NUM\_ITER}
      \State \py{g\_sgm} $\gets$ \Call{Normalize}{\py{a\_sgm}}  \textcolor{green!50!black}{\scriptsize \# \py{Each FGT node's children's values sum to 1}}

      \EndFor
\EndFunction
\end{algorithmic}
\end{algorithm}
\begin{algorithm}
\caption{FindReachProbability}
\label{cs:findreach}
\begin{algorithmic}[1]
\Function{findrpr}{\py{g\_sgm, i\_cfr}} \textcolor{green!50!black}{\scriptsize \# \py{g\_sgm:current round's strategy.} $\sigma^t$  \py{in Eq. \eqref{eq:RM}}}
\State \py{i\_h} $\gets$ \py{\# nodes in current round's root level}
\State \py{t\_rpr} $\gets$ \py{ones(i\_h)}
\State \py{d\_rpr} $\gets$ \py{\{\}}

\For{\py{i\_trn} \textbf{in} \py{range(4)}}
    \For{\py{i\_ply} \textbf{in} \py{range(2)}}
        \vspace{2mm}
        \State \py{i\_htp} $\gets$ \py{i\_hnd\_i\_trn\_i\_ply}
        \State \py{t\_sgm} $\gets$ \py{g\_sgm[i\_htp]}
        \State \py{t\_sgm} $\gets$ \py{ones\_like(t\_sgm)} \textbf{if} \py{i\_ply == i\_cfr}
        \State \py{t\_edg} $\gets$ \py{gt\_edg[i\_htp]} \textcolor{green!50!black}{\scriptsize \# \py{CUDA tensors gt\_edg contains FGT edges}}
        \State \py{d\_rpr[i\_htp]} $\gets$ \py{t\_sgm * t\_rpr[t\_edg]}
        \State \py{t\_rpr} $\gets$ \py{d\_rpr[i\_htp]}
        \vspace{2mm}
    \EndFor
\EndFor
\State \textbf{return} \py{d\_rpr}
\EndFunction
\end{algorithmic}
\end{algorithm}

\begin{algorithm}
\caption{FindUtility}
\begin{algorithmic}[1]
\Function{FindUtility}{\py{t\_utl}, \py{g\_sgm}, \py{a\_sgm}, \py{g\_reg}, \py{g\_rp0}, \py{g\_rp1}}
\Statex \textcolor{green!50!black}{\scriptsize \# \py{Input: gt\_edg: FGT edges, gc\_edg: FGT branch factor}}
\Statex \textcolor{green!50!black}{\scriptsize \# \py{TOL = 1e-5, a\_sgm: average strategy, g\_rp0, g\_rp1: reach probs for Alex \& Bob resp.}}
\Statex \textcolor{green!50!black}{\scriptsize \# \py{g\_sgm: current strategy}}
    \vspace{0.5mm}
    \For{\py{i\_trn} \textbf{in} \Call{reversed}{\py{range(4)}}}
        \For{\py{i\_ply} \textbf{in} \Call{reversed}{\py{range(2)}}}
            \vspace{2mm}
            \State \py{i\_htp} $\gets$ \py{i\_hnd\_i\_trn\_i\_ply}
            \State \py{i\_h} $\gets$ \py{\# nodes in current layer's root-level}
            \vspace{2mm}
            \State \py{t\_ply, t\_opp} $\gets$ \py{g\_rp0[i\_htp], g\_rp1[i\_htp]} \textbf{if} \py{i\_ply == 1}
            \State \py{t\_ply, t\_opp} $\gets$ \py{g\_rp1[i\_htp], g\_rp0[i\_htp]}  \textbf{if} \py{i\_ply == 0}
            \vspace{2mm}
            \State \py{c\_edg,t\_edg,t\_sgm}$\gets$\py{gc\_edg[i\_htp], gt\_edg[i\_htp],g\_sgm[i\_htp]}
            \vspace{2mm}
            \State \py{t\_pt2} $\gets$ \py{zeros(i\_h)}
            \State \py{t\_pt2.scatter\_add\_(0,t\_edg,t\_sgm*t\_utl)}
            \vspace{2mm}
            \State \py{t\_reg} $\gets$ \py{(1-2*i\_ply)*t\_opp*(t\_utl-t\_pt2[t\_edg])}
            \State \py{t\_utl} $\gets$ \py{t\_pt2}
            \vspace{2mm}
            \State \py{t\_msk} $\gets$ \py{g\_reg[\texttt{i\_htp}] >= TOL}
            \State \py{t\_fct} $\gets$ \py{zeros\_like(g\_reg[i\_htp])}
            \State \py{t\_fct[t\_msk]} $\gets$ \py{i\_cnt**1.5/(1+i\_cnt**1.5)}
            \State \py{t\_fct[{\textasciitilde}t\_msk]} $\gets$ \py{0.5}
            \State \py{g\_reg[i\_htp]} $\gets$ \py{t\_fct*g\_reg[i\_htp]+t\_reg}
            \State \py{t\_rts} $\gets$ \py{clamp(1000*g\_reg[i\_htp], min=TOL)}
            \vspace{2mm}
            \State \py{t\_sum} $\gets$ \py{zeros(i\_h)}
            \State \py{t\_sum.scatter\_add\_(0, t\_edg, t\_rts)}
            \vspace{2mm}
            \State \py{t\_msk} $\gets$ \py{t\_sum[t\_edg] <= TOL}
            \State \py{t\_sgm[{\textasciitilde}t\_msk]} $\gets$ \py{t\_rts[{\textasciitilde}t\_msk]/t\_sum[t\_edg][{\textasciitilde}t\_msk]}
            \State \py{t\_sgm[t\_msk]} $\gets$ \py{(1/c\_edg[t\_edg])[t\_msk]}
            \vspace{2mm}
            \State \py{a\_sgm[i\_htp]}$\gets$\py{(1-1/(1+i\_cnt)**2)*a\_sgm[i\_htp] + t\_sgm*t\_ply}
            \State \py{g\_sgm[i\_htp]} $\gets$ \py{t\_sgm}
            \vspace{2mm}
        \EndFor
    \EndFor
    \vspace{0.5mm}
    \State \Return \py{t\_utl}, \py{a\_sgm}, \py{g\_sgm}, \py{g\_reg}
    \vspace{2mm}
\EndFunction
\end{algorithmic}
\label{cs:findutil}
\end{algorithm}

\newpage

%% file: exsampling.tex
\subsection{External Sampling}\label{sec:exsample}
In this section, we describe the procedure for external sampling used during the unfolding process. Although the present work does not rely on or develop specific sampling techniques, we include this explanation for the sake of completeness and future reference. The external sampling method outlined here will play a key role in subsequent extensions of this research, where sampling-based approaches will be more actively explored and integrated into the modeling framework.

In external sampling, a single action is selected for the opponent based on their current strategy. This selection is performed within FGT, as illustrated in Figure~\ref{fig:exsample_fgm}. To further aid understanding, Figure~\ref{fig:tfgmexsampling} illustrates how the FGT tensor \py{t\_fgm} is updated/sampled. CodeSnippet~\ref{alg:exsampling} details the procedure used in external sampling.

\input{figs/fig00_exsampling}
\input{figs/fig01_exsampling}

\input{algs/exsampling}

%% file: figs/fig00_exsampling.tex
\begin{figure}
\centering
\scalebox{0.5}{
    \begin{tikzpicture}

        \node[rotate=0] at (-2, -4) {\Large Game Tree}; 
        \node[rotate=0] at (11, -4) {\Large Full Game Tree}; 
 \draw[->, very thick, decorate, decoration={snake, amplitude=2mm, segment length=5mm}] (2.5cm, -9cm) -- (4.5cm, -9cm);
        \node[rotate=0] at (3.5, -8) {\Large Unfold w/}; 
        \node[rotate=0] at (3.5, -10) {\Large Ext. Sampling};
        
      \drawboxwitharrows{0}{blue!50/0.4, red/0.05, green/-0.2}{dummy}{\py{0}}  {-2.5cm,  -8cm}{none}
      \drawboxwitharrows{0}{blue!50/0.4, orange/0.05}{dummy}{\py{1}}  {-1.5cm,   -8cm}{none}

      \drawboxwitharrows{0}{blue!50/0.4, red/0.05, green/-0.2}{dummy}                                       {\py{0}}    {-4cm,  -10cm}{none}
      \drawboxwitharrows{0}{blue!50/0.4, red/0.05, green/-0.2}{dummy}                                       {\py{1}}    {-3cm,  -10cm}{none}
      \drawboxwitharrows{0}{blue!50/0.4, orange/0.05}{dummy}                                       {\py{2}}    {-2cm,  -10cm}{none}
      \drawboxwitharrows{0}{blue!50/0.4, orange/0.05}{dummy}                                       {\py{3}}    {-1cm,  -10cm}{none}
      \drawboxwitharrows{0}{blue!50/0.4, orange/0.05}{dummy}                                       {\py{4}}    {0cm,   -10cm}{none}
       \draw (-4.5cm,    -10.5cm) -- (-2.5cm, -10.5cm);
        \draw (-2.2cm, -10.5cm) -- (0.4cm, -10.5cm);
         \drawboxwitharrows{0}{}{dummy}{\py{0}}  {9cm,   -7cm}{green}
         \drawboxwitharrows{0}{}{dummy}{\py{0}}  {10cm,  -7cm}{red}
         \drawboxwitharrows{0}{}{dummy}{\py{0}}  {11cm,  -7cm}{blue!50}
         \drawboxwitharrows{0}{}{dummy}{\py{1}}  {12cm,  -7cm}{orange}
         \drawboxwitharrows{0}{}{dummy}{\py{1}}  {13cm,  -7cm}{blue!50}

         \drawboxwitharrows{0}{}{dummy}{\py{0}}  {6cm,   -11cm}{green}
         \drawboxwitharrows{0}{}{dummy}{\py{0}}  {7cm,   -11cm}{red}
         \drawboxwitharrows{0}{}{dummy}{\py{0}}  {8cm,   -11cm}{blue!50}
          \drawboxwitharrows{0}{}{dummy}{\py{1}} {9cm,   -11cm}{green}
         \drawboxwitharrows{0}{}{dummy}{\py{1}}  {10cm,  -11cm}{red}
         \drawboxwitharrows{0}{}{dummy}{\py{1}}  {11cm,  -11cm}{blue!50}
         \drawboxwitharrows{0}{}{dummy}{\py{2}}  {12cm,  -11cm}{orange}
         \drawboxwitharrows{0}{}{dummy}{\py{2}}  {13cm,  -11cm}{blue!50}
         \drawboxwitharrows{0}{}{dummy}{\py{3}}  {14cm,  -11cm}{orange}
         \drawboxwitharrows{0}{}{dummy}{\py{3}}  {15cm,  -11cm}{blue!50}
         \drawboxwitharrows{0}{}{dummy}{\py{4}}  {16cm,  -11cm}{orange}
         \drawboxwitharrows{0}{}{dummy}{\py{4}}  {17cm,  -11cm}{blue!50}


          \draw[thick, decoration={markings, mark=at position 0.5 with {\arrow{>}}}, postaction={decorate}] (9, -7.25) to[out=-80, in=60] (9, -10.75); 

          \draw[thick, decoration={markings, mark=at position 0.5 with {\arrow{>}}}, postaction={decorate}] (10, -7.25) to[out=-80, in=60] (10, -10.75);

          \draw[thick, decoration={markings, mark=at position 0.5 with {\arrow{>}}}, postaction={decorate}] (11, -7.25) to[out=-80, in=60] (11, -10.75); 

          \draw[thick, decoration={markings, mark=at position 0.5 with {\arrow{>}}}, postaction={decorate}] (12, -7.25) to[out=-80, in=60] (14, -10.75); 

          \draw[thick, decoration={markings, mark=at position 0.5 with {\arrow{>}}}, postaction={decorate}] (13, -7.25) to[out=-80, in=60] (17, -10.75);


         \node[rotate=0] at (9, -11.5) {\py{{0}}}; 
         \node[rotate=0] at (10, -11.5) {\py{{1}}}; 
         \node[rotate=0] at (11, -11.5) {\py{{2}}}; 
         \node[rotate=0] at (14, -11.5) {\py{{3}}}; 
         \node[rotate=0] at (17, -11.5) {\py{{4}}}; 
    \end{tikzpicture}
    }
\caption{An illustration of how \py{t\_edg} is constructed under ex. sampling regime.}
    \label{fig:exsample_fgm}
\end{figure}
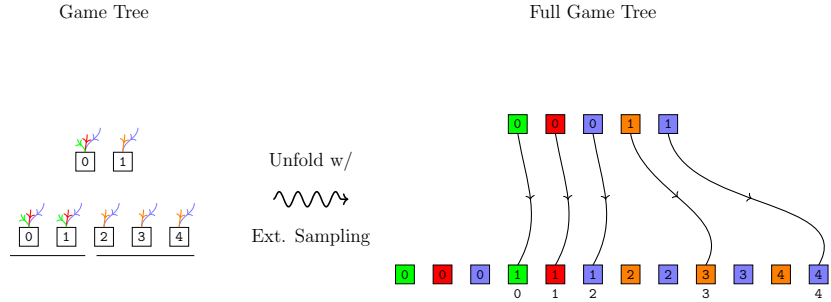

%% file: figs/fig01_exsampling.tex
\begin{figure}[h]
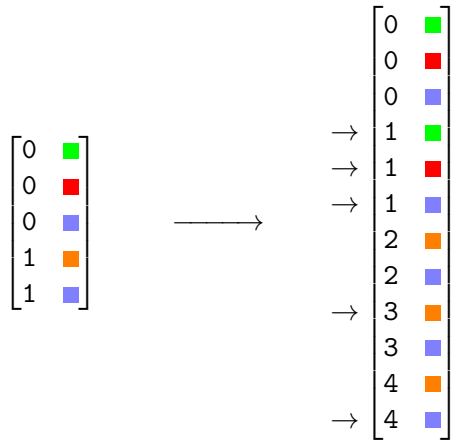

  \centering
\[
\begin{bmatrix}
\py{0} & \tikz\fill[green] (0,0) rectangle (0.2,0.2); \\
\py{0} & \tikz\fill[red] (0,0) rectangle (0.2,0.2); \\
\py{0} & \tikz\fill[fill=blue!50] (0,0) rectangle (0.2,0.2); \\
\py{1} & \tikz\fill[orange] (0,0) rectangle (0.2,0.2); \\
\py{1} & \tikz\fill[fill=blue!50] (0,0) rectangle (0.2,0.2); 
\end{bmatrix}
\;
\qquad
\xrightarrow{\hspace{1cm}}
\qquad
\begin{matrix}
 \\
 \\
 \\
\rightarrow \\
\rightarrow \\
\rightarrow \\
\\
\\
\rightarrow \\
\\
\\
\rightarrow \\
\end{matrix}
\begin{bmatrix}
\py{0} & \tikz\fill[green] (0,0) rectangle (0.2,0.2); \\
\py{0} & \tikz\fill[red] (0,0) rectangle (0.2,0.2); \\
\py{0} & \tikz\fill[fill=blue!50] (0,0) rectangle (0.2,0.2); \\
\py{1} & \tikz\fill[green] (0,0) rectangle (0.2,0.2); \\
\py{1} & \tikz\fill[red] (0,0) rectangle (0.2,0.2); \\
\py{1} & \tikz\fill[fill=blue!50] (0,0) rectangle (0.2,0.2); \\
\py{2} & \tikz\fill[orange] (0,0) rectangle (0.2,0.2); \\
\py{2} & \tikz\fill[fill=blue!50] (0,0) rectangle (0.2,0.2); \\
\py{3} & \tikz\fill[orange] (0,0) rectangle (0.2,0.2); \\
\py{3} & \tikz\fill[fill=blue!50] (0,0) rectangle (0.2,0.2); \\
\py{4} & \tikz\fill[orange] (0,0) rectangle (0.2,0.2); \\
\py{4} & \tikz\fill[fill=blue!50] (0,0) rectangle (0.2,0.2);
\end{bmatrix}
\]
  \caption{An illustration of how \py{t\_fgm} is updated under ext. sampling regime.}
  \label{fig:tfgmexsampling}
\end{figure}

%% file: algs/exsampling.tex
\begin{algorithm}
\caption{ExternalSampling}
\label{alg:exsampling}
\begin{algorithmic}[1]
\State Input \py{t\_edg,c\_edg,t\_sgm}
\Statex\textcolor{green!50!black}{\scriptsize\py{\#t\_edg links lower to upper nodes, while c\_edg specifies the degree of each upper node. \#Ex: t\_edg = [0,1,2,0,1,2,3,4,3,4,3,4], c\_edg = t\_brf$\otimes$c\_scr = [2,2,2,3,3]}}
\State \py{t\_inv = argsort(t\_edg, stable=True)} 
\Statex\textcolor{green!50!black}{\scriptsize\py{\#t\_inv = [0,3,1,4,2,5,6,8,10,7,9,11]}}
\Statex\textcolor{green!50!black}{\scriptsize\py{\#use t\_inv when sorting edges.1st green maps to index 0, 2nd green maps to index 3, etc}}
\State \py{t\_idx = empty\_like(t\_inv)}
\State \py{t\_idx[t\_inv] $\gets$ arange(len(t\_inv))}
\Statex\textcolor{green!50!black}{\scriptsize\py{\#t\_idx =[ 0,  2,  4,  1,  3,  5,  6,  9,  7, 10,  8, 11]}}
\Statex\textcolor{green!50!black}{\scriptsize\py{\#use t\_idx  when mapping back sorted edges}}
\State \py{i\_max = c\_edg.max()}
\State \py{t\_msk = arange(i\_max).unsqueeze(0) < c\_edg.unsqueeze(1)}
\Statex\textcolor{green!50!black}{\scriptsize\py{\#t\_msk = tensor([[ T,  T, F],
        [ T,  T, F],
        [ T,  T, F],
        [ T,  T,  T],
        [ T,  T,  T]])}}
\State \py{t\_mtx = zeros\_like(t\_msk)}
\State \py{t\_mtx[t\_msk]  $\gets$ t\_sgm[t\_inv]}
\State \py{t\_smp = multinomial(t\_mtx).squeeze(1)}
\Statex\textcolor{green!50!black}{\scriptsize\py{\#sampled indices for each row}}
\State \py{t\_gps = cat([0, c\_edg.cumsum()[:-1]])}
\State  \py{t\_res = zeros\_like(t\_sgm, dtype=bool)}
\State \py{t\_res[t\_gps + t\_smp] $\gets$ True}
\State \py{t\_res $\gets$ t\_res[t\_idx]}
\end{algorithmic}
\end{algorithm}



%% file: tree.tex
\section{Experimental Evaluation}\label{sec:experimental}
\begin{figure}[htbp]
  \centering
  \includegraphics[width=1.0\textwidth]{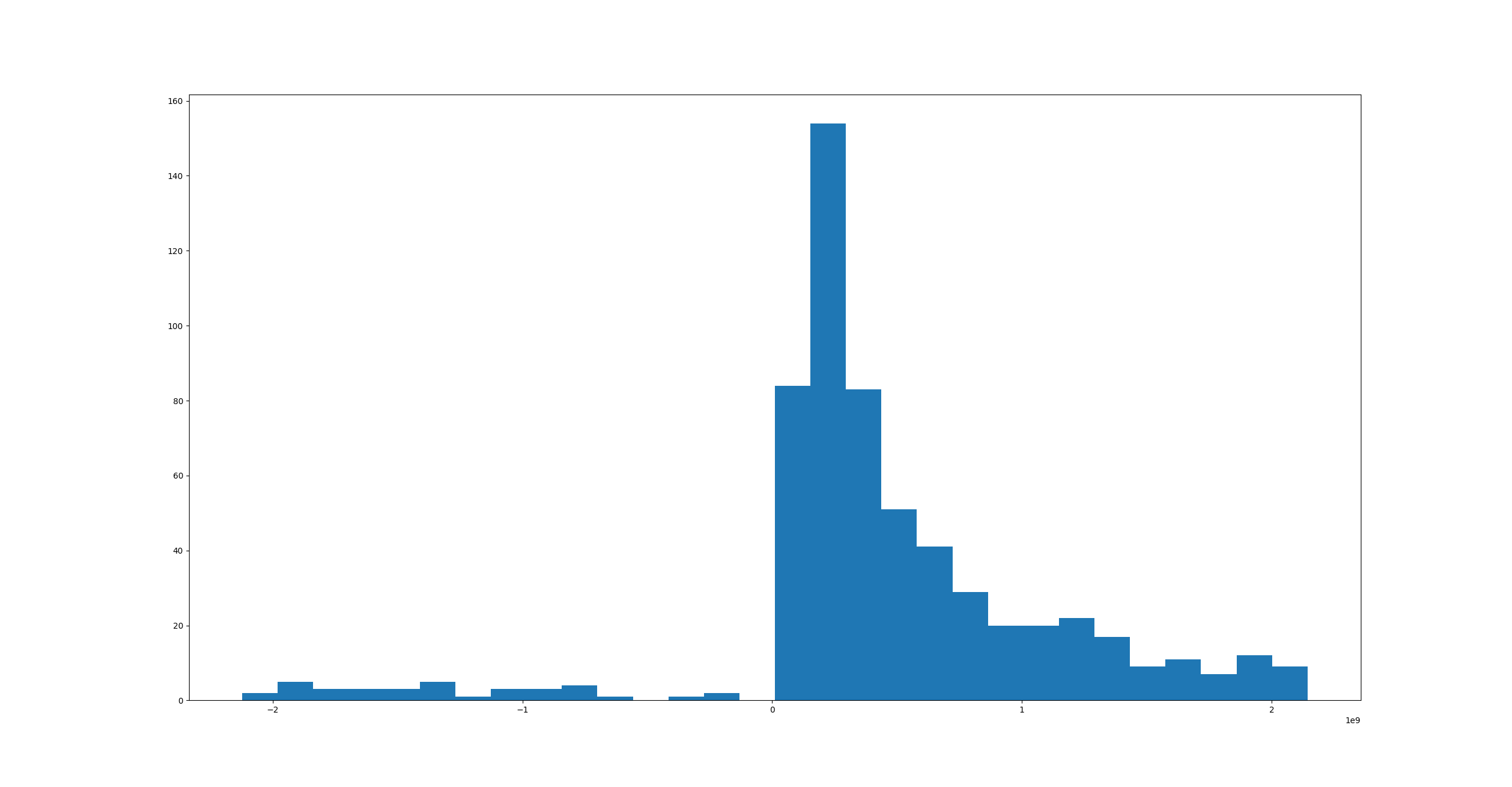}
  \caption{Distribution of game tree sizes (i.e., strategy profile table shapes) across 500 randomly generated decks.}
  \label{fig:Dsizes}
\end{figure}

We trained more than 500 randomly generated decks using the CFR algorithm and obtain strategy profiles similar (stored in \py{.parquete}) to the one shown in Table~\ref{tab:deck9_exp}. Figure~\ref{fig:Dsizes} shows the distribution of the sizes of the resulting full game trees corresponding to the shapes of the strategy profile tables.

We also illustrate a near-Nash equilibrium strategy, computed via CFR, in Figure~\ref{fig:game_9} and Table~\ref{tab:deck9_exp}. Figure~\ref{fig:game_9} shows a sample game between two near-Nash strategies, while Table~\ref{tab:deck9_exp} lists their first-turn strategies in the first round.


Once we have the strategy profiles at our disposal, we fit an XGBoost model to predict $100 \times \sigma$, using root mean squared error (RMSE) as the evaluation metric and setting \py{gamma=1} as the regularization parameter. This allows us to pass the current infoset to the model in order to obtain a near-Nash strategy for the active player. The main advantage of this approach is that it enables the simulation of multiple self-play games in parallel on the GPU. Specifically, if \py{N} denotes the number of self-play instances, then the infoset tensor at each turn has shape \py{[N, 58]}. Sampling based on the learned strategy profile is performed using the same method introduced in CodeSnippet~\ref{alg:exsampling}.

\begin{figure}[htbp]
  \centering
  \includegraphics[width=1.2\textwidth]{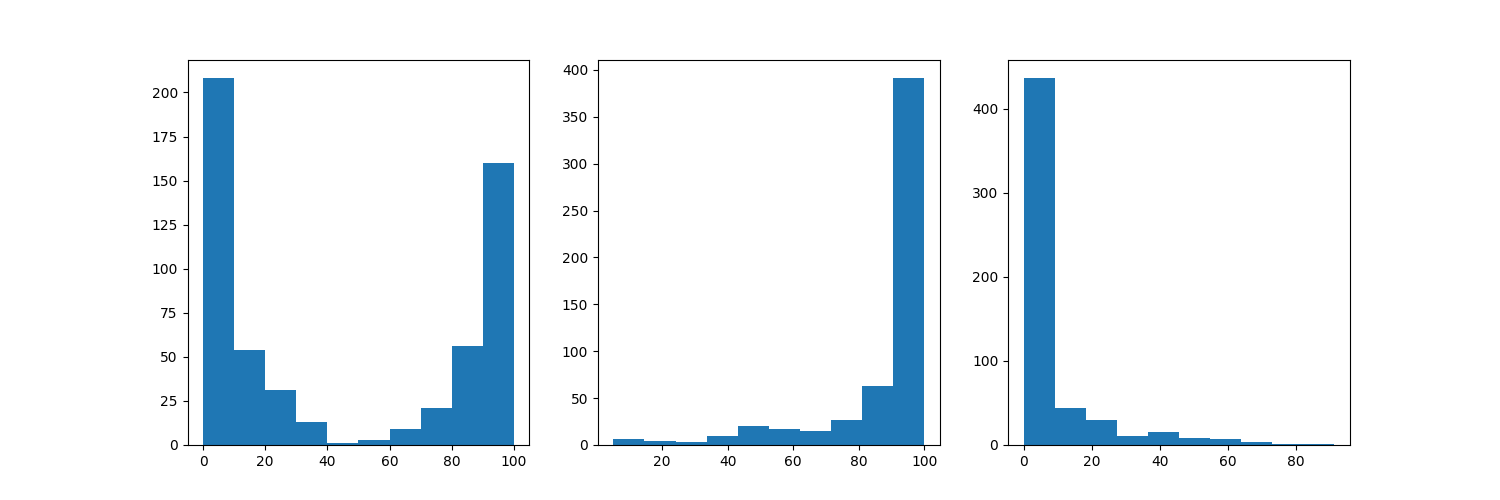}
  \caption{Left panel shows near-Nash vs. near-Nash, middle panel shows near-Nash vs. Random, and the right panel shows Random vs. near-Nash}
  \label{fig:MMMRRM}
\end{figure}

With the ability to run multiple self-play games in parallel, we can now estimate the fair value of each deck composition. To do so, we let two near-Nash equilibrium strategies compete against each other using a given deck. The percentage of games won by Alex, divided by 100, is what we define as the fair value of that deck. As a first sanity check during development, we evaluated our model against a random strategy. Figure~\ref{fig:MMMRRM} presents the results: the left panel shows near-Nash vs. near-Nash, the middle panel shows near-Nash vs. Random, and the right panel shows Random vs. near-Nash.

\input{figs/game_9.tex}

\begin{table}[ht]
\centering
\begin{tabular}{|c|c|c|c|}
\hline
\# J & \# $\clubsuit$ & $v$ & std \\
\hline
=\; 0 & $\leq 7$            & 0.06  & 0.03 \\
=\; 0 & $> 7$               & 0.22  & 0.11 \\
=\; 1 & $\leq 5$            & 0.06  & 0.02 \\
=\; 1 & $> 5$               & 0.25  & 0.12 \\
=\; 2 & $\leq 6$            & 0.36  & 0.15 \\
=\; 2 & $> 6$               & 0.54  & 0.17 \\
=\; 3 & $\leq 5$            & 0.58  & 0.15 \\
=\; 3 & $> 5$               & 0.76  & 0.10 \\
=\; 4 & $\leq 4$            & 0.66  & 0.11 \\
=\; 4 & $> 4$               & 0.85  & 0.09 \\
\hline
\end{tabular}
\caption{Deck fair values vs. Jacks and Clubs held by Alex.}
\label{tab:tree_paths}
\end{table}

Another step in our analysis is to relate the fair value of a deck to key elements of its composition. For example, it is clear that holding Jack cards or Club cards provides a strategic advantage to the player. To investigate this relationship, we fit decision trees to subsets of decks, each categorized by the number of Jack cards held by Alex. We then examine how the decision tree branches based on the number of Clubs held by Alex. Table~\ref{tab:tree_paths} illustrates this relationship: the more Jacks or Clubs Alex holds, the higher the fair value of the deck tends to be; providing supporting evidence that our CFR algorithm has converged to near-Nash equilibrium strategies.

\begin{figure}[htbp]
  \centering
  \includegraphics[width=1.0\textwidth]{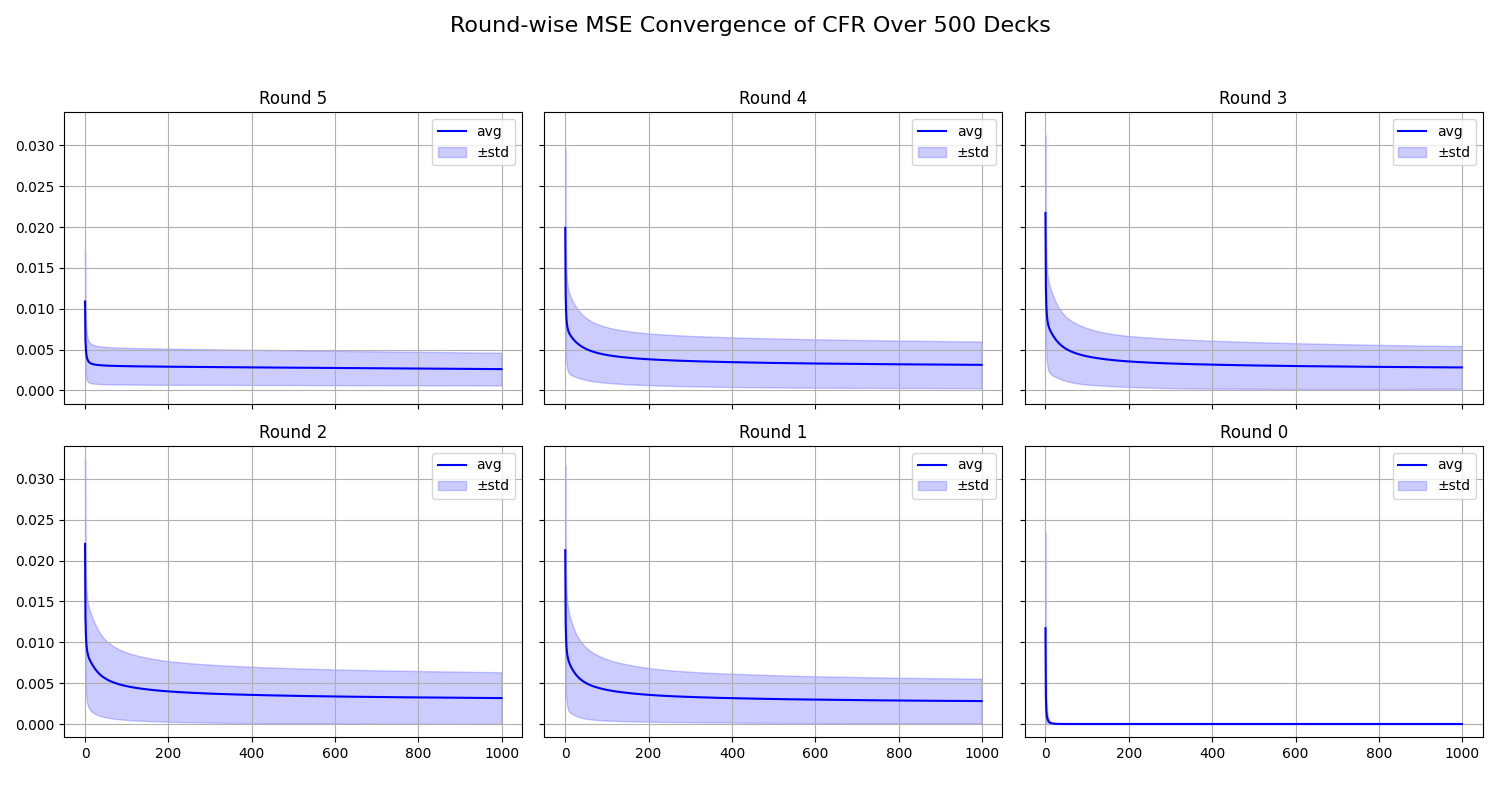}
\caption{Convergence of CFR for 500 random decks. The plot shows the average $\pm$ standard deviation bands of the mean squared error (MSE) to the final CFR vector, measured separately for each round. As observed, the last round converges the fastest due to having the fewest nodes in the FT representation.}
  \label{fig:cfrconv}
\end{figure}

\begin{figure}[htbp]
  \centering
  \includegraphics[width=0.7\textwidth]{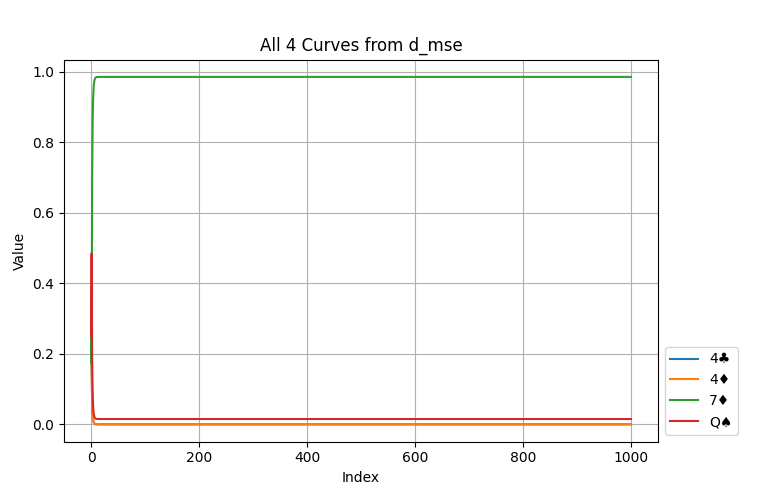}
  \caption{$\sigma$ values for Alex's first turn using the deck from Figure~\ref{fig:game_9}}
  \label{fig:deck9_000}
\end{figure}

We next examine the convergence behavior of CFR. To this end, we measure the mean squared error (MSE) between each intermediate strategy profile $\bar{\sigma}^t$ and the final CFR strategy vector. Since our approach computes strategies on a round-by-round basis, we provide separate plots for each round to visualize the convergence. It is important to note that, as shown in CodeSnippet~\ref{cs:findutil}, we use the DCFR algorithm from~\cite{brown2019solving} with the following parameters:
\[
\gamma = 2,\quad \alpha = 1.5,\quad \beta = 0.
\]
Figure \ref{fig:cfrconv} shows the CFR convergence behavior. Finally, Figure \ref{fig:deck9_000} shows the $\sigma$ value for the first turn of Alex for Deck of Figure \ref{fig:game_9}. 

%% file: figs/game_9.tex
\newpage 
\begin{figure}[H]  
\begin{center}
    \scriptsize
    \caption*{Sample Game Played Between Near-Nash Strategies}
\resizebox{1\textwidth}{!}{%
\begin{tabular}{|>{\centering\arraybackslash}p{0.1\textwidth}|
                >{\centering\arraybackslash}p{0.2\textwidth}|
                >{\centering\arraybackslash}p{0.2\textwidth}|
                >{\centering\arraybackslash}p{0.7\textwidth}|
                >{\centering\arraybackslash}p{0.05\textwidth}|
                 >{\centering\arraybackslash}p{0.7\textwidth}|
                >{\centering\arraybackslash}p{0.05\textwidth}|
                >{\centering\arraybackslash}p{0.05\textwidth}|
                 >{\centering\arraybackslash}p{0.05\textwidth}|
                  >{\centering\arraybackslash}p{0.05\textwidth}|
                   >{\centering\arraybackslash}p{0.05\textwidth}|
                   >{\centering\arraybackslash}p{0.05\textwidth}|
                   >{\centering\arraybackslash}p{0.05\textwidth}|
                   >{\centering\arraybackslash}p{0.05\textwidth}|}
\hline
\textbf{Stage}& \textbf{Alex} & \textbf{Bob}& \textbf{Pool} & \textbf{Lay} & \textbf{Pick} & \textbf{Acl} &  \textbf{Bcl} & \textbf{Apt} &  \textbf{Bpt} & \textbf{Asr} & \textbf{Bsr} & \textbf{$\Delta$} & \textbf{L}\\
\hline
\rowcolor{myrowcolor}

 0\_0\_0 &$\FoC\;\FoD\;\SeD\;\QC$&$\ThD\;\ThH\;\FiC\;\KS$&$\AC\;\AS\;\ND\;\KD$&$\SeD$&&0 &0 &0 &0 &0 &0 &0 &0\\ \hline

 0\_0\_1 &$\FoC\;\FoD\;\QC$&$\ThD\;\ThH\;\FiC\;\KS$&$\AC\;\AS\;\SeD\;\ND\;\KD$&$\ThD$&$\AC\;\SeD$&0 &1 &0 &1 &0 &0 &0 &B\\ \hline
 \rowcolor{myrowcolor}

 0\_1\_0 &$\FoC\;\FoD\;\QC$&$\ThH\;\FiC\;\KS$&$\AS\;\ND\;\KD$&$\FoD$&&0 &1 &0 &1 &0 &0 &0 &B\\ \hline

 0\_1\_1 &$\FoC\;\QC$&$\ThH\;\FiC\;\KS$&$\AS\;\FoD\;\ND\;\KD$&$\ThH$&&0 &1 &0 &1 &0 &0 &0 &B\\ \hline
 \rowcolor{myrowcolor}

 0\_2\_0 &$\FoC\;\QC$&$\FiC\;\KS$&$\AS\;\ThH\;\FoD\;\ND\;\KD$&$\QC$&&0 &1 &0 &1 &0 &0 &0 &B\\ \hline

 0\_2\_1 &$\FoC$&$\FiC\;\KS$&$\AS\;\ThH\;\FoD\;\ND\;\QC\;\KD$&$\KS$&$\KD$&0 &1 &0 &1 &0 &0 &0 &B\\ \hline
 \rowcolor{myrowcolor}

 0\_3\_0 &$\FoC$&$\FiC$&$\AS\;\ThH\;\FoD\;\ND\;\QC$&$\FoC$&$\ThH\;\FoD$&1 &1 &0 &1 &0 &0 &0 &A\\ \hline

 0\_3\_1 &&$\FiC$&$\AS\;\ND\;\QC$&$\FiC$&&1 &1 &0 &1 &0 &0 &0 &A\\ \hline
 \rowcolor{myrowcolor}
\hline

 1\_0\_0 &$\SiC\;\SiD\;\NH\;\JC$&$\SiH\;\SeC\;\JD\;\KH$&$\AS\;\FiC\;\ND\;\QC$&$\JC$&$\AS\;\FiC\;\ND$&3 &1 &2 &0 &0 &0 &-1 &A\\ \hline

 1\_0\_1 &$\SiC\;\SiD\;\NH$&$\SiH\;\SeC\;\JD\;\KH$&$\QC$&$\SiH$&&3 &1 &2 &0 &0 &0 &-1 &A\\ \hline
 \rowcolor{myrowcolor}

 1\_1\_0 &$\SiC\;\SiD\;\NH$&$\SeC\;\JD\;\KH$&$\SiH\;\QC$&$\NH$&&3 &1 &2 &0 &0 &0 &-1 &A\\ \hline

 1\_1\_1 &$\SiC\;\SiD$&$\SeC\;\JD\;\KH$&$\SiH\;\NH\;\QC$&$\SeC$&&3 &1 &2 &0 &0 &0 &-1 &A\\ \hline
 \rowcolor{myrowcolor}

 1\_2\_0 &$\SiC\;\SiD$&$\JD\;\KH$&$\SiH\;\SeC\;\NH\;\QC$&$\SiD$&&3 &1 &2 &0 &0 &0 &-1 &A\\ \hline

 1\_2\_1 &$\SiC$&$\JD\;\KH$&$\SiD\;\SiH\;\SeC\;\NH\;\QC$&$\KH$&&3 &1 &2 &0 &0 &0 &-1 &A\\ \hline
 \rowcolor{myrowcolor}

 1\_3\_0 &$\SiC$&$\JD$&$\SiD\;\SiH\;\SeC\;\NH\;\QC\;\KH$&$\SiC$&&3 &1 &2 &0 &0 &0 &-1 &A\\ \hline

 1\_3\_1 &&$\JD$&$\SiC\;\SiD\;\SiH\;\SeC\;\NH\;\QC\;\KH$&$\JD$&$\SiC\;\SiD\;\SiH\;\SeC\;\NH$&3 &3 &2 &1 &0 &0 &-1 &B\\ \hline
 \rowcolor{myrowcolor}
\hline

 2\_0\_0 &$\FiD\;\NC\;\TeH\;\KC$&$\AD\;\SiS\;\EC\;\ES$&$\QC\;\KH$&$\NC$&&3 &3 &0 &0 &0 &0 &0 &0\\ \hline

 2\_0\_1 &$\FiD\;\TeH\;\KC$&$\AD\;\SiS\;\EC\;\ES$&$\NC\;\QC\;\KH$&$\SiS$&&3 &3 &0 &0 &0 &0 &0 &0\\ \hline
 \rowcolor{myrowcolor}

 2\_1\_0 &$\FiD\;\TeH\;\KC$&$\AD\;\EC\;\ES$&$\SiS\;\NC\;\QC\;\KH$&$\KC$&$\KH$&4 &3 &0 &0 &0 &0 &0 &A\\ \hline

 2\_1\_1 &$\FiD\;\TeH$&$\AD\;\EC\;\ES$&$\SiS\;\NC\;\QC$&$\EC$&&4 &3 &0 &0 &0 &0 &0 &A\\ \hline
 \rowcolor{myrowcolor}

 2\_2\_0 &$\FiD\;\TeH$&$\AD\;\ES$&$\SiS\;\EC\;\NC\;\QC$&$\FiD$&$\SiS$&4 &3 &0 &0 &0 &0 &0 &A\\ \hline

 2\_2\_1 &$\TeH$&$\AD\;\ES$&$\EC\;\NC\;\QC$&$\ES$&&4 &3 &0 &0 &0 &0 &0 &A\\ \hline
 \rowcolor{myrowcolor}

 2\_3\_0 &$\TeH$&$\AD$&$\EC\;\ES\;\NC\;\QC$&$\TeH$&&4 &3 &0 &0 &0 &0 &0 &A\\ \hline

 2\_3\_1 &&$\AD$&$\EC\;\ES\;\NC\;\TeH\;\QC$&$\AD$&$\TeH$&4 &3 &0 &1 &0 &0 &0 &B\\ \hline
 \rowcolor{myrowcolor}
\hline

 3\_0\_0 &$\TwC\;\ThC\;\JS\;\QS$&$\SeS\;\ED\;\EH\;\QH$&$\EC\;\ES\;\NC\;\QC$&$\QS$&$\QC$&5 &3 &0 &0 &0 &0 &-1 &A\\ \hline

 3\_0\_1 &$\TwC\;\ThC\;\JS$&$\SeS\;\ED\;\EH\;\QH$&$\EC\;\ES\;\NC$&$\SeS$&&5 &3 &0 &0 &0 &0 &-1 &A\\ \hline
 \rowcolor{myrowcolor}

 3\_1\_0 &$\TwC\;\ThC\;\JS$&$\ED\;\EH\;\QH$&$\SeS\;\EC\;\ES\;\NC$&$\TwC$&$\NC$&7 &3 &2 &0 &0 &0 &-1 &A\\ \hline

 3\_1\_1 &$\ThC\;\JS$&$\ED\;\EH\;\QH$&$\SeS\;\EC\;\ES$&$\EH$&&7 &3 &2 &0 &0 &0 &-1 &A\\ \hline
 \rowcolor{myrowcolor}

 3\_2\_0 &$\ThC\;\JS$&$\ED\;\QH$&$\SeS\;\EC\;\EH\;\ES$&$\JS$&$\SeS\;\EC\;\EH\;\ES$&8 &3 &3 &0 &0 &0 &-1 &A\\ \hline

 3\_2\_1 &$\ThC$&$\ED\;\QH$&&$\QH$&&8 &3 &3 &0 &0 &0 &-1 &A\\ \hline
 \rowcolor{myrowcolor}

 3\_3\_0 &$\ThC$&$\ED$&$\QH$&$\ThC$&&8 &3 &3 &0 &0 &0 &-1 &A\\ \hline

 3\_3\_1 &&$\ED$&$\ThC\;\QH$&$\ED$&$\ThC$&8 &4 &3 &0 &0 &0 &-1 &B\\ \hline
 \rowcolor{myrowcolor}
\hline

 4\_0\_0 &$\FoH\;\FoS\;\SeH\;\JH$&$\TwD\;\FiH\;\NS\;\TeS$&$\QH$&$\JH$&&0 &0 &0 &0 &0 &0 &2 &0\\ \hline

 4\_0\_1 &$\FoH\;\FoS\;\SeH$&$\TwD\;\FiH\;\NS\;\TeS$&$\JH\;\QH$&$\TwD$&&0 &0 &0 &0 &0 &0 &2 &0\\ \hline
 \rowcolor{myrowcolor}

 4\_1\_0 &$\FoH\;\FoS\;\SeH$&$\FiH\;\NS\;\TeS$&$\TwD\;\JH\;\QH$&$\SeH$&&0 &0 &0 &0 &0 &0 &2 &0\\ \hline

 4\_1\_1 &$\FoH\;\FoS$&$\FiH\;\NS\;\TeS$&$\TwD\;\SeH\;\JH\;\QH$&$\NS$&$\TwD$&0 &0 &0 &0 &0 &0 &2 &B\\ \hline
 \rowcolor{myrowcolor}

 4\_2\_0 &$\FoH\;\FoS$&$\FiH\;\TeS$&$\SeH\;\JH\;\QH$&$\FoH$&$\SeH$&0 &0 &0 &0 &0 &0 &2 &A\\ \hline

 4\_2\_1 &$\FoS$&$\FiH\;\TeS$&$\JH\;\QH$&$\FiH$&&0 &0 &0 &0 &0 &0 &2 &A\\ \hline
 \rowcolor{myrowcolor}

 4\_3\_0 &$\FoS$&$\TeS$&$\FiH\;\JH\;\QH$&$\FoS$&&0 &0 &0 &0 &0 &0 &2 &A\\ \hline

 4\_3\_1 &&$\TeS$&$\FoS\;\FiH\;\JH\;\QH$&$\TeS$&&0 &0 &0 &0 &0 &0 &2 &A\\ \hline
 \rowcolor{myrowcolor}
\hline

 5\_0\_0 &$\AH\;\ThS\;\TeC\;\QD$&$\TwH\;\TwS\;\FiS\;\TeD$&$\FoS\;\FiH\;\TeS\;\JH\;\QH$&$\TeC$&&0 &0 &0 &0 &0 &0 &2 &0\\ \hline

 5\_0\_1 &$\AH\;\ThS\;\QD$&$\TwH\;\TwS\;\FiS\;\TeD$&$\FoS\;\FiH\;\TeC\;\TeS\;\JH\;\QH$&$\FiS$&&0 &0 &0 &0 &0 &0 &2 &0\\ \hline
 \rowcolor{myrowcolor}

 5\_1\_0 &$\AH\;\ThS\;\QD$&$\TwH\;\TwS\;\TeD$&$\FoS\;\FiH\;\FiS\;\TeC\;\TeS\;\JH\;\QH$&$\ThS$&&0 &0 &0 &0 &0 &0 &2 &0\\ \hline

 5\_1\_1 &$\AH\;\QD$&$\TwH\;\TwS\;\TeD$&$\ThS\;\FoS\;\FiH\;\FiS\;\TeC\;\TeS\;\JH\;\QH$&$\TeD$&&0 &0 &0 &0 &0 &0 &2 &0\\ \hline
 \rowcolor{myrowcolor}

 5\_2\_0 &$\AH\;\QD$&$\TwH\;\TwS$&$\ThS\;\FoS\;\FiH\;\FiS\;\TeC\;\TeD\;\TeS\;\JH\;\QH$&$\AH$&$\TeC$&1 &0 &1 &0 &0 &0 &2 &A\\ \hline

 5\_2\_1 &$\QD$&$\TwH\;\TwS$&$\ThS\;\FoS\;\FiH\;\FiS\;\TeD\;\TeS\;\JH\;\QH$&$\TwS$&$\FoS\;\FiS$&1 &0 &1 &0 &0 &0 &2 &B\\ \hline
 \rowcolor{myrowcolor}

 5\_3\_0 &$\QD$&$\TwH$&$\ThS\;\FiH\;\TeD\;\TeS\;\JH\;\QH$&$\QD$&$\QH$&1 &0 &1 &0 &0 &0 &2 &A\\ \hline

 5\_3\_1 &&$\TwH$&$\ThS\;\FiH\;\TeD\;\TeS\;\JH$&$\TwH$&&1 &0 &1 &0 &0 &0 &2 &A\\ \hline
\hline 

CleanUp &&&&&&1 &0 &5 &0 &0 &0 &2 &A\\ \hline
\end{tabular}
}
\end{center}
\caption{Some highlight of sound decisions made by both players: Since both players have full knowledge of each other's hands, Alex should avoid playing either $\FoC$ or $\FoD$. Doing so would allow Bob to respond immediately by collecting $\AC$ and $\AS$ with his $\FiC$, thereby securing 2 points and at least two clubs. Assuming instead that Alex plays $\AC$ and $\AS$ on his first move, Bob's optimal response is to collect using his $\FiC$. Finally, if Alex plays $\SeD$, Bob should respond by picking up one Ace using either his $\ThD$ or $\ThH$. It is emphasized that the strategy profile in Table~\ref{tab:deck9_exp} reflects all of these observations.
}
\label{fig:game_9}
\end{figure}

\begin{table}[htbp]
\centering
\scriptsize

\begin{tabular}{rrrrrrrrrrrrrrrr}
$\AC$ & $\AS$ & $\ThD$ & $\ThH$ & $\FoC$ & $\FoD$ & $\FiC$ & $\SeD$ & $\ND$ & $\QC$ & $\KD$ & $\KS$ & H & T & P & $\sigma$ \\
\hline
110 & 110 & 105 & 105 & 100 & 1   & 105 & 100 & 110 & 100 & 110 & 105 & 0 & 0 & 0 & 0.00 \\
110 & 110 & 105 & 105 & 1   & 100 & 105 & 100 & 110 & 100 & 110 & 105 & 0 & 0 & 0 & 0.00 \\
110 & 110 & 105 & 105 & 100 & 100 & 105 & 1   & 110 & 100 & 110 & 105 & 0 & 0 & 0 & 1.00 \\
110 & 110 & 105 & 105 & 100 & 100 & 105 & 100 & 110 & 1   & 110 & 105 & 0 & 0 & 0 & 1.00 \\
110 & 110 & 105 & -1  & 100 & 1   & 105 & 100 & 110 & 100 & 110 & 105 & 0 & 0 & 1 & 0.00 \\
110 & 110 & -1  & 105 & 100 & 1   & 105 & 100 & 110 & 100 & 110 & 105 & 0 & 0 & 1 & 0.00 \\
-10 & -10 & 105 & 105 & 100 & -9  & 109 & 100 & 110 & 100 & 110 & 105 & 0 & 0 & 1 & 1.00 \\
110 & 110 & 105 & 105 & 100 & 1   & 105 & 100 & 110 & 100 & -10 & 109 & 0 & 0 & 1 & 0.00 \\
110 & 110 & -1  & 105 & 1   & 100 & 105 & 100 & 110 & 100 & 110 & 105 & 0 & 0 & 1 & 0.00 \\
110 & 110 & 105 & -1  & 1   & 100 & 105 & 100 & 110 & 100 & 110 & 105 & 0 & 0 & 1 & 0.00 \\
110 & 110 & 105 & 105 & 1   & 100 & 105 & 100 & 110 & 100 & -10 & 109 & 0 & 0 & 1 & 0.00 \\
-10 & -10 & 105 & 105 & -9  & 100 & 109 & 100 & 110 & 100 & 110 & 105 & 0 & 0 & 1 & 1.00 \\
110 & 110 & 105 & 105 & 100 & 100 & 105 & 1   & 110 & 100 & -10 & 109 & 0 & 0 & 1 & 0.00 \\
110 & -10 & 109 & 105 & 100 & 100 & 105 & -9  & 110 & 100 & 110 & 105 & 0 & 0 & 1 & 0.25 \\
-10 & 110 & 105 & 109 & 100 & 100 & 105 & -9  & 110 & 100 & 110 & 105 & 0 & 0 & 1 & 0.25 \\
110 & 110 & 105 & 105 & 100 & 100 & -1  & 1   & 110 & 100 & 110 & 105 & 0 & 0 & 1 & 0.00 \\
-10 & 110 & 109 & 105 & 100 & 100 & 105 & -9  & 110 & 100 & 110 & 105 & 0 & 0 & 1 & 0.25 \\
110 & -10 & 105 & 109 & 100 & 100 & 105 & -9  & 110 & 100 & 110 & 105 & 0 & 0 & 1 & 0.25 \\
110 & 110 & -1  & 105 & 100 & 100 & 105 & 100 & 110 & 1   & 110 & 105 & 0 & 0 & 1 & 0.00 \\
110 & 110 & 105 & 105 & 100 & 100 & 105 & 100 & 110 & 1   & -10 & 109 & 0 & 0 & 1 & 1.00 \\
110 & 110 & 105 & 105 & 100 & 100 & -1  & 100 & 110 & 1   & 110 & 105 & 0 & 0 & 1 & 0.00 \\
110 & 110 & 105 & -1  & 100 & 100 & 105 & 100 & 110 & 1   & 110 & 105 & 0 & 0 & 1 & 0.00 \\
\end{tabular}
\caption{Near-Nash strategies for first turn of Figure~\ref{fig:game_9}'s deck}
\label{tab:deck9_exp}
\end{table}
\newpage

%% file: fwork.tex
\section{Future Work}

We conclude the paper by outlining two promising directions for future research.

First, we need to extend the current framework to the more realistic and challenging setting where players do not have access to the opponent’s hand information. This extension could proceed in two stages. First, the possible opponent deck compositions can be narrowed down to a small set of plausible candidates. Then, leveraging the external sampling technique introduced in Section~\ref{sec:exsample}, one could apply methods similar to those proposed in~\cite{brown2019deep} to train general tree-based models and use CFR to solve Pasur in the most general setting. The development and evaluation of this more general framework are left to future work.

Second, it would be valuable to investigate whether a single XGBoost model can be trained to generalize across multiple strategy profiles. This investigation remains within the setting where both players have full knowledge of each other’s hands, with the added complexity that multiple deck decompositions may be possible and gradually revealed as the game progresses. A key challenge in this context arises when decks share partial similarity—for example, two decks may have identical first-round compositions but differ in later rounds. Such differences can influence strategies even in the early stages, causing the same information sets to map to different actions. To address this, strategy training must account for this additional source of randomness.

Finally, another important direction for future work is to approximate the exploitability of the strategies trained in this paper. This would help validate convergence toward a true Nash equilibrium. Exploitability quantifies how far a strategy profile deviates from equilibrium, lower values indicate closer approximation to optimal play. A natural starting point would be the simplified setting where both players have full knowledge of each other's hands. Once exploitability is well-understood in this restricted scenario, approximation techniques can be applied to the general case to estimate the exploitability of the learned strategies. 

%% file: appendix.tex
\section{Appendix: RepeatBlocks}\label{sec:appendix}
RepeatBlocks is a generalization of \py{torch.repeat\_interleave} in which, instead of repeating individual elements, entire blocks of elements are repeated. Each block has a custom size and a custom repetition count.  The idea is best explained through an example. Consider the following
 \[
\py{t\_org = [1,2,3,4,5,6], t\_bsz = [2,3,1],t\_rpt = [1,3,2]}
 \]
Here \py{t\_org,t\_bsz,t\_rpt} are the input tensor, block sizes, and repeat counts respectively. The desired output denoted by $\py{t\_org} \otimes_{\py{t\_bsz}}\py{t\_rpt}$ is 
\[
\py{t\_org} \otimes_{\py{t\_bsz}}\py{t\_rpt = [1, 2, 3, 4, 5, 3, 4, 5, 3, 4, 5, 6, 6]}
\]
For convenience, let us denote \py{t\_rbk} = \py{t\_org} $\otimes_{\py{t\_bsz}}$\py{t\_rpt}.
To construct \py{t\_rbk}, it suffices to build the corresponding index tensor \py{t\_idx}.
\[
\py{t\_idx = [0, 1, 2, 3, 4, 2, 3, 4, 2, 3, 4, 5, 5]}
\]
Note that \py{t\_rbk = t\_org[t\_idx]}. Next, consider the corresponding start indices of each repeated block 
\[
\py{t\_blk = [0, 0, 2, 2, 2, 2, 2, 2, 2, 2, 2, 5, 5],}
\]
It is emphasized that using input tensors, \py{t\_blk} is given as below.
\[
 \py{t\_blk = (cat([0,t\_bsz[:-1]])$\otimes$\py{t\_rpt})}\otimes (\underbrace{\py{t\_bsz$\otimes$t\_rpt}}_{:=\py{t\_bls}}) 
\]
Deducting \py{t\_blk} from \py{t\_idx}, we arrive at the following tensor.
\[
\py{t\_pos = [0, 1, 0, 1, 2, 0, 1, 2, 0, 1, 2, 0, 0]}
\]
Next, 
\[
\py{arange(t\_blk.shape[0])-t\_pos = [ 0,  0,  2,  2,  2,  5,  5,  5,  8,  8,  8, 11, 12]}
\]
 where a simple examination shows that the RHS is equal
\[
\py{cumsum(cat([0, t\_bls[:-1]]))$\otimes$t\_bls}.
\]
Putting pieces together, we arrive at Algorithm \ref{alg:RepeatBlocks}.
\begin{algorithm}
\caption{RepeatBlocks }
\label{alg:RepeatBlocks}
\begin{algorithmic}[1]
\State Input \py{t\_org, t\_bsz, t\_rpt}
\Statex \textcolor{green!50!black}{\scriptsize \# \py{t\_org:input tensor, t\_bsz:blocks sizes, t\_rpt:repeat counts}}
\Statex \textcolor{green!50!black}{\scriptsize \# Ex: \py{t\_org = [1,2,3,4,5,6], t\_bsz = [2,3,1],t\_rpt = [1,3,2]}}
\State \py{t\_bgn} $\gets$ \py{cat([0,t\_bsz[:-1]])}$\otimes$\py{t\_rpt}
\Statex \textcolor{green!50!black}{\scriptsize \# Compute beginning indices of each output's block}
\Statex \textcolor{green!50!black}{\scriptsize \# Ex: \py{t\_bgn = [0, 2, 2, 2, 5, 5]}}
\State \py{t\_bls} $\gets$ \py{t\_bsz}$\otimes$\py{t\_rpt}
\Statex \textcolor{green!50!black}{\scriptsize \# Compute sizes of each output's block}
\Statex \textcolor{green!50!black}{\scriptsize \# Ex: \py{t\_bls = [2, 3, 3, 3, 1, 1]}}
\State \py{t\_blk} $\gets$ \py{t\_bgn}$\otimes$\py{t\_bls}
\Statex \textcolor{green!50!black}{\scriptsize \# t\_blk[i] = j i.e., output[i] belongs to block j}
\Statex \textcolor{green!50!black}{\scriptsize \# Ex: \py{t\_blk = [0, 0, 2, 2, 2, 2, 2, 2, 2, 2, 2, 5, 5]}}
\State \py{t\_csm} $\gets$ \py{cumsum(0,t\_bls[:-1])}$\otimes$\py{t\_bls}
\Statex \textcolor{green!50!black}{\scriptsize \# Ex: \py{t\_csm=[ 0,  2,  5,  8, 11, 12]$\otimes$t\_bls=[ 0,  0,  2,  2,  2,  5,  5,  5,  8,  8,  8, 11, 12]}}
\State \py{t\_pos} $\gets$ \py{arange(i\_blk)-t\_csm}
\Statex \textcolor{green!50!black}{\scriptsize \# Ex: \py{t\_pos=[0, 1, 0, 1, 2, 0, 1, 2, 0, 1, 2, 0, 0]}}
\State \py{t\_idx} $\gets$ \py{t\_pos+t\_blk}
\Statex \textcolor{green!50!black}{\scriptsize \# Ex: \py{t\_idx=[0, 1, 2, 3, 4, 2, 3, 4, 2, 3, 4, 5, 5]}}
\State \py{t\_rbk} $\gets$ \py{t\_org[t\_idx]}
\Statex \textcolor{green!50!black}{\scriptsize \# Ex: \py{t\_rbk=[1, 2, 3, 4, 5, 3, 4, 5, 3, 4, 5, 6, 6]}}

\State Output \py{t\_rbk} 
\Statex \textcolor{green!50!black}{\scriptsize \# denoted by \py{t\_org}$\otimes_{\py{t\_bsz}}$\py{t\_rpt}}
\end{algorithmic}
\end{algorithm}